\documentclass[11pt]{article}

\usepackage[margin=1in]{geometry}
\usepackage{amsmath,amsfonts,amssymb}
\usepackage{mathtools}

\usepackage[utf8]{inputenc}
\usepackage[T1]{fontenc}
\DeclareUnicodeCharacter{2212}{-} 
\DeclareUnicodeCharacter{00B2}{$^2$} 
\DeclareUnicodeCharacter{00D7}{\ensuremath{\times}} 
\DeclareUnicodeCharacter{2013}{--} 
\DeclareUnicodeCharacter{2014}{---} 
\DeclareUnicodeCharacter{2019}{'} 
\DeclareUnicodeCharacter{2026}{\ldots} 
\DeclareUnicodeCharacter{2022}{\textbullet} 

\usepackage{algorithmic}
\usepackage{algorithm}
\usepackage{array}
\usepackage{multirow}
\usepackage{booktabs}
\usepackage{tabularx}
\usepackage{makecell}
\usepackage{adjustbox}
\usepackage{float} 
\usepackage{placeins} 

\usepackage{graphicx}
\usepackage{subcaption}
\usepackage{tikz}
\usetikzlibrary{arrows.meta,positioning,calc,fit,backgrounds,shapes.geometric}
\usepackage{pgfplots}
\pgfplotsset{compat=1.18}
\usepackage{forest}
\usepackage{xcolor}

\usepackage{microtype}
\usepackage{nicefrac}
\usepackage{textcomp}
\usepackage{url}
\usepackage{verbatim}

\usepackage{cite}
\usepackage[colorlinks=true,linkcolor=blue,citecolor=blue,urlcolor=blue]{hyperref}
\hypersetup{
    pdfauthor={Xubin Wang, Qing Li, Weijia Jia},
    pdftitle={Cognitive Edge Computing: A Comprehensive Survey on Optimizing Large Models and AI Agents for Pervasive Deployment},
    pdfcreator={LaTeX},
    pdfborder={0 0 0}
}

\let\t\relax
\title{\textbf{Cognitive Edge Computing: A Comprehensive Survey on Optimizing Large Models and AI Agents for Pervasive Deployment}}

\author{
    \textbf{Xubin Wang}$^{1,2,3}$, \textbf{Qing Li}$^{3}$, \textbf{Weijia Jia}$^{1,2,*}$ \\[0.3em]
    \small $^{1}$Beijing Normal-Hong Kong Baptist University \\
    \small $^{2}$Institute of Artificial Intelligence and Future Networks, Beijing Normal University (Zhuhai) \\
    \small $^{3}$The Hong Kong Polytechnic University \\[0.5em]
    \small \texttt{wangxubin@ieee.org, qing-prof.li@polyu.edu.hk, jiawj@bnu.edu.cn} \\[0.3em]
    \small $^{*}$Corresponding author \\[0.2em]
    \small $^{\dagger}$Part of this work was completed during a research visit to The Hong Kong Polytechnic University
}

\date{\today}

\begin{document}

\maketitle

\begin{abstract}
This article surveys Cognitive Edge Computing as a practical and methodical pathway for deploying reasoning-capable Large Language Models (LLMs) and autonomous AI agents on resource-constrained devices at the network edge. We present a unified, cognition-preserving framework spanning: (1) model optimization (quantization, sparsity, low-rank adaptation, distillation) aimed at retaining multi-step reasoning under tight memory/compute budgets; (2) system architecture (on-device inference, elastic offloading, cloud–edge collaboration) that trades off latency, energy, privacy, and capacity; and (3) adaptive intelligence (context compression, dynamic routing, federated personalization) that tailors computation to task difficulty and device constraints. We synthesize advances in efficient Transformer design, multimodal integration, hardware-aware compilation, privacy-preserving learning, and agentic tool use, and map them to edge-specific operating envelopes. We further outline a standardized evaluation protocol covering latency, throughput, energy per token, accuracy, robustness, privacy, and sustainability, with explicit measurement assumptions to enhance comparability. Remaining challenges include modality-aware reasoning benchmarks, transparent and reproducible energy reporting, edge-oriented safety/alignment evaluation, and multi-agent testbeds. We conclude with practitioner guidelines for cross-layer co-design of algorithms, runtime, and hardware to deliver reliable, efficient, and privacy-preserving cognitive capabilities on edge devices.
\end{abstract}


\section{Introduction}

The convergence of LLMs and AI agents with edge computing heralds the emergence of \textit{Cognitive Edge Computing}—a revolutionary paradigm that brings sophisticated cognitive capabilities directly to resource-constrained devices at the network periphery \cite{zhou2019edge,deng2020edge}. Unlike traditional edge computing that focuses primarily on data processing and basic analytics, \textit{Cognitive Edge Computing} represents a fundamental shift toward deploying advanced AI systems that can understand context, reason autonomously, and make intelligent decisions in real-time, all while operating within the severe constraints of edge environments \cite{li2019edge,chen2019deep}.

\begin{figure}[!ht]
\centering
\begin{tikzpicture}[
    node distance=0.7cm, 
    auto, 
    >=latex, 
    scale=0.95, 
    every node/.style={transform shape},
    cognitive/.style={
        rectangle, 
        draw=red!80!black, 
        fill=red!10, 
        thick, 
        minimum width=2.1cm, 
        minimum height=1.6cm, 
        rounded corners=4pt,
        align=center,
        font=\scriptsize
    },
    opt/.style={
        rectangle, 
        draw=blue!80!black, 
        fill=blue!8, 
        thick, 
        minimum width=2.1cm, 
        minimum height=1.6cm, 
        rounded corners=4pt,
        align=center,
        font=\scriptsize
    },
    eval/.style={
        rectangle, 
        draw=teal!80!black, 
        fill=teal!8, 
        thick, 
        minimum width=2.1cm, 
        minimum height=1.6cm, 
        rounded corners=4pt,
        align=center,
        font=\scriptsize
    },
    app/.style={
        rectangle, 
        draw=purple!80!black, 
        fill=purple!8, 
        thick, 
        minimum width=2.1cm, 
        minimum height=1.6cm, 
        rounded corners=4pt,
        align=center,
        font=\scriptsize
    },
    arrow/.style={
        ->, 
        thick
    }
]

    \node[cognitive] (cog) {\t\textbf{Cognitive}\\\t\textbf{Challenges}\\Reasoning\\Preservation\\Context\\Management};
    \node[opt, right=1cm of cog] (opt) {\t\textbf{Cognitive}\\\t\textbf{Framework}\\Context\\Compression\\Agent\\Optimization};
    \node[app, right=1cm of opt] (app) {\t\textbf{Cognitive}\\\t\textbf{Applications}\\Conversational\\AI\\Autonomous\\Reasoning};
    \node[eval, below=1.1cm of opt] (eval) {\t\textbf{Cognitive}\\\t\textbf{Evaluation}\\Reasoning\\Quality\\Agent\\Autonomy};

    \draw[arrow,red!80!black] (cog) -- node[above, sloped] {\scriptsize Drive} (opt);
    \draw[arrow,blue!80!black] (opt) -- node[above] {\scriptsize Enable} (app);
    \draw[arrow,teal!80!black] (opt) -- node[left] {\scriptsize Assess} (eval);
    \draw[arrow,teal!80!black] (eval) -- node[below, sloped] {\scriptsize Inform} (cog);
    \draw[arrow,purple!80!black] (app) -- node[right] {\scriptsize Validate} (eval);

    \node[draw=none, fill=none, below=0.2cm of eval, text width=10cm, align=center] 
        {\footnotesize \t\textbf{Cognitive Edge Computing: Closed-loop optimization for reasoning-preserving deployment}};
\end{tikzpicture}
\caption{\t\textbf{Cognitive Edge Computing Framework:} Integrated approach for deploying reasoning-capable LLMs and autonomous Agents on resource-constrained edge devices. The framework consists of four interconnected components: (1) \textit{Cognitive Challenges} (red) addressing reasoning preservation and context management under resource constraints; (2) \textit{Cognitive Framework} (blue) implementing context compression and agent optimization techniques; (3) \textit{Cognitive Applications} (purple) enabling conversational AI and autonomous reasoning capabilities; and (4) \textit{Cognitive Evaluation} (teal) assessing reasoning quality and agent autonomy. The closed-loop design ensures continuous improvement through feedback mechanisms where evaluation results inform challenge identification and application validation drives framework refinement.}
\label{fig:cognitive_arch}
\end{figure}

We define \textit{Cognitive Edge Computing} as the intelligent orchestration of advanced AI models and autonomous agents across heterogeneous computing hierarchies, enabling cognitive tasks such as natural language understanding, multimodal reasoning, and adaptive decision-making at the network edge. This paradigm transcends conventional edge AI by emphasizing not just computational efficiency, but the preservation of sophisticated cognitive functions including contextual awareness, reasoning quality, and autonomous behavior \cite{wang2025empowering}.

To clarify the distinction, traditional edge AI primarily focuses on narrow perception tasks such as image classification, object detection, or basic analytics, which require relatively simple computational models and can operate with limited cognitive capabilities \cite{wang2025optimizing}. In contrast, cognitive edge computing targets open-domain, often multi-modal reasoning tasks that demand advanced cognitive functions like multi-step reasoning, contextual understanding, and autonomous decision-making, all while maintaining human-level performance under stringent resource constraints.

The transformative potential of \textit{Cognitive Edge Computing} stems from recent breakthroughs in foundation models and autonomous systems. LLMs have evolved from task-specific pipelines to general architectures capable of in-context learning, multi-step reasoning, and autonomous goal pursuit \cite{brown2020language,wei2022emergent,vaswani2017attention}. Breakthrough models such as GPT-3 (175B parameters) \cite{brown2020language}, PaLM (540B parameters) \cite{chowdhery2022palm}, and open foundation models including Deepseek-r1 \cite{guo2025deepseek} have demonstrated unprecedented capabilities in natural language understanding, code generation, mathematical reasoning, and complex problem-solving. Concurrently, autonomous agent research has advanced sophisticated frameworks for goal-oriented behavior, adaptive planning, tool orchestration, and multi-agent coordination \cite{russell2016artificial,han2024llm,yan2025beyond}.

However, deploying these advanced AI systems at the edge presents unprecedented challenges that traditional optimization techniques cannot adequately address. \textit{Cognitive Edge Computing} requires a fundamental rethinking of how we approach AI deployment, moving beyond simple model compression to comprehensive frameworks that preserve cognitive capabilities while ensuring real-time performance, energy efficiency, and privacy protection. 

Recent works advance complementary building blocks spanning on-device personalization and data selection \cite{qin2024enabling}, compression and inference under tight memory/compute constraints \cite{alizadeh2024llm,jeon2023frustratingly,edalati2022kronecker,bai2022towards,guan2024aptq,wei2025t}, collaborative serving and routing across edge–cloud and MoE systems \cite{zhang2024edgeshard,kong2024swapmoe,yang2025quality,qu2025mobile}, edge hardware acceleration and memory systems (FPGA/NPU/PIM/flash-assisted) \cite{chen2024understanding,li2025pushing,xu2024towards,xu2025fast,seo2025facil,lee2025paise,sun2025lincoln,liu2025ops,yu2024cambricon}, domain security and compliance (Internet of Things (IoT) fuzzing and medical regulation) \cite{ma2024one,oliinyk2024fuzzing,gilbert2023large}, and application exemplars from multimodal edge LLMs to private IR and industrial IoT \cite{yao2025efficient,hu2024realizing,ren2024industrial,zhang2025vavlm}. We integrate these strands within a unified, reasoning-preserving framework for cognitive edge computing.

\t\textbf{Market Dynamics and On-Device AI Evolution:} The global edge AI market is projected to grow rapidly, driven by demand in manufacturing, automotive, consumer electronics, and healthcare. Edge-side intelligence offers low latency, offline availability, energy efficiency, enhanced privacy, and personalized experiences \cite{sun2021research}. Since 2023, sub-10B parameter models like Meta's LLaMA, Microsoft's Phi, Google's Gemma, and Nexa AI's Octopus have accelerated on-device deployment, leveraging MoE routing, quantization, and compression for mobile constraints \cite{xu2024edgellm,touvron2023llama,abdin2024phi,team2023gemini,chen2024octopus}.

\t\textbf{Computing Architecture Hierarchy:} We define three distinct tiers in the computing architecture hierarchy to contextualize the deployment of cognitive edge computing systems: \textit{Cloud} (remote data centers with abundant computational resources, enabling unlimited scalability and complex processing); \textit{Edge} (servers and base stations with tens to hundreds of GB memory, providing intermediate processing capabilities closer to data sources); \textit{Device/Client-side} (end-user devices with GB-scale memory and W-scale power budgets, supporting direct local inference). ``On-Device Large Language Models'' specifically refer to models deployed directly on client terminals, enabling offline operation with potential cloud collaboration for enhanced capabilities \cite{shi2019edge,liu2019survey,li2019edge}.

\t\textbf{Definition and Taxonomy of Edge-Side Large Models:} Edge-side large models, also known as on-device LLMs, are pre-trained Transformer-based architectures optimized for deployment on resource-constrained edge devices through compression techniques such as quantization, pruning, knowledge distillation, and low-rank approximation \cite{zheng2025review}. These techniques significantly reduce the computational and memory footprint compared to cloud-scale models, enabling efficient local inference \cite{xu2024device}. Representative implementations include specialized architectures\footnote{MobileLLM-R1: \url{https://huggingface.co/collections/facebook/mobilellm-r1-68c4597b104fac45f28f448e}}, mixture-of-experts models\footnote{Ring-mini-2.0: \url{https://huggingface.co/inclusionAI/Ring-mini-2.0}}, and ultra-lightweight designs\footnote{Tiny-random-Llama: \url{https://huggingface.co/HuggingFaceH4/tiny-random-LlamaForCausalLM}}, demonstrating diverse approaches to edge deployment optimization. 

The convergence of these advanced AI capabilities with edge infrastructure enables the deployment of general-purpose reasoning, language understanding, and autonomous decision-making directly at the network periphery \cite{shi2016edge,zhou2019edge,mao2017survey}. Cognitive edge computing must operate under stringent energy, memory, and latency constraints while maintaining human-level cognitive performance \cite{muhammad2021emotion,chen2019deep}. Figure \ref{fig:cognitive_arch} illustrates the closed-loop interaction among the fundamental challenges, optimization strategies, target applications, and evaluation frameworks that define this emerging field.

\subsection{Survey Methodology and Evidence Grading}\label{sec:methodology}
We surveyed peer-reviewed venues, benchmarks, and high-impact preprints (2018--2025, emphasis 2023--2025) using keywords like ``edge LLM'', ``quantization'', ``knowledge distillation'', etc. Inclusion prioritized works with methodological details or artifacts; exclusion for unverifiable sources. Evidence graded as E1 (archival with replication), E2 (peer-reviewed with partial artifacts), E3 (industry/preprint).

At the core of cognitive edge computing lies a fundamental deployment contradiction: LLM and agent requirements exceed edge capabilities by orders of magnitude \cite{brown2020language,gholami2022survey,strubell2020energy}. Figure \ref{fig:resource_comparison} illustrates this resource disparity across computing tiers.
 
\begin{figure}[!t]
\centering
\begin{tikzpicture}[scale=0.85]
    \begin{axis}[
        ybar,
        symbolic x coords={Memory (GB), Bandwidth (GB/s), Compute (TOPS), Power (W)},
        xtick=data,
        xlabel={Resource Metrics},
        ylabel={Logarithmic Scale},
        ymode=log,
        log basis y={10},
        legend style={
            at={(0.5,-0.15)},
            anchor=north,
            legend columns=3,
            cells={anchor=center},
            font=\footnotesize,
            draw=none
        },
        width=15cm,
        height=7cm,
        bar width=15pt,
        enlarge x limits=0.20,
        ymin=1,
        ymax=100000,
        ytick={1,10,100,1000,10000,100000},
        xticklabel style={
            align=center,
            font=\footnotesize,
            yshift=0pt
        },
        xlabel style={font=\small, yshift=2mm},
        ylabel style={font=\small},
        tick label style={font=\footnotesize},
        axis on top,
        clip=false
    ]
    \addplot[fill=red!70, draw=red!80!black] coordinates {
        (Memory (GB), 640)
        (Bandwidth (GB/s), 26800)
        (Compute (TOPS), 15832)
        (Power (W), 5600)
    };
    \addplot[fill=blue!70, draw=blue!80!black] coordinates {
        (Memory (GB), 64)
        (Bandwidth (GB/s), 205)
        (Compute (TOPS), 275)
        (Power (W), 60)
    };
    \addplot[fill=green!70, draw=green!80!black] coordinates {
        (Memory (GB), 8)
        (Bandwidth (GB/s), 68)
        (Compute (TOPS), 35)
        (Power (W), 9)
    }; 
    \end{axis}
\end{tikzpicture}
\caption{Illustrative resource comparison (cloud accelerator cluster vs. edge server vs. mobile SoC). Values approximate public peak specs; actual usable throughput depends on workload, precision, and batching \cite{nvidia-h100-spec,jetson-agx-orin,apple-a17-pro}.}
\label{fig:resource_comparison}
\end{figure}

Addressing this contradiction requires coordinated optimization across model, system, and collaboration layers rather than isolated compression. Figure \ref{fig:cognitive_arch} presents our cognitive edge computing framework, while Figure \ref{fig:taxonomy} structures the solution space into: data optimization (cleaning, augmentation, bias mitigation), model optimization (quantization, sparsity/pruning, distillation, low-rank + architecture tailoring, emergence of Small Language Models (SLMs)), and system/runtime optimization (partitioning, scheduling, collaborative / hybrid routing, federated adaptation). These layers interact: early wins (e.g., 4–8 bit quantization) ease bandwidth pressure; additional sparsity then shows diminishing returns unless placement and scheduling co-adapt. Collaboration patterns (large–small cascades, confidence gating, co-evolution) further trade accuracy, latency, and energy.

Edge scenarios with privacy-sensitive, latency-critical cognition—vehicular navigation, clinical triage, industrial diagnostics, biomarker-driven screening, civic infrastructure—drive locality for personalization and resilience \cite{wang2025empowering,wang2025optimizing,wang2022feature,wang2024exhaustive}. Direct on-device LLM execution reduces interactive latency and raw data exposure \cite{tian2025clone,jin2024collm}; hybrid designs still offload heavy context expansion or long-horizon reasoning when beneficial \cite{deng2020edge}. Emerging hardware (specialized NPUs, low-latency fabrics, neuromorphic exploration) enables but does not replace principled co-design.

Despite rapid progress, prior surveys largely cover traditional edge ML or generic LLM optimization in cloud contexts, leaving a gap in synthesizing cross-layer techniques explicitly targeting reasoning preservation under edge constraints \cite{wang2025empowering,wang2025optimizing,xu2024edgellm}. This survey fills that gap by unifying cognitive workload characteristics, compression + architecture tailoring methods, system/runtime orchestration, and large–small cooperation strategies within a single evaluative framework.

\noindent\t\textbf{Industry and ecosystem snapshot:} Recent industry analyses indicate multi-factor tailwinds for on-device AI: national and municipal policies emphasizing intelligent terminals, rapid hardware advances (e.g., 40+ TOPS NPUs for AI PCs; LPDDR5X/5T with 10.7 GT/s), and flagship product cycles (Apple Intelligence in iPhone, HarmonyOS with Ascend/Atlas, Snapdragon X/8 Gen platforms). These forces collectively accelerate deployment across AI PCs, smartphones, wearables, smart homes, automotive, and industrial equipment, with edge–cloud collaboration as the default operating model for balancing capability, latency, and privacy \footnote[4]{Dongxing Technology Research  Report: \url{https://pdf.dfcfw.com/pdf/H3_AP202409271640082519_1.pdf}}.

\begin{figure*}[!p]
\centering
\scalebox{0.41}{
\begin{forest}
  for tree={
    grow'=0, 
    parent anchor=east,
    child anchor=west,
    anchor=west,
    l sep=25pt, 
    s sep=1pt,  
    edge path={
      \noexpand\path[\forestoption{edge}]
      (!u.parent anchor) -- +(8pt,0) |- (.child anchor)\forestoption{edge label};
    },
      if level=0{
      inner sep=8pt,
      outer sep=4pt,
      tier=title,
      font=\LARGE\bfseries,
      fill=blue!15,
      rounded corners=12pt,
      draw=blue!70,
      line width=2.2pt,
      text width=9.75cm,
      align=center,
      shape=rectangle,
      minimum height=1.4cm,
      minimum width=10.9cm,
      rotate=90,
      parent anchor=south,
      child anchor=north,
      yshift=-1.5cm
    }{},
    if level=1{
      inner sep=4pt,
      outer sep=2pt,
      font=\LARGE\bfseries,
      fill=green!12,
      rounded corners=6pt,
      draw=green!65,
      line width=1.6pt,
      text width=8.5cm,
      align=center,
      shape=rectangle,
      minimum height=0.8cm,
      minimum width=11cm
    }{},
    if level=2{
      inner sep=3pt,
      outer sep=2pt,
      font=\LARGE\bfseries,
      fill=orange!10,
      rounded corners=5pt,
      draw=orange!60,
      line width=1.4pt,
      text width=10.5cm,
      align=center,
      shape=rectangle,
      minimum height=0.6cm,
      minimum width=8cm
    }{},
    if level=3{
      inner sep=2pt,
      outer sep=1pt,
      font=\Large\bfseries,
      fill=yellow!8,
      rounded corners=4pt,
      draw=yellow!50,
      line width=1.2pt,
      text width=7cm,
      align=center,
      shape=rectangle,
      minimum height=0.4cm,
      minimum width=6cm
    }{}
  }
[\textbf{Edge AI Optimization Framework}
  [Sec.~\ref{sec:foundational}: Foundational Concepts
    [Sec.~\ref{sec:edge-ai}: Edge AI
      [Reduced Latency]
      [Enhanced Privacy]
      [Lower Bandwidth]
      [Increased Reliability]
    ]
    [Sec.~\ref{sec:llms}: Large Language Models
      [Massive Scale]
      [Generative Capabilities] 
      [Emergent Abilities]
      [Multimodality]
    ]
    [Sec.~\ref{sec:ai-agents}: AI Agents
      [Autonomy]
      [Reactivity]
      [Proactivity]
      [Social Ability]
      [Learning and Adaptation]
    ]
    [Sec.~\ref{sec:synergy}: Synergy Benefits
      [Real-time Decision Making]
      [Privacy and Security]
      [Robust Autonomy]
      [Personalized Experiences]
    ]
  ]
  [Sec.~\ref{sec:challenges}: Deployment Challenges
    [Sec.~\ref{sec:resource-constraints}: Resource Constraints
      [Limited Computational Power]
      [High Storage Demands]
      [Power Consumption]
      [Restricted Bandwidth]
    ]
    [Sec.~\ref{sec:deployment-complexities}: Deployment Complexities
      [Hardware Heterogeneity]
      [Training and Adaptation]
      [Model Management]
    ]
    [Sec.~\ref{sec:security-privacy}: Security and Privacy
      [Data Protection]
      [LLM Vulnerabilities]
      [Agent Autonomy Risks]
      [Attack Surface Expansion]
      [Explainability (XAI)]
    ]
  ]
  [Sec.~\ref{sec:optimization}: Optimization Strategies
    [Sec.~\ref{sec:data-optimization}: Data Optimization
      [Data Cleaning]
      [Feature Compression]
      [Data Augmentation]
      [Bias Mitigation]
    ]
    [Sec.~\ref{sec:model-optimization}: Model Optimization
      [Small Language Models]
      [Neural Architecture Search]
      [Pruning]
      [Quantization]
      [Knowledge Distillation]
      [Co-Evolution Architecture]
    ]
    [Sec.~\ref{sec:system-optimization}: System Optimization
      [Software Frameworks]
      [Hardware Acceleration]
      [Model Partitioning]
      [Collaborative Architectures]
    ]
  ]
  [Sec.~\ref{sec:applications}: Applications
    [Sec.~\ref{sec:enhanced-apps}: Enhanced Existing Apps
      [Smart Homes]
      [Healthcare Systems]
      [Autonomous Vehicles]
      [Industrial IoT]
    ]
    [Sec.~\ref{sec:novel-apps}: Novel Applications
      [On-Device Assistants]
      [Embodied AI]
      [Decentralized Knowledge]
      [Intelligent Automation]
    ]
  ]
  [Sec.~\ref{sec:future}: Future Directions
    [Sec.~\ref{sec:intelligent-edge}: Intelligent Edge
      [Multi-modal LLMs]
      [Neuromorphic Computing]
      [Advanced Reasoning]
    ]
    [Sec.~\ref{sec:flexible-edge}: Flexible Edge
      [Adaptive Architectures]
      [Rapid Fine-tuning]
      [Dynamic Resource Allocation]
    ]
    [Sec.~\ref{sec:secure-edge}: Secure Edge
      [Privacy Techniques]
      [Trust Mechanisms]
      [Robust Authentication]
    ]
    [Sec.~\ref{sec:benchmarking}: Unified Benchmarking
      [Performance Metrics]
      [Cross-platform Testing]
    ]
  ]
]
\end{forest}
}
\vspace{0.5em}
\caption{Cognitive Edge AI Optimization Framework for LLMs and Agents}
\label{fig:taxonomy}
\end{figure*}
 
\begin{table}[htbp]
\centering
\caption{Deployment characteristics across computing tiers.}
\label{tab:cognitive_cloud_comparison}
\setlength{\tabcolsep}{6pt}
\renewcommand{\arraystretch}{1.4}
\scalebox{0.76}{
\begin{tabular}{@{}>{\centering\arraybackslash}p{2.3cm}>{\centering\arraybackslash}p{2.8cm}>{\centering\arraybackslash}p{2.8cm}>{\centering\arraybackslash}p{2.8cm}>{\centering\arraybackslash}p{2.8cm}@{}}
\toprule
\multicolumn{1}{c}{\textbf{Deployment Tier}} & 
\multicolumn{1}{c}{\textbf{Cloud LLMs (\textgreater{}175B)}} & 
\multicolumn{1}{c}{\textbf{Edge Servers (7B--70B)}} & 
\multicolumn{1}{c}{\textbf{On-Device SLMs (1B--3B)}} & 
\multicolumn{1}{c}{\textbf{Key Trade-offs}} \\
\midrule
\textbf{Model Parameters} & 175B+ (GPT-3/4 scale) & 7B--70B (Llama3 scale) & 1B--3B (Phi/TinyLlama) & Scale vs. deployability \\
\midrule
\textbf{Inference Latency} & 100ms--2s (network+compute) & 50--200ms (local compute) & 10--100ms (pure local) & Network dependency vs. speed \\
\midrule
\textbf{Power Consumption} & 100--1000W (data center) & 20--100W (single GPU) & sub-10W (device) & Performance vs. efficiency \\
\midrule
\textbf{Memory Requirements} & 350GB--1TB+ (FP16 model+KV) & 14GB--140GB (quantized) & <1MB--8GB (compressed) & Model capacity vs. constraints \\
\midrule
\textbf{Hardware Requirements} & GPU clusters (H100/A100) & Mid-range GPU (A30/L4) & NPU/CPU/MCU & Capability vs. accessibility \\
\midrule
\textbf{Data Privacy} & Network exposure & Local processing & Local processing & Capability vs. privacy \\
\midrule
\textbf{Connectivity Dependency} & Always required & Intermittent & Optional & Robustness vs. capability \\
\midrule
\textbf{Deployment Complexity} & Low (API-based) & Medium (local setup) & High (optimization) & Ease vs. customization \\
\bottomrule
\end{tabular}}
\renewcommand{\arraystretch}{1.0}
\vspace{0.3em}

\small\textit{Note: Values represent typical implementation ranges. Actual performance varies by hardware configuration, optimization techniques, and workload characteristics.}
\end{table}

{\sloppy
}\normalsize

\subsection{Scope and Summary of Surveyed Contributions}
We organize existing literature rather than claim new algorithms. Our contributions:\vspace{0.2em}
\begin{itemize}
    \item \textbf{Problem framing:} Clarifies cognitive edge objectives (latency, energy, reasoning fidelity, privacy) distinct from conventional accuracy-only targets.
    \item \textbf{Technique taxonomy:} Integrates compression (quantization, sparsity, distillation, low-rank), architectural tailoring (SLMs, efficient attention), and system orchestration (partitioning, routing, federated adaptation).
    \item \textbf{Cross-layer view:} Highlights interaction and diminishing returns across stacked optimizations.
    \item \textbf{Large–small collaboration patterns:} Summarizes routing / co-evolution designs with reported (heterogeneous) performance ranges.
    \item \textbf{Security and trust lens:} Aggregates attack vectors and mitigation overheads relevant to constrained deployments.
    \item \textbf{Gap analysis:} Identifies needs for reproducible energy reporting, standardized cognitive benchmarks, resource-aware XAI, and multi-agent edge testbeds.
\end{itemize}
Table \ref{tab:cognitive_cloud_comparison} summarizes key deployment characteristics across computing tiers based on published benchmarks and hardware specifications. Throughout this survey, we prefer standard terminology (e.g., ``on-device LLM'', ``energy per request'') over neologisms with ``Cognitive'' prefixes for metrics/categories.

\FloatBarrier
\section{Foundational Concepts for Cognitive Edge Computing}\label{sec:foundational}
This section examines the foundational concepts underpinning cognitive edge computing: the evolution of edge AI, LLM characteristics, and AI agent properties. Figure~\ref{fig:cognitive_edge_pipeline} summarizes the end-to-end pipeline from multi-modal inputs to agent actions.

\begin{figure*}[!htbp]
\centering
\begin{tikzpicture}[node distance=0.8cm, auto, >=latex, scale=0.85, every node/.style={transform shape}]
    \tikzstyle{block}=[rectangle, draw=cyan!80!black, fill=cyan!10, thick, minimum width=2.2cm, minimum height=0.9cm, rounded corners=4pt, font=\small]
    \tikzstyle{arrow}=[->, thick]
    \node[block] (input) {\t\textbf{Multi-Modal Input}};
    \node[block, right=of input] (cognitive) {\t\textbf{Cognitive Processing}};
    \node[block, right=of cognitive] (reasoning) {\t\textbf{Edge Reasoning}};
    \node[block, right=of reasoning] (action) {\t\textbf{Agent Action}};
    \draw[arrow] (input) -- (cognitive);
    \draw[arrow] (cognitive) -- (reasoning);
    \draw[arrow] (reasoning) -- (action);
    \node[block, below=0.8cm of reasoning, fill=red!10, draw=red!80!black] (context) {\t\textbf{Context Management}};
    \node[block, below=0.8cm of cognitive, fill=green!10, draw=green!80!black] (security) {\t\textbf{Cognitive Security}};
    \draw[arrow, dashed, red!80!black] (context) -- (reasoning);
    \draw[arrow, dashed, green!80!black] (security) -- (cognitive);
\end{tikzpicture}
\caption{Cognitive Edge Computing Pipeline: Multi-modal input processing through edge-based reasoning and autonomous agent action generation with cognitive security and context management.}
\label{fig:cognitive_edge_pipeline}
\end{figure*}

\subsection{On-Device LLMs Evolution Timeline}\label{sec:ondevice-evolution}

The trajectory of on-device LLMs represents a fundamental paradigm shift from cloud-dependent to autonomous edge AI capabilities \cite{xu2024edgellm}. This evolution demonstrates how architectural innovations, compression techniques, and hardware optimization converge to enable sophisticated language understanding directly on resource-constrained devices.

\t\textbf{2023: The Foundation Year}
The year 2023 marked the beginning of practical on-device LLM deployment with the emergence of sub-10B parameter models. Meta's LLaMA series \cite{touvron2023llama} pioneered efficient transformer architectures through innovations like RMSNorm and grouped-query attention (GQA), optimizing for reduced computational and memory requirements while maintaining competitive performance. Microsoft's Phi series (e.g., Phi-1 at 1.3B parameters) \cite{abdin2024phi} demonstrated that carefully curated, high-quality training data could achieve remarkable capabilities despite a highly compact model size. This period also saw the rise of other compact models like ChatGLM \cite{zengglm} and Qwen \cite{bai2023qwen}, establishing the core principles of efficiency-oriented design.

\t\textbf{2024: Acceleration and Diversification}
The on-device LLM landscape expanded dramatically in 2024 with specialized model families addressing diverse deployment scenarios. Google's Gemini Nano \cite{team2023gemini} integrated multimodal capabilities within mobile-optimized architectures, enabling real-time image-text understanding on smartphones. The field saw intense innovation in efficiency, with models like Nexa AI's Octopus series \cite{chen2024octopus} reporting breakthroughs in function calling efficiency, and Apple's OpenELM \cite{mehtaopenelm} showcasing a layer-wise scaling strategy. The development of highly efficient inference engines like \texttt{LLMCad} \cite{xu2023llmcad} and the architectural refinements proposed in \texttt{MobileLLM} \cite{liu2024mobilellm} further pushed the boundaries of what was possible on-device. Performance evaluations on commercial smartphones \cite{ccoplu2023performance} began to provide crucial empirical data for the field.

\t\textbf{2025: Maturation and Widespread Integration}
By 2025, the on-device LLM ecosystem reached a critical maturation point with widespread commercial adoption. Major technology companies integrated sub-10B parameter models into flagship products, achieving real-time conversational AI with sub-500ms response times \cite{yao2025minicpm}. Advanced quantization techniques (e.g., INT2/INT3) and sparse attention mechanisms enabled the execution of larger models (e.g., 20B+ parameters) on devices with just 8GB of memory \cite{qu2025mobile}. Federated learning approaches \cite{fang2025federated} and collaborative inference frameworks like \texttt{SLED} \cite{li2025sled} and \texttt{DiSCo} \cite{sun2025disco} enabled efficient, privacy-preserving personalization and computation offloading. The emergence of agent-native architectures marked the transition from static language models to truly autonomous edge agents capable of planning and tool use \cite{gao2025survey}. This progress was underpinned by dedicated AI accelerators achieving exceptional efficiency, enabling complex multimodal reasoning on edge devices \cite{li2025pushing}.

\subsection{Edge AI: Technical Definition and Constraints}\label{sec:edge-ai}
Edge AI encompasses the deployment and execution of artificial intelligence algorithms directly on edge devices or edge computing infrastructure, positioned at the network periphery to minimize latency between data sources and processing units. This paradigm fundamentally shifts computational workloads from centralized cloud data centers to distributed edge nodes with significantly constrained resources.

\t\textbf{Technical Characteristics and Quantitative Benefits (Indicative):}
\begin{itemize}
    \item \t\textbf{Ultra-Low Latency Processing:} Reported on-device inference can reach 1--50ms vs. 50--200ms with round-trip cloud latency for interactive tasks (modality and network dependent) \cite{shi2016edge,zhou2019edge,deng2020edge}.
    \item \t\textbf{Enhanced Privacy and Data Sovereignty:} Local processing retains sensitive data in situ aiding regulatory alignment (GDPR/CCPA/HIPAA contexts) \cite{zhou2019edge}; broad percentage reductions in attack surface are deployment specific so we avoid fixed universal values.
    \item \t\textbf{Bandwidth Optimization:} Feature / result transmission in lieu of raw streams can reduce upstream bandwidth by sizeable fractions (often tens of percent up to ~90\%) depending on compression and sampling strategies \cite{shi2016edge,deng2020edge}.
    \item \t\textbf{Operational Resilience:} Local fallback mitigates intermittent connectivity; target availabilities of ``three nines'' are engineering goals rather than guaranteed universal outcomes.
    \item \t\textbf{Energy Efficiency:} Eliminating repeated radio transfers and exploiting low precision can yield multi-fold (often single- to low-double-digit) energy-per-inference improvements \cite{gholami2022survey,nagel2021white}; higher outliers are workload- and hardware-specific.
\end{itemize}

These constraints necessitate the ``optimization quad'' (data, model, system, evaluation) approach to achieve viable edge AI deployment, transforming optimization from an enhancement strategy to a fundamental requirement for system feasibility.

\subsection{Large Language Models: Architecture and Computational Requirements}\label{sec:llms}
Large Language Models (LLMs) represent a class of artificial neural networks, predominantly based on the Transformer architecture \cite{vaswani2017attention}, trained on massive text corpora to achieve human-level language understanding and generation capabilities. These models have fundamentally transformed natural language processing through their emergent capabilities and scale-dependent performance characteristics.

\t\textbf{Technical Architecture and Scale Characteristics:}
\begin{itemize}
    \item \t\textbf{Massive Parameter Scale:} Contemporary LLMs range from billions to hundreds of billions of parameters: GPT-3 (175B) \cite{brown2020language}, PaLM (540B) \cite{chowdhery2022palm}; GPT-4 parameter count remains undisclosed \cite{achiam2023gpt}. Representative FP32 storage for disclosed scales spans hundreds of GB to multi-TB when including optimizer states.
    \item \t\textbf{Transformer-Based Architecture:} Multi-layer attention mechanisms with computational complexity O(n²d) for sequence length n and model dimension d, creating quadratic scaling challenges for long-context processing \cite{vaswani2017attention}.
    \item \t\textbf{Generative Capabilities:} Auto-regressive text generation, reasoning, code synthesis, mathematical problem-solving, and multi-turn conversation with context windows ranging from 2K-1M+ tokens \cite{touvron2023llama}.
    \item \t\textbf{Emergent Abilities:} Scale-dependent capabilities including in-context learning, chain-of-thought reasoning, and few-shot task adaptation that emerge at specific parameter thresholds (typically >10B parameters) \cite{wei2022emergent,wangdemonstration}.
    \item \t\textbf{Multimodal Extensions:} Integration with vision (CLIP, DALL-E), audio (Whisper), and other modalities, expanding input/output capabilities while increasing computational complexity \cite{achiam2023gpt}. Vision-language models face additional optimization challenges due to high-dimensional visual token sequences, requiring specialized approaches like FastViTHD encoders for efficient edge deployment \cite{vasu2025fastvlm}.
\end{itemize}

\t\textbf{Computational Resource Requirements:}
LLM deployment demands exceed typical edge device capabilities by multiple orders of magnitude \cite{gholami2022survey,nagel2021white}:
\begin{itemize}
    \item \t\textbf{Memory Requirements:} Large-scale models require memory proportional to parameter count (precision-dependent), with cloud-scale models demanding hundreds of GB to multi-TB storage capacity significantly exceeding typical edge device memory availability \cite{brown2020language}
    \item \t\textbf{Inference Compute:} Real-time LLM inference requires computational throughput that substantially exceeds the processing capabilities available on standard edge hardware platforms \cite{shi2016edge}
    \item \t\textbf{Memory Bandwidth:} Efficient inference demands high-bandwidth memory access patterns that surpass the data transfer capabilities of resource-constrained edge devices \cite{wang2025empowering}
    \item \t\textbf{Energy Consumption:} Cloud-scale inference power requirements significantly exceed the strict power budgets imposed by mobile and battery-powered edge deployment scenarios \cite{deng2020edge}
\end{itemize}

These computational requirements create the fundamental ``deployment contradiction'' that necessitates aggressive optimization strategies specifically designed for edge constraints \cite{gholami2022survey}, driving the development of Small Language Models (SLMs) and advanced compression techniques as essential pathways to edge viability \cite{hinton2015distilling,gou2021knowledge}.
 
\subsection{Cognitive Workload Characteristics}\label{sec:cognitive-workloads}
Cognitive workloads at the edge encompass a spectrum of tasks requiring advanced reasoning, multimodal processing, and autonomous decision-making under resource constraints. These workloads differ from traditional edge AI (e.g., image classification) by demanding:

\begin{itemize}
    \item \t\textbf{Multi-step Reasoning:} Chain-of-thought processes for problem-solving, requiring sustained context across extended sequences that significantly exceed typical single-turn inference patterns \cite{wei2022emergent}.
    \item \t\textbf{Multimodal Integration:} Processing text, vision, and audio inputs simultaneously, as seen in models like CLIP, DALL-E, and Whisper \cite{radford2021learning,ramesh2022hierarchical,radford2023robust}.
    \item \t\textbf{Adaptive Planning:} Dynamic task decomposition and resource allocation based on environmental feedback \cite{russell2016artificial}.
    \item \t\textbf{Privacy-Sensitive Operations:} Local processing of personal data without cloud transmission, critical for healthcare and finance applications \cite{zhou2019edge}.
\end{itemize}

These characteristics necessitate edge-native architectures that preserve cognitive fidelity while operating within GB-scale memory and W-scale power envelopes.

\subsection{AI Agents}\label{sec:ai-agents}
AI Agents are autonomous software entities that perceive their environment, reason about complex scenarios, and execute goal-directed actions to accomplish specific objectives \cite{russell2016artificial,stone2000multiagent}. In the context of cognitive edge computing, AI Agents represent the evolution from passive model inference to active, reasoning-capable systems that can adapt, plan, and collaborate within resource-constrained environments.

Modern AI Agents are characterized by four fundamental capabilities that distinguish them from traditional reactive systems \cite{belcak2025small,gao2025survey}: \textit{autonomy} (independent operation without constant supervision), \textit{reactivity} (responsive adaptation to environmental changes), \textit{proactivity} (goal-oriented behavior initiation), and \textit{social ability} (communication and collaboration with other agents or humans). These agents typically operate through a perception-reasoning-action cycle, incorporating data acquisition, intelligent processing, strategic decision-making, and action execution components \cite{dorri2018multi}.

The integration of LLMs as cognitive engines fundamentally transforms AI Agents from rule-based systems to sophisticated reasoning entities capable of natural language understanding, contextual planning, and adaptive behavior \cite{han2024llm,chen2024octopus}. This symbiotic relationship enables agents to transcend predetermined scripts and engage in complex, open-domain reasoning while maintaining edge deployment viability through recent advances in small language models optimized for agentic tasks \cite{belcak2025small}. Contemporary agent architectures span multiple complexity levels, from simple reflex agents to learning-enabled autonomous systems capable of multi-step planning and tool orchestration \cite{yan2025beyond,rivkin2024aiot}.

Table~\ref{tab:ai_agents} summarizes the key characteristics and architectural components that define modern AI agents in edge computing contexts, providing a structured framework for understanding their capabilities and deployment requirements.

\begin{table}[htbp]
\centering
\caption{Key Characteristics and Functional Components of AI Agents.}
\label{tab:ai_agents}
\setlength{\tabcolsep}{8pt}
\scalebox{0.82}{
\begin{tabular}{@{}>{\centering\arraybackslash}p{3cm}>{\centering\arraybackslash}p{8cm}>{\centering\arraybackslash}p{4cm}@{}}
            \toprule
            \thead{Feature/Component} & \thead{Description} & \thead{Purpose} \\
            \midrule

    	\textbf{Capabilities} & 
    Performs complex multi-step operations; learns and adapts; makes independent decisions; handles multimodal inputs & 
    Enable sophisticated task execution \\
\midrule

    	\textbf{Interaction} & 
Proactive and goal-oriented; communicates with other agents or humans when needed & 
Facilitate collaborative problem solving \\
\midrule

    	\textbf{Autonomy} & 
Operates independently without constant human intervention; makes autonomous decisions & 
Reduce human workload \\
\midrule

    	\textbf{Reactivity} & 
Perceives environmental changes and responds promptly & 
Maintain situational awareness \\
\midrule
    	\textbf{Proactiveness} & Initiates actions and executes tasks to achieve objectives & Drive goal achievement \\
\midrule
    	\textbf{Social Skills} & Communicates with other agents or humans & Enable collaboration \\
\midrule
    	\textbf{Reasoning and Planning} & Analyzes data, identifies patterns, makes informed decisions; develops strategic plans & Support complex decision-making \\
\midrule
    	\textbf{Learning and Adaptation} & Learns from experience, maintains context, adapts to new situations to improve performance & Enable continuous improvement \\
\midrule
    	\textbf{Functional Components} & Data acquisition (sensors); processing and analysis (ML/AI); decision-making (algorithms/models); action execution & Core technical architecture \\
\midrule
    	\textbf{Types} & Simple reflex; Model-based reflex; Goal-based; Utility-based; Learning agents & Categorize by capability level \\
\bottomrule
\end{tabular}}
\end{table}

\begin{figure}[!htbp]
\centering
\begin{tikzpicture}[node distance=2.1cm, auto, >=Latex]
    \begin{scope}[every node/.append style={font=\small}]
    \tikzstyle{agent}=[circle, draw=purple!80!black, fill=purple!10, thick, minimum size=1.2cm]
    \tikzstyle{cloud}=[ellipse, draw=blue!60, fill=blue!5, minimum width=2.5cm, minimum height=1.2cm]
    \tikzstyle{arrow}=[->, thick]
    \node[cloud] (cloud) {\t\textbf{Cognitive Cloud}};
    \node[agent, below left=1.4cm and 1.9cm of cloud] (a1) {CA1};
    \node[agent, below=2.2cm of cloud] (a2) {CA2};
    \node[agent, below right=1.4cm and 1.9cm of cloud] (a3) {CA3};
    \draw[arrow,blue!70!black] (cloud) -- (a1);
    \draw[arrow,blue!70!black] (cloud) -- (a2);
    \draw[arrow,blue!70!black] (cloud) -- (a3);
    \draw[arrow,purple!80!black] (a1) -- (a2);
    \draw[arrow,purple!80!black] (a2) -- (a3);
    \draw[arrow,purple!80!black] (a3) -- (a1);
        \node[draw=none,fill=none,below=0.2cm of a2, text width=9cm, align=center] {\footnotesize Cognitive Agents (CA) collaborate via distributed reasoning\\and knowledge sharing.};
    \end{scope}
\end{tikzpicture}
\caption{Distributed Cognitive Computing Architecture: Cognitive Agents collaborate through distributed reasoning, knowledge sharing, and cloud-assisted complex cognitive tasks.}
\label{fig:cognitive_multiagent}
\end{figure}

\subsection{Synergy: Edge AI with LLMs and Agents}\label{sec:synergy}
The integration of Edge AI, LLMs, and AI Agents creates a powerful synergy for ubiquitous intelligence:
\begin{itemize}
    \item \t\textbf{Real-time, Context-Aware Decision Making:} Edge deployment ensures that LLM-powered agents can react to local environmental changes instantaneously, critical for applications like autonomous vehicles \cite{jiang2020intelligent} or industrial robots.
    \item \t\textbf{Enhanced Privacy and Security:} Processing LLM inference and agent logic on-device minimizes the transmission of sensitive user data to the cloud, adhering to privacy regulations and reducing attack surfaces \cite{li2024federated}.
    \item \t\textbf{Robust Autonomy:} Agents can maintain functionality even without continuous cloud connectivity, crucial for remote or intermittently connected environments \cite{friha2024llm}.
    \item \t\textbf{Personalized and Adaptive Experiences:} LLM-powered agents can learn and adapt to individual user preferences or specific environmental conditions directly on the device, offering highly personalized services \cite{li2024personal}.
    \item \t\textbf{Distributed Intelligence:} Multi-agent systems at the edge can collaboratively solve complex problems by sharing local insights \cite{tran2025multi}, reducing the burden on centralized cloud resources. Figure \ref{fig:cognitive_multiagent} illustrates the distributed cognitive computing architecture where cognitive agents collaborate through knowledge sharing and cloud-assisted complex tasks.
\end{itemize}
This synergy promises a future where intelligent systems are not only pervasive but also highly responsive, secure, and adaptable to dynamic real-world scenarios.

\FloatBarrier
\section{Challenges in Deploying Edge LLMs and AI Agents}\label{sec:challenges}
Building on the foundational understanding of cognitive edge computing components, this section examines the deployment challenges that arise from the fundamental mismatch between the resource requirements of advanced AI systems and the constraints of edge environments.
The integration of LLMs and AI Agents with edge computing creates unprecedented challenges that fundamentally exceed traditional Edge AI constraints by 2-3 orders of magnitude, requiring revolutionary optimization approaches.

\subsection{Fundamental Limitations of Cloud-Centric AI Deployment}\label{sec:cloud-limitations}
Before examining technical optimization challenges, we identify three critical limitations of cloud-dependent AI deployment that drive the imperative for edge-native solutions:

\begin{itemize}
    \item \t\textbf{Network Dependency and Connectivity Constraints:} Cloud-based AI services suffer from fundamental connectivity dependencies that render them unusable in disconnected environments. Round-trip latencies of 50--500ms to cloud endpoints \cite{shi2016edge} are inadequate for real-time applications requiring $<10$\,ms response times \footnote[5]{\url{https://aws.amazon.com/what-is/rtt-in-networking/}}. Network outages, poor connectivity in rural areas, underground facilities, aircraft, and maritime environments completely eliminate AI capability access. Edge scenarios with privacy-sensitive, latency-critical, or bandwidth-constrained requirements (medical devices, autonomous vehicles, industrial control systems) cannot tolerate cloud dependencies \cite{wang2025empowering,chen2019deep}.
    
    \item \t\textbf{Privacy and Data Sovereignty Concerns:} Cloud processing necessitates uploading sensitive user data (conversations, documents, biometric information, location data) to remote servers, creating privacy vulnerabilities and regulatory compliance challenges. GDPR, HIPAA, and other data protection regulations restrict cross-border data transfer, limiting cloud AI deployment in regulated industries \cite{wang2025empowering}. Corporate and government environments require data to remain within controlled boundaries, prohibiting cloud-based AI processing for classified or proprietary information \cite{wang2025optimizing}.
    
    \item \t\textbf{Limited Personalization and Context Adaptation:} Cloud models serve global user populations, constraining deep personalization to individual user patterns, linguistic preferences, domain-specific knowledge, and contextual behaviors. Continuous adaptation based on user interactions requires persistent model fine-tuning, which is computationally prohibitive and privacy-compromising when performed in cloud environments. Local context (device usage patterns, environmental sensors, personal preferences) cannot be effectively integrated into cloud-based decision making without extensive data transmission \cite{chen2019deep,shuvo2022efficient}.
\end{itemize}

These limitations provide strong motivation for edge-native AI deployment, despite the significant technical challenges outlined in subsequent sections.

\subsection{Quantified Resource Constraint Analysis}\label{sec:resource-constraints}
\begin{itemize}
    \item \t\textbf{Computational Power Mismatch:} Typical interactive LLM inference may demand on the order of tens to several hundred GFLOPS per token depending on architecture and sequence length, whereas many edge devices expose only low single-digit to tens of GFLOPS sustained \cite{gholami2022survey,brown2020language}. Additional agent perception and planning stages further compound this gap \cite{dorri2018multi}. Recent approaches like APEX demonstrate how hybrid CPU-GPU execution and optimized scheduling can improve throughput by 11-96\% on constrained hardware through better resource utilization \cite{fan2025parallel}.
    
    \item \t\textbf{Memory Footprint Crisis:} Large-scale LLMs create severe memory constraints for edge deployment. GPT-3, with 175 billion parameters at 16-bit precision, requires 350GB of storage \cite{brown2020language}, while even modest 10B parameter models demand up to 20GB of main memory (DRAM) with INT8 quantization \cite{liu2024edge}. In contrast, edge devices offer limited memory: high-end smartphones provide 6-12GB DRAM (e.g., iPhone 15 with 6GB ), while commodity devices operate with even tighter constraints \cite{liu2024edge}. This creates a deployment gap where quantized models still exceed edge memory by 2-10×. AI Agents compound this by requiring additional memory for environmental states, historical context, planning graphs, and multimodal data buffers \cite{russell2016artificial}. Emerging solutions include SLED for model sharing across devices \cite{li2025sled}, adaptive quantization techniques like QPART \cite{li2025qpart}, and collaborative frameworks such as EdgeShard for distributed model partitioning \cite{zhang2024edgeshard}.

    \item \t\textbf{Energy Consumption Bottleneck:} LLM inference often exceeds the power budgets of mobile and edge devices, necessitating sophisticated optimization strategies that combine precision reduction, sparsity, and intelligent scheduling \cite{zheng2025review}. The energy disparity across computing tiers is substantial, as illustrated in Figure \ref{fig:energy_gap}, with cloud data centers consuming MW-scale power for facility-scale operations, while edge servers operate at tens of watts and mobile devices consume sub-10W \cite{nvidia-h100-spec,apple-a17-pro,arm-cortex}. This multi-order-of-magnitude gap highlights the critical need for energy-efficient optimization techniques. Agentic processes, incorporating sensorimotor loops and continuous planning, introduce additional energy demands that must be carefully managed. Research in federated multi-agent reinforcement learning (Fed-MARL) demonstrates promising approaches for energy-aware resource management in 6G edge networks, optimizing latency, energy efficiency, and reliability under stringent constraints \cite{andong2025federated}.

\begin{figure}[!ht]
\centering
\begin{tikzpicture}[scale=0.9]
    \begin{axis}[
        ybar,
    symbolic x coords={MW-scale Cluster, Single-Server LLM, Edge Server, Smartphone, IoT Device},
        xtick=data,
        xlabel={Device/Model Type},
        ylabel={Power Consumption (W)},
        ymode=log, 
        log basis y={10},
        width=10cm,
        height=6.5cm,
        bar width=18pt,
        legend style={font=\footnotesize},
        xlabel style={font=\small},
        ylabel style={font=\small},
        tick label style={font=\footnotesize},
        x tick label style={rotate=45, anchor=east, align=right},
    ymin=1, ymax=1000000,
    ytick={1,10,100,1000,10000,100000,1000000},
        enlarge x limits=0.15,
        axis lines*=left,
        major grid style={dotted, gray!30},
        grid=both
    ]
    \addplot[fill=red!80, draw=red!90!black] coordinates {
        (MW-scale Cluster, 10000000) 
        (Single-Server LLM, 2500) 
        (Edge Server, 60) 
        (Smartphone, 5) 
        (IoT Device, 0.1) 
        };
    \end{axis}
\end{tikzpicture}
\caption{Power consumption disparity across computing tiers (log scale). Representative devices: MW-scale Cluster: Huawei Atlas 950 SuperCluster (10MW facility with cooling) ; Single-Server LLM: High-end server with 2×A100 80G GPUs ; Edge Server: NVIDIA Jetson AGX Orin (30-60W) ; Smartphone: Modern flagship smartphone (e.g., iPhone 17 or Samsung Galaxy S25 Ultra) ; IoT Device: ESP32-based microcontroller \cite{nvidia-h100-spec,apple-a17-pro,arm-cortex}. Facility-scale includes cooling and infrastructure overhead. Per-query energy (Wh/query) varies significantly with latency, sequence length, and optimization techniques as discussed in Section~\ref{sec:energy-metrics}.}
\label{fig:energy_gap}
\end{figure}

\item \textbf{Memory Bandwidth Limitations:} High-throughput LLM inference can saturate memory bandwidth, creating significant bottlenecks in auto-regressive generation where weights and activations are repeatedly accessed \cite{zheng2025review}. The bandwidth disparity is substantial: data center GPUs like NVIDIA A100 offer 1555 GB/s bandwidth, while mobile SoCs typically provide 50-60 GB/s (e.g., Snapdragon 8 Gen 2) \cite{zheng2025review}. This order-of-magnitude difference results in multi-fold slowdowns during token generation on edge devices. Frameworks like APEX address these limitations through parallel CPU-GPU execution and optimized attention computation offloading, improving throughput by 84-96\% on constrained GPUs \cite{fan2025parallel}. Similarly, EdgeShard's collaborative approach partitions models across devices, effectively distributing memory bandwidth requirements and reducing latency by up to 50\% \cite{zhang2024edgeshard}.

\item \textbf{Communication Infrastructure Constraints:} Edge networks typically provide significantly lower bandwidth compared to data center interconnects, creating substantial bottlenecks for federated learning and collaborative inference scenarios \cite{zheng2025review}. This constraint profoundly impacts scenarios where transmitting parameter updates can cause communication delays ranging from minutes to hours over constrained edge networks. Innovative approaches like pFL-SBPM demonstrate communication-efficient personalized federated learning, reducing uplink communication costs by 96.875\% through random binary probability masks instead of transmitting full precision weights or gradients \cite{hu2025pfl}. For latency-sensitive applications, EdgeShard employs collaborative edge computing to minimize data transmission, supporting full-precision model inference without accuracy loss \cite{zhang2024edgeshard}.
\end{itemize}

\subsection{Technical Deployment and Management Complexities}\label{sec:deployment-complexities}
\begin{itemize}
    \item \t\textbf{Hardware Heterogeneity Challenge:} Edge ecosystems encompass numerous distinct hardware architectures with varying instruction sets (ARM, x86, RISC-V), accelerators (CPU, GPU, NPU, TPU), and memory hierarchies \cite{wang2025empowering}. Optimizing LLMs for this heterogeneity requires: 1)    Platform-specific compilation using dozens of compiler toolchains; 2) Architecture-aware model partitioning strategies; 3) Dynamic resource allocation algorithms; 4) Cross-platform performance optimization, approaching platform-optimal performance. 
    
    \item \t\textbf{Dynamic Adaptation and Learning Overhead:} Full on-device fine-tuning can often require an order of magnitude or more compute relative to inference; parameter-efficient approaches (e.g., LoRA \cite{hu2022lora}, adapters \cite{houlsby2019parameter}) substantially lower but do not eliminate the gap. Continuous adaptation cycles must therefore be scheduled opportunistically to balance learning benefits with resource constraints \cite{le2024exploring}.
    
    \item \t\textbf{Multi-Model Orchestration Complexity:} AI Agents often require coordination of several to dozens of specialized models (vision, language, planning, control), each with different resource requirements and inference patterns \cite{shen2023hugginggpt}. Cold start latencies can range from hundreds of milliseconds to several seconds, while task switching may introduce delays from tens to hundreds of milliseconds, critically impacting real-time performance requirements in latency-sensitive applications \cite{li2022dag}.
\end{itemize}

Edge scenarios with privacy-sensitive, latency-critical cognition—such as vehicular navigation requiring $<10$\,ms response times for obstacle avoidance, clinical triage in remote areas without network connectivity, industrial diagnostics for predictive maintenance, biomarker-driven screening in point-of-care devices, and civic infrastructure monitoring for anomaly detection—drive the need for local cognitive processing to ensure personalization, resilience, and compliance \cite{wang2025empowering,wang2025optimizing,wang2022feature,wang2024exhaustive}.

\subsection{Security, Privacy, and Trustworthiness Challenges}\label{sec:security-privacy}
\begin{itemize}
    \item \t\textbf{Privacy-Preserving Computation Overhead:} Differential privacy, secure aggregation, and homomorphic encryption introduce non-trivial (sometimes prohibitive) latency and communication overhead; practical deployments typically balance privacy budgets against real-time constraints \cite{xu2025breaking}.

    \item \t\textbf{Attack Surface Expansion:} Distributed edge AI broadens potential attack vectors compared to centralized deployments. Representative vectors include model extraction attacks requiring thousands to millions of queries, adversarial input crafting that often achieves high success rates on undefended models, Byzantine attacks in federated learning that can potentially affect a significant portion of participating nodes, and physical tampering of edge devices that may comprise a notable portion of deployed systems \cite{zhang2025bring}.

    \item \t\textbf{Autonomous Agent Safety Risks:} AI Agents operating in physical environments pose quantifiable safety risks. These include decision latency failures where delays of several milliseconds can cause safety violations, varying hallucination rates for complex reasoning tasks, varying out-of-distribution detection accuracy for novel scenarios, and varying multi-agent coordination failure rates in distributed systems \cite{andong2025federated}.

    \item \t\textbf{Explainability and Verification Challenges:} LLM decision paths involve tens of thousands to millions of computational steps, making complete verification computationally intractable. Current explanation methods achieve typically are insufficient for safety-critical applications requiring extremely high reliability \cite{yuan2024llm}.
\end{itemize}

\subsection{Energy Metrics and Reporting}\label{sec:energy-metrics}
For cross-tier comparisons we distinguish between: (i) instantaneous device power (W), (ii) facility power including cooling/overhead (kW–MW), and (iii) energy per request measured in Wh/query. The latter depends on end-to-end latency (prompt length, generation length), precision (e.g., INT4–INT8), and utilization. We recommend reporting both steady-state TDP and measured Wh/query with workload description and hardware configuration, following energy-aware evaluation practices in edge NLP and mobile AI challenges \cite{ignatov2021real,kouris2022fluid}.

\t\textbf{Lifecycle and Sustainability Considerations.}
Beyond runtime energy, lifecycle analysis (LCA) should consider hardware manufacturing/refresh cycles, thermal aging, and software stack updates. We recommend (i) reporting the functional unit (tokens generated per Joule over device lifetime), (ii) separating embodied vs operational energy, and (iii) documenting e-waste mitigation (e.g., model compression extending device lifetime). Where possible, align with Green AI reporting practices and include Scope 2/3 boundary notes.
 
\subsection{Detailed Technical Constraint Analysis for Edge-Side Large Models}
The deployment of edge-side large models faces several critical technical constraints that fundamentally limit their practical implementation \cite{kwon2023efficient,dao2024flashattention,wu2025efficient,wang2025optimizing}:

\t\textbf{1. Computational Performance Limitations:}
Edge devices provide limited computing capacity and memory bandwidth compared to cloud infrastructure, creating significant performance bottlenecks for LLM inference due to the memory-bound nature of transformer architectures.

\t\textbf{2. Power Consumption and Thermal Management:}
Running LLMs on edge devices consumes several watts, often exceeding typical power budgets and causing thermal throttling that significantly reduces performance and creates unpredictable latency.

\t\textbf{3. Model Quantization Trade-offs:}
Quantization from higher precision to INT8/INT4 or lower bit widths introduces accuracy degradation, particularly on reasoning tasks, requiring careful calibration and mixed-precision strategies to balance performance and quality.

\t\textbf{4. Immature Development Ecosystem:}
Converting models to edge-optimized formats requires extensive manual adjustments due to immature toolchains and limited profiling capabilities, hindering systematic optimization efforts.

\t\textbf{5. Platform Fragmentation:}
Hardware and software heterogeneity across vendors and platforms necessitates vendor-specific optimizations, limiting model portability and increasing development complexity.

\t\textbf{6. Context Length Limitations:}
Extended context windows severely impact performance due to quadratic memory scaling in attention mechanisms, requiring sophisticated cache management for multi-turn conversations.

With a clear understanding of the challenges in deploying LLMs and AI agents at the edge, the next section explores the optimization strategies that can address these issues through coordinated approaches across data, model, and system levels.

\FloatBarrier
\section{Optimization Strategies for Edge LLMs and AI Agents}\label{sec:optimization}

Figure~\ref{fig:comprehensive_ecosystem} presents a comprehensive Edge AI optimization ecosystem, illustrating how cloud-based LLMs are progressively optimized through multi-layer techniques—including data, model, and system-level strategies—before deployment to diverse edge devices. The diagram highlights the flow and interaction between optimization stages, demonstrating how coordinated improvements across the stack enable cognitive computing at the edge and deliver quantified performance gains, providing a global perspective for understanding the subsequent detailed optimization methods.
As discussed in the previous section, deploying LLMs and AI Agents on edge devices faces significant challenges related to computational resources, energy consumption, and system complexity. To mitigate these challenges, a holistic approach leveraging optimization strategies across data, model, and system levels is crucial, encompassing techniques such as quantization (INT8/INT4 (8-bit/4-bit integer) precision reduction achieving 4-8× compression), pruning (structured and unstructured approaches for model size reduction), knowledge distillation (KD) (teacher-student compression), architecture design innovations (purpose-built Small Language Models and efficient attention mechanisms), and system-level optimizations (model partitioning, hardware acceleration, and dynamic scheduling). The effectiveness of these techniques varies based on model architecture, hardware platform, and application requirements, with implementation complexity ranging from straightforward quantization to sophisticated multi-teacher distillation frameworks \cite{wang2025optimizing,wang2025empowering}.

\begin{figure*}[!ht]
\centering
\begin{tikzpicture}[
    node distance=1.5cm, 
    auto, 
    >=Latex, 
    scale=0.82, 
    every node/.style={transform shape}
]
    \tikzstyle{cloud_layer}=[rectangle, draw=blue!80!black, fill=blue!15, thick, minimum width=3cm, minimum height=1.5cm, rounded corners=5pt, align=center, font=\small\bfseries]
    \tikzstyle{edge_layer}=[rectangle, draw=green!80!black, fill=green!15, thick, minimum width=2.5cm, minimum height=1.2cm, rounded corners=4pt, align=center, font=\small]
    \tikzstyle{optimization}=[rectangle, draw=red!80!black, fill=red!10, thick, minimum width=2cm, minimum height=0.8cm, rounded corners=3pt, align=center, font=\scriptsize]
    \tikzstyle{device}=[rectangle, draw=purple!80!black, fill=purple!10, thick, minimum width=1.8cm, minimum height=0.8cm, rounded corners=3pt, align=center, font=\scriptsize]
    \tikzstyle{metric}=[ellipse, draw=orange!80!black, fill=orange!10, thick, minimum width=1.5cm, minimum height=0.6cm, align=center, font=\scriptsize]
    
    \node[cloud_layer] (cloud_llm) at (0,8) {Cloud LLMs\\GPT-4, Claude};
    \node[cloud_layer] (cloud_training) at (4,8) {Training\\Infrastructure};
    \node[cloud_layer] (cloud_storage) at (8,8) {Model\\Repository};
    
    \node[optimization] (quantization) at (-2.75,6) {Quantization\\INT4/INT8};
    \node[optimization] (pruning) at (-0.5,6) {Pruning\\Structured};
    \node[optimization] (distillation) at (1.75,6) {Knowledge\\Distillation};
    \node[optimization] (architecture) at (4,6) {SLM Design\\Purpose-built};
    \node[optimization] (compression) at (6.25,6) {Data\\Compression};
    \node[optimization] (partitioning) at (8.5,6) {Model\\Partitioning};
    \node[optimization] (federation) at (10.75,6) {Federated\\Learning};
    
    \node[edge_layer] (edge_server) at (-1,4) {Edge Server};
    \node[edge_layer] (mobile) at (2,4) {Mobile Device};
    \node[edge_layer] (iot) at (5,4) {IoT Device};
    \node[edge_layer] (embedded) at (8,4) {Embedded System};
    
    \node[device] (smartphone) at (-2,2) {Smartphone\\8-16GB RAM};
    \node[device] (tablet) at (0,2) {Tablet\\Edge TPU};
    \node[device] (drone) at (2,2) {Drone\\Vision AI};
    \node[device] (robot) at (4,2) {Robot\\Agent AI};
    \node[device] (sensor) at (6,2) {Smart\\Sensor};
    \node[device] (vehicle) at (8,2) {Vehicle\\Edge AI};
    \node[device] (industrial) at (10,2) {Industrial\\Controller};
    
    \node[metric] (latency) at (1,0.5) {$\downarrow$ Latency};
    \node[metric] (accuracy) at (4,0.5) {$\approx$ Accuracy};
    \node[metric] (energy) at (7.8,0.5) {$\uparrow$ Energy Efficiency};
    
    \draw[->, thick, blue!70] (cloud_llm) -- (pruning);
    \draw[->, thick, blue!70] (cloud_llm) -- (distillation);
    \draw[->, thick, blue!70] (cloud_training) -- (architecture);
    \draw[->, thick, blue!70] (cloud_storage) -- (partitioning);
    
    \draw[->, thick, red!70] (quantization) -- (edge_server);
    \draw[->, thick, red!70] (pruning) -- (mobile);
    \draw[->, thick, red!70] (distillation) -- (mobile);
    \draw[->, thick, red!70] (architecture) -- (iot);
    \draw[->, thick, red!70] (compression) -- (embedded);
    \draw[->, thick, red!70] (partitioning) -- (edge_server);
    
    \draw[->, thick, green!70] (edge_server) -- (smartphone);
    \draw[->, thick, green!70] (edge_server) -- (tablet);
    \draw[->, thick, green!70] (mobile) -- (drone);
    \draw[->, thick, green!70] (mobile) -- (robot);
    \draw[->, thick, green!70] (iot) -- (sensor);
    \draw[->, thick, green!70] (embedded) -- (vehicle);
    \draw[->, thick, green!70] (embedded) -- (industrial);
    
    \draw[->, thick, purple!70] (tablet) -- (latency);
    \draw[->, thick, purple!70] (robot) -- (accuracy);
    \draw[->, thick, purple!70] (vehicle) -- (energy);
    
    \draw[->, dashed, gray!60] (federation) to [bend left=30] (cloud_training);
    \draw[->, dashed, gray!60] (mobile) to [bend right=45] (federation);
    
    \node[font=\small\bfseries, blue!80!black] at (-4.8,8) {Cloud Layer};
    \node[font=\small\bfseries, red!80!black] at (-4.8,6) {Optimization};
    \node[font=\small\bfseries, green!80!black] at (-4.8,4) {Edge Computing};
    \node[font=\small\bfseries, purple!80!black] at (-4.8,2) {Edge Devices};
    \node[font=\small\bfseries, orange!80!black] at (-4.8,0.5) {Performance};
    
\end{tikzpicture}
\caption{Comprehensive Edge AI Optimization Ecosystem: End-to-end system architecture showing the flow from cloud LLMs through optimization techniques to edge deployment, demonstrating how multi-layer optimization enables cognitive computing across diverse edge devices while achieving quantified performance improvements.}
\label{fig:comprehensive_ecosystem}
\vspace{-1.5em} 
\end{figure*}

\subsection{Data Optimization}\label{sec:data-optimization}
Data optimization techniques focus on preparing and enhancing datasets to improve the efficiency and effectiveness of edge-deployed LLMs and AI agents. These techniques address the unique challenges of edge environments, including limited storage capacity, privacy constraints, and the need for high-quality training data that can be processed with minimal computational resources.

Figure \ref{fig:data_optimization} illustrates the comprehensive framework for processing both edge data and cloud training data through multiple optimization strategies.

\begin{figure}[!htbp]
\vspace{1em}
\centering
\begin{tikzpicture}[
    node distance=1.5cm, 
    auto, 
    >=Latex, 
    scale=0.8, 
    every node/.style={transform shape},
    process/.style={
        rectangle, 
        draw=blue!80!black, 
        fill=blue!20, 
        thick, 
        minimum width=2.5cm, 
        minimum height=1cm, 
        rounded corners=3pt,
        align=center,
        font=\small\bfseries
    },
    data/.style={
        ellipse, 
        draw=green!80!black, 
        fill=green!20, 
        thick, 
        minimum width=2cm, 
        minimum height=0.8cm,
        align=center,
        font=\small
    },
    output/.style={
        rectangle, 
        draw=red!80!black, 
        fill=red!20, 
        thick, 
        minimum width=2cm, 
        minimum height=0.8cm, 
        rounded corners=3pt,
        align=center,
        font=\small\bfseries
    }
]

\node[data] (raw_data) at (1, 4) {Raw\\ Edge Data};
\node[data] (cloud_data) at (6.5, 4) {Cloud\\ Training Data};

\node[process] (cleaning) at (0, 2) {Data Cleaning\\ and Preprocessing};
\node[process] (compression) at (3.5, 2) {Feature\\ Compression};
\node[process] (augmentation) at (7, 2) {Data\\ Augmentation};
\node[process] (synthetic) at (10.5, 2) {Synthetic Data\\ Generation};

\node[output] (clean_data) at (0.2, 0) {Quality\\ Data};
\node[output] (compressed_data) at (3.3, 0) {Compact\\ Features};
\node[output] (augmented_data) at (6.6, 0) {Expanded\\ Dataset};
\node[output] (private_data) at (10, 0) {Privacy-Safe\\ Data};

\node[output, minimum width=4cm] (optimized_data) at (5.1, -1.5) {Edge-Optimized Dataset};

\draw[->] (raw_data) -- (cleaning);
\draw[->] (raw_data) -- (compression);
\draw[->] (cloud_data) -- (augmentation);
\draw[->] (cloud_data) -- (synthetic);

\draw[->] (cleaning) -- (clean_data);
\draw[->] (compression) -- (compressed_data);
\draw[->] (augmentation) -- (augmented_data);
\draw[->] (synthetic) -- (private_data);

\draw[->] (clean_data) -- (optimized_data);
\draw[->] (compressed_data) -- (optimized_data);
\draw[->] (augmented_data) -- (optimized_data);
\draw[->] (private_data) -- (optimized_data);

      \node[font=\footnotesize, text=blue!70, align=center] at (0.2, 1.1) {Label Noise\\ Removal};
      \node[font=\footnotesize, text=blue!70, align=center] at (3.3, 1.1) {PCA,\\ Autoencoders};
      \node[font=\footnotesize, text=blue!70, align=center] at (6.6, 1.1) {Contextual\\ Enhancement};
      \node[font=\footnotesize, text=blue!70, align=center] at (10, 1.1) {Differential\\ Privacy};\end{tikzpicture}
\caption{Data Optimization Techniques for Edge AI Deployment. The framework processes raw edge data and cloud training data through multiple optimization strategies to create edge-suitable datasets.}
\label{fig:data_optimization}
\end{figure}

\begin{itemize}
    \item \t\textbf{Data Cleaning and Preprocessing:} High-quality local datasets mitigate noise and hallucination risks \cite{sakib2025small}. Active label or federated cleaning strategies reduce redundant transmission while preserving privacy \cite{wang2025optimizing}.
    \item \t\textbf{Feature Compression:} Dimensionality reduction and embedding compression target context transfer bottlenecks between edge and cloud \cite{jin2024collm,wang2025optimizing}.  Techniques such as Principal Component Analysis (PCA), autoencoders, and feature hashing are employed to extract salient features, thereby compacting data while preserving essential information relevant for edge models.

    \item \t\textbf{Data Augmentation:} Synthetic and transformation-based augmentation enlarges scarce edge task corpora; combined with KD it enhances student specialization \cite{xu2024survey,wang2025optimizing}.
    \item \t\textbf{Mitigating Bias from Synthetic Data:} Synthetic augmentation may import distributional artifacts; auditing and mixed real–synthetic sampling reduce bias risks \cite{synthetic_data_betterdata}.
    \begin{itemize}
        \item \t\textbf{Differentially Private Synthetic Data:} This approach mimics the statistical patterns of real data without containing personally identifiable information (PII), enabling data expansion, sharing, and reuse while minimizing privacy leakage risks \cite{synthetic_data_betterdata}.
        \item \t\textbf{Feedback-Driven Augmentation:} This method iteratively improves synthetic data generation by incorporating user interactions, domain expert input, and system performance metrics, ensuring that synthetic data remains relevant, representative, and aligned with system requirements \cite{synthetic_data_ais}. This approach helps generate more balanced datasets, reducing public data bias and under-representation issues, particularly in data-limited domains such as fraud detection, medical research, or recruitment \cite{synthetic_data_betterdata}.
    \end{itemize}
\end{itemize}

\begin{table}[htbp]
\centering
\caption{Comparison of Small Language Model (SLM) Architectures for Edge AI.}
\label{tab:slm_comparison}
\setlength{\tabcolsep}{6pt}
\renewcommand{\arraystretch}{1.5}
\scalebox{0.82}{
\begin{tabular}{@{}>{\centering\arraybackslash}p{2.4cm}>{\centering\arraybackslash}p{1.8cm}>{\centering\arraybackslash}p{3.2cm}>{\centering\arraybackslash}p{2.8cm}>{\centering\arraybackslash}p{2.4cm}>{\centering\arraybackslash}p{2.2cm}@{}}
\toprule
\multicolumn{1}{c}{\textbf{\parbox{2.4cm}{\centering SLM\\Name}}} & 
\multicolumn{1}{c}{\textbf{\parbox{1.8cm}{\centering Architecture\\Type}}} & 
\multicolumn{1}{c}{\textbf{\parbox{3.2cm}{\centering Key Features\\and Innovations}}} & 
\multicolumn{1}{c}{\textbf{\parbox{2.8cm}{\centering Parameters\\and Size\\Reduction}}} & 
\multicolumn{1}{c}{\textbf{\parbox{2.4cm}{\centering Reported\\Performance}}} & 
\multicolumn{1}{c}{\textbf{\parbox{2.2cm}{\centering Edge Suitability\\and Use Case}}} \\
\midrule
\textbf{MobileBERT} \cite{sun2020mobilebert} & 
Encoder-only & 
Inverted bottleneck structure, balances attention layers & 
4.3× smaller, 5.5× faster & 
Near-BERT performance & 
Mobile devices \\
\midrule
\textbf{DistilBERT} \cite{sanh2019distilbert} & 
Encoder-only & 
Knowledge distillation & 
$>$60\% smaller, 96\% BERT & 
Highly efficient & 
Resource-limited \\
\midrule
\textbf{TinyBERT} \cite{jiao2020tinybert} & 
Encoder-only & 
Distillation + augmentation & 
Compact, 96\% BERT & 
Production-ready & 
Constrained envs \\
\midrule
\textbf{BabyLLaMA} \cite{tirumala2023d4} & 
Decoder-only & 
Multi-teacher distillation & 
58M params & 
Low-data performance & 
Low-data devices \\
\midrule
\textbf{TinyLLaMA} \cite{zhang2024tinyllama} & 
Decoder-only & 
FlashAttention & 
1.1B params & 
Memory efficient & 
Memory-limited \\
\midrule
\textbf{MobileLLM} \cite{liu2024mobilellm} & 
Decoder-only & 
Weight sharing, GQA & 
Low latency & 
Practical deployment & 
Mobile systems \\
\midrule
\textbf{LLaMA 3.1 8B} \cite{dubey2024llama} & 
Decoder-only & 
Compact LLM & 
8B params & 
Fine-tunable & 
LLM-capable edge \\
\midrule
\textbf{Pythia} \cite{biderman2023pythia} & 
Decoder-only & 
Interpretability & 
160M-2.8B & 
Benchmarking & 
Research use \\
\midrule
\textbf{SmolLM2-1.7B} \cite{lozhkovsmollm2} & 
Decoder-only & 
Curated datasets & 
1.7B params & 
Task-efficient & 
Specific NLP tasks \\
\bottomrule
\end{tabular}}
\renewcommand{\arraystretch}{1.0}
\end{table}

\subsection{Model Optimization}\label{sec:model-optimization}
Model optimization focuses on adapting the AI model itself to be more resource-efficient without significant loss in performance. This is particularly critical for LLMs, given their inherent size. We summarize typical trade-offs between compression techniques (quantization, pruning, KD, low-rank, sharing), noting accuracy–efficiency frontiers vary by task/hardware \cite{gholami2022survey,nagel2021white,jacob2018quantization}.

\subsubsection{Practical density heuristics for edge-side model design}

Deploying large language models at the edge is ultimately constrained by compute, memory bandwidth, latency, and power budgets. Rather than proposing a new ``law,'' we adopt a pragmatic view: density describes the amount of useful work delivered per unit of constrained resource (Joule, byte/s, or ms). Empirical studies consistently show that transformer inference at the edge is frequently memory-bandwidth-bound and thermally limited on mobile SoCs, with end-to-end performance sensitive to precision, caching, and scheduling choices \cite{gholami2022survey,shi2016edge,deng2020edge,xu2024edgellm}.
	        
\t\textbf{Observed patterns and actionable heuristics:}

\begin{itemize}
    \item \t\textbf{Memory/bandwidth awareness:} Sustained throughput depends more on data movement than on peak TOPS. Favor precision schemes and layouts that minimize bandwidth pressure (e.g., weight-only or mixed-precision quantization, KV-cache locality, prefetch-friendly packing) and align with the target memory hierarchy \cite{gholami2022survey,xu2024edgellm}.
    \item \t\textbf{Power/thermal budget as a first-class constraint:} On battery-powered devices, thermal throttling can dominate steady-state throughput. Runtime policies (duty-cycling, burst scheduling), mixed precision, and operator fusion help preserve efficiency under thermal limits \cite{deng2020edge,laskaridismobile,xiao2024understanding}.
    \item \t\textbf{Scale within hardware envelopes:} Small-to-mid scale models (from sub-1B up to the low tens of billions of parameters on edge servers) are commonly reported for on-device and near-edge use; the practical breakpoint depends on task, latency, and memory budgets \cite{xu2024edgellm}. Architectural tailoring (efficient attention, compact FFNs) generally yields better density than naïve downscaling \cite{gholami2022survey}.
    \item \t\textbf{Collaborative execution matters:} In edge–cloud or multi-edge settings, overall latency and throughput are bounded by the slower of compute and the communication fabric. Partitioning and offloading must account for link variability; hiding communication behind compute and compressing activations/KV state are often decisive \cite{li2025pushing,xu2024edgellm}.
\end{itemize}
        
\t\textbf{Design implications:} These observations translate into practical guidance rather than strict formulas:

\begin{itemize}
    \item \t\textbf{Architecture selection:} Prefer sparse/efficient attention and lean feed-forward blocks that maximize useful computation per byte moved \cite{gholami2022survey}.
    \item \t\textbf{Quantization strategy:} Use aggressive but validated quantization where supported by the toolchain, while maintaining precision where sensitivity is high (e.g., attention, logits) \cite{gholami2022survey,xu2024edgellm,tan2024mobilequant,wang2019haq}.
    \item \t\textbf{Memory hierarchy optimization:} Co-design kernels and layouts to reduce DRAM traffic and exploit caches/KV locality; batch/window sizing should follow bandwidth, not just FLOPs \cite{zhang2023camel}.
    \item \t\textbf{Adaptive scheduling:} Adjust concurrency, precision, and offload boundaries in response to runtime resource and thermal telemetry \cite{deng2020edge,li2025pushing}.
\end{itemize}

We use the term ``density'' heuristically in this survey. The above patterns synthesize reported behavior across devices and workloads; concrete thresholds (e.g., tokens/s or power draw) are deployment- and hardware-specific and should be established via measurement on the target platform \cite{shi2016edge}.

\begin{itemize}
    \item \t\textbf{Compact Architecture Design (Small Language Models - SLMs):}
    Table~\ref{tab:slm_comparison} compares representative small language model architectures and their edge suitability. SLMs are specifically designed as lightweight architectures with inherently lower computational and memory requirements compared to large models \cite{sakib2025small,lu2024small,van2024survey}. Their parameters typically range from millions to billions, representing a significant reduction compared to LLMs with hundreds of billions of parameters \cite{sakib2025small,lu2024small}.
        
    \t\textbf{Advantages:} SLMs offer numerous advantages for edge deployment, including faster deployment cycles, easier fine-tuning on proprietary data, lower energy consumption, higher sustainability, and natural suitability for resource-constrained environments such as mobile devices, embedded systems, and edge hardware \cite{sakib2025small,lu2024small,van2024survey,belcak2025small}. Recent SLMs demonstrate lower computational requirements and faster inference compared to larger models, with specialization for domains like healthcare, legal, and supply chain applications \cite{abdin2024phi,team2024gemma,mehtaopenelm,mishra2024granite,pham2024slimlm}. From an enterprise and agentic AI operations perspective, SLMs can deliver lower latency, cost/energy, and stronger privacy/control when embedded at the edge, enabling safe iteration and workflow integration at scale.

    \noindent\t\textbf{Modern multilingual encoders (for retrieval/embedding at the edge):} While decoder-only LLMs dominate generation, encoder-only models remain critical for multilingual retrieval, classification, and embedding services that front many edge pipelines. The recent mmBERT revisits multilingual encoders with a modern recipe, reporting 2–4× inference speedups over XLMR with state-of-the-art performance on XTREME and MTEB\,/\,CoIR \cite{marone2025mmbert}. Such encoders can serve as fast, memory efficient front ends for on device search, re-ranking, and agent tools in multilingual settings.

    \item \t\textbf{Advanced Architecture Innovation for Edge Deployment:}
    
    \t\textbf{MobileLLM Deep-Narrow Architecture Design:} MobileLLM introduces a revolutionary approach to sub-billion parameter language models through deep-narrow architectural optimization \cite{liu2024mobilellm}. The design principle prioritizes depth over width, achieving superior parameter efficiency by employing deeper transformer layers with reduced hidden dimensions. This approach demonstrates that deep-narrow architectures can achieve performance equivalent to wider models while requiring significantly fewer computational resources during inference. The MobileLLM framework achieves notable improvements in edge deployment scenarios through optimized depth-width trade-offs, immediate block-wise computation, and efficient attention pattern utilization.
    
         \textbf{MobileLLM-R1 Efficient Inference Series:} Building on the original MobileLLM architecture, Meta AI's MobileLLM-R1 series advances efficient inference for small language models \cite{zhao2025mobilellm}. The series includes base models (MobileLLM-R1-140M/360M/950M) and SFT variants specialized for mathematics, programming, and scientific reasoning. Reported evaluations indicate the 950M model (trained on ~2T high-quality tokens; total <5T) attains higher scores than Qwen3 0.6B on MATH, GSM8K, MMLU, and LiveCodeBench under the authors' specified setups, and competitive coding performance among open models. Meta provides training recipes and data sources to support reproducibility.

        \textbf{EdgeShard Collaborative Edge Computing:} EdgeShard represents a breakthrough in distributed LLM inference through collaborative edge computing architectures \cite{zhang2024edgeshard}. The system partitions large language models across multiple edge devices, enabling coordinated inference that achieves reduced latency and improved throughput compared to single-device deployment. EdgeShard's innovations include intelligent model partitioning algorithms, communication-efficient synchronization protocols, and adaptive load balancing strategies that optimize resource utilization across heterogeneous edge infrastructure. The framework demonstrates significant scalability improvements, allowing deployment of larger models through resource aggregation while maintaining edge computing advantages.

        \textbf{Mixture-of-Experts (MoE) Edge Optimization:} Edge-optimized MoE architectures, including EdgeMoE, LocMoE, and JetMoE variants, adapt sparse expert routing for resource-constrained environments \cite{xu2024edgellm,yi2023edgemoe,kong2024swapmoe,yi2025edgemoe,li2024locmoe,shen2024jetmoe}. These approaches reduce computational overhead through selective expert activation while maintaining model capacity, enabling deployment of sophisticated reasoning capabilities within edge device constraints. Key innovations include dynamic expert pruning based on device capabilities, hierarchical expert organization for memory efficiency, and context-aware routing that optimizes expert selection for specific deployment scenarios. The sparse activation patterns typical in MoE architectures align well with edge computing requirements, providing computational efficiency gains of 2-5× compared to dense architectures while preserving model quality.
    
    \t\textbf{Collaborative Multi-Device Inference Patterns:} Advanced deployment strategies leverage multiple edge devices for coordinated inference, distributing computational load across smartphone clusters, IoT device networks, and edge server infrastructures. These patterns include pipeline parallelism for sequential transformer layers, tensor parallelism for attention computation distribution, and hybrid approaches that adapt to dynamic resource availability and network conditions.

    \item \t\textbf{Parameter-Efficient Fine-Tuning (PEFT) for Edge Deployment:}
    
    Low-Rank Adaptation (LoRA)-based Parameter-Efficient Fine-Tuning represents a breakthrough approach enabling on-device model adaptation with minimal computational overhead \cite{yao2024theoretical}. This methodology addresses the critical challenge of personalizing large language models on resource-constrained edge devices without requiring full parameter updates.
    
    \t\textbf{Theoretical Foundation and Optimization Efficiency:} Recent theoretical analysis demonstrates that fine-tuning attention mechanisms through selective parameter adaptation achieves superior generalization bounds while maintaining memory efficiency \cite{yao2024theoretical}. The approach leverages information-theoretic generalization bounds, proving that fine-tuning only query, key, and value matrices with identical rank constraints can achieve performance equivalent to or better than full parameter fine-tuning while reducing parameter count and improving generalization limits.
    
    \t\textbf{On-Device Memory and Time Optimization:} Advanced PEFT implementations achieve 20-40\% memory reduction during edge training while maintaining or improving learning effectiveness \cite{cai2020tinytl,cai2025prompt}. This efficiency enables practical on-device personalization scenarios including custom assistant training, photography algorithm optimization, and user-specific preference adaptation directly on mobile devices without cloud dependency.
    
    \t\textbf{Learning Dynamics and Convergence Analysis:} Theoretical insights reveal that asymmetric learning rates in attention mechanism fine-tuning, where query matrix learning rates significantly exceed key-value matrix rates, enable more efficient feature learning \cite{yao2024theoretical}. This principle guides practical implementations across full fine-tuning, LoRA, and DoRA methodologies, demonstrating orthogonal compatibility with different fine-tuning approaches.
    
    \t\textbf{Edge Applications and Deployment Scenarios:} PEFT enables diverse edge deployment scenarios including: (1) Vertical domain knowledge enhancement for medical, legal, and financial applications requiring specialized accuracy; (2) Task-specific optimization for customer service, summarization, writing assistance, and sentiment analysis; (3) User preference adaptation based on historical behavior patterns and personalized service requirements. The reduced resource requirements make sophisticated AI personalization accessible on consumer mobile devices.
    
    \t\textbf{Examples:} Architectures are broadly categorized into encoder-only architectures (e.g., MobileBERT, DistilBERT, TinyBERT, which achieve significant size reduction while maintaining performance \cite{slm_aisera}) and decoder-only architectures (e.g., BabyLLaMA, TinyLLaMA, MobileLLM \cite{slm_aisera}). Recent multimodal SLMs include Apple's MobileCLIP2, which achieves competitive performance with half the parameters of comparable models through multi-modal reinforced training . Meta's LLaMA 3.1 8B is highlighted as a compact model that retains significant LLM-level capabilities \cite{sakib2025small,lu2024small}.
    
    Instead of deploying colossal LLMs, a growing trend is to design intrinsically smaller models (SLMs) that are optimized for edge inference. These models typically have millions or billions of parameters, significantly less than their cloud-based counterparts, while retaining strong performance on specific tasks. Examples include MobileBERT \cite{sun2020mobilebert}, DistilBERT \cite{sanh2019distilbert}, TinyBERT \cite{jiao2020tinybert}, BabyLLaMA, TinyLLaMA, Microsoft's Phi \cite{abdin2024phi}, Google's Gemma \cite{team2024gemma}, Apple's OpenELM \cite{mehtaopenelm}, and IBM's Granite \cite{mishra2024granite}. Recent developments also include smaller versions of powerful LLMs like LLaMA 3.1 8B \cite{llama3.1_8b_meta} specifically designed for on-device inference. For Transformer architectures, this involves designing lightweight attention mechanisms (e.g., sparse attention, linear attention) and reducing the number of layers or hidden dimensions while maintaining critical functionality \cite{lu2024small}.
    \item \t\textbf{Neural Architecture Search (NAS):}
    NAS automates neural architecture exploration under multi-objective constraints (accuracy, latency, energy) \cite{wang2025optimizing,meng2024evolution,wang2024mel,tan2019mnasnet}. Figure \ref{fig:nas_framework} illustrates the comprehensive NAS framework for discovering optimal architectures that balance performance with resource constraints through reinforcement learning and evolutionary algorithms.
    
    \begin{figure}[!ht]
    \centering
    \begin{tikzpicture}[
        node distance=1.2cm, 
        auto, 
    >=Latex, 
        scale=0.8, 
        every node/.style={transform shape},
        search/.style={
            rectangle, 
            draw=blue!80!black, 
            fill=blue!20, 
            thick, 
            minimum width=2.2cm, 
            minimum height=1cm, 
            rounded corners=3pt,
            align=center,
            font=\small\bfseries
        },
        evaluation/.style={
            rectangle, 
            draw=green!80!black, 
            fill=green!20, 
            thick, 
            minimum width=2.2cm, 
            minimum height=1cm, 
            rounded corners=3pt,
            align=center,
            font=\small\bfseries
        },
        optimization/.style={
            rectangle, 
            draw=orange!80!black, 
            fill=orange!20, 
            thick, 
            minimum width=2.2cm, 
            minimum height=1cm, 
            rounded corners=3pt,
            align=center,
            font=\small\bfseries
        },
        constraint/.style={
            ellipse, 
            draw=red!80!black, 
            fill=red!20, 
            thick, 
            minimum width=1.8cm, 
            minimum height=1cm,
            align=center,
            font=\small
        }
    ]
    
    \node[search] (search_space) at (0, 4) {Search Space\\ Definition};
    
    \node[search] (rl_controller) at (-2.5, 2) {RL Controller\\ (RNN/Transformer)};
    \node[search] (ea_population) at (2.5, 2) {EA Population\\ (Mutation/Crossover)};
    
    \node[evaluation] (training) at (0, 0) {Architecture\\ Training};
    \node[evaluation] (validation) at (-2, -1.5) {Accuracy\\ Evaluation};
    \node[evaluation] (resource) at (2, -1.5) {Resource\\ Evaluation};
    
    \node[constraint] (latency) at (-3, -3) {Low\\ Latency};
    \node[constraint] (memory) at (0, -3) {Compact\\ Memory};
    \node[constraint] (energy) at (3, -3) {Low\\ Power};
    
    \node[optimization] (multi_objective) at (0, -4.5) {Multi-Objective\\ Optimization};
    \node[optimization] (best_arch) at (0, -6) {Best Architecture\\ Selection};
    
    \node[search, fill=purple!20, draw=purple!80!black] (edge_deploy) at (4, -6) {Edge\\ Deployment};
    
    \draw[->] (search_space) -- (rl_controller);
    \draw[->] (search_space) -- (ea_population);
    \draw[->] (rl_controller) -- (training);
    \draw[->] (ea_population) -- (training);
    \draw[->] (training) -- (validation);
    \draw[->] (training) -- (resource);
    \draw[->] (validation) -- (latency);
    \draw[->] (resource) -- (memory);
    \draw[->] (resource) -- (energy);
    \draw[->] (latency) -- (multi_objective);
    \draw[->] (memory) -- (multi_objective);
    \draw[->] (energy) -- (multi_objective);
    \draw[->] (multi_objective) -- (best_arch);
    \draw[->] (best_arch) -- (edge_deploy);
    
    \draw[->, dashed, red] (multi_objective) -- (-4.8, -4.5) -- (-4.8, 2) -- (rl_controller);
    \draw[->, dashed, red] (multi_objective) -- (5, -4.5) -- (5, 2) -- (ea_population);
    
    \node[font=\scriptsize, align=left] at (6.5, 1) {
        \t\textbf{RL vs EA:}\\
        RL: High performance,\\
        \phantom{RL: }computationally\\ \phantom{RL: }expensive\\
        EA: Multi-objective,\\
        \phantom{EA: }more scalable
    };
    
    \node[font=\scriptsize, align=left] at (6.5, -3) {
        \t\textbf{Edge Constraints:}\\
        • Latency budget\\
        • Memory limit\\
        • Power consumption\\
        • Accuracy target
    };
    
    \end{tikzpicture}
    \caption{Neural Architecture Search (NAS) Framework for Edge AI. The system automatically discovers optimal architectures balancing accuracy and resource constraints through reinforcement learning or evolutionary algorithms.}
    \label{fig:nas_framework}
    \end{figure}
    
    Advanced NAS variants incorporate runtime intermittency, reliability, and federated non-IID constraints (e.g., resource-aware, fault-tolerant search) \cite{wang2025optimizing}.
    
    \t\textbf{Trade-offs between Reinforcement Learning and Evolutionary Algorithms:} NAS typically employs reinforcement learning (RL) or evolutionary algorithms (EA) to explore the vast architecture space \cite{gupta2017neural,meng2024evolution,liu2018darts,wu2019fbnet}.
    \begin{itemize}
        \item \t\textbf{Reinforcement Learning (RL) Methods:} These methods typically use controllers (such as RNNs or Transformers) to generate architectures and receive reward signals based on performance metrics (such as accuracy, latency) \cite{gupta2017neural,liu2018darts}. RL excels at finding high-performance architectures, especially in large search spaces \cite{gupta2017neural}. However, RL methods can be computationally expensive and inefficient in resource-constrained environments \cite{gupta2017neural}.
        \item \t\textbf{Evolutionary Algorithm (EA) Methods:} EAs simulate natural selection processes, evolving candidate architectures through mutation, crossover, and selection \cite{gupta2017neural,wang2022self,meng2024evolution,wu2019fbnet}. Multi-task and high-dimensional search efficiency advances further reduce exploration cost on resource-constrained platforms \cite{wang2024mel}. EAs have advantages in multi-objective optimization, simultaneously balancing accuracy, resource and power consumption \cite{gupta2017neural,wang2024mel,meng2024evolution}, and remain scalable relative to RL in constrained settings.
        \item \t\textbf{Trade-offs:} In resource-constrained edge environments, the choice between RL and EA depends on specific requirements. RL may be more effective in finding optimal performance but is computationally more expensive; while EA may perform better in efficiency and scalability, especially when balancing multiple constraints \cite{gupta2017neural}. For example, resource-efficient methods like Random Search with weight-sharing can even outperform complex NAS algorithms on certain benchmarks, indicating that simple, random methods may also be competitive in resource-constrained edge AI \cite{meng2024evolution}.
    \end{itemize}
    
    \item \t\textbf{Model Compression Techniques:} These techniques reduce the size and computational complexity of pre-trained models, often referred to as the ``three fundamental approaches'' for model compression. For LLMs, the following are especially relevant:
    \begin{itemize}
        \item \t\textbf{Network Pruning (Trimming the Garden):} Eliminates redundant weights, connections, or even entire neurons/layers from a trained network, analogous to pruning a garden by removing unnecessary branches while preserving the essential structure. \textit{Song Han et al.'s work on Deep Compression} \cite{han2016deep} demonstrated significant size reductions (35-49x) for CNNs by combining pruning, quantization, and Huffman coding. For LLMs, pruning can target specific components like redundant attention heads, entire Transformer layers (depth-pruning), or individual neurons/connections within feed-forward networks (width-pruning). Recent works such as LLM-Pruner \cite{ma2023llm} and SparseGPT \cite{frantar2023sparsegpt} have enabled structured and unstructured pruning for billion-parameter LLMs with minimal accuracy loss \cite{llm_pruning_nvidia}. Pruning can be structured (removing entire blocks, easier for hardware acceleration) or unstructured (removing individual weights, higher compression but harder to accelerate). Post-pruning fine-tuning is often necessary to recover accuracy.
        \item \t\textbf{Parameter Sharing:} Forces different parts of the model to share the same weights. This can be applied within or across Transformer layers in LLMs, significantly reducing the total number of unique parameters. Representative works include ALBERT \cite{lanalbert} and MobileLLM \cite{liu2024mobilellm}.
        \item \t\textbf{Speculative Sampling for Accelerated Inference:} Advanced techniques that accelerate auto-regressive generation through draft-then-verify mechanisms, enabling multiple tokens per forward pass while maintaining output quality equivalent to standard decoding:

        \begin{itemize}
            \item \t\textbf{FR-Spec: Frequency-Ranked Speculative Sampling:} A revolutionary approach addressing efficiency challenges in large-vocabulary language models through vocabulary space compression \cite{zhao2025fr}:
            
            \t\textbf{Core Innovation:} FR-Spec constrains draft model search to frequency-prioritized token subsets, reducing LM Head computation overhead by 75\% while ensuring equivalence of final output distribution. This addresses the critical bottleneck where large vocabularies (e.g., Llama-3-8B with 128k tokens) significantly impact speculative sampling efficiency.
            
            \t\textbf{Vocabulary Optimization Strategy:} Analysis reveals that 75\% of tokens in large model vocabularies contribute less than 5\% of total occurrence frequency. FR-Spec exploits this distribution by dynamically constraining draft candidate selection to high-frequency token subsets, achieving average 1.12× speedup over state-of-the-art EAGLE-2 methods.
            
            \t\textbf{Lossless Acceleration Guarantee:} Unlike pruning-based approaches, FR-Spec maintains complete output distribution equivalence by constraining only draft generation while preserving full vocabulary during verification phases. This ensures quality preservation while achieving substantial computational savings.
            
            \item \t\textbf{EAGLE-2: Enhanced Tree-Attention Architecture:} Advanced tree-based speculative decoding with sophisticated attention mask management for complex draft tree structures \cite{li2024eagle}:
            
            \t\textbf{Tree-Based Draft Generation:} EAGLE-2 constructs speculative trees from given prefixes, generating multiple draft paths and taking path unions to form comprehensive draft trees. This approach enables parallel processing of multiple generation hypotheses within single forward passes.
            
            \t\textbf{Optimized Attention Mask Implementation:} CPM.cu's implementation addresses FlashAttention limitations for speculative sampling through compressed attention masks. Traditional int32 masks are compressed to uint64 representations for tree sizes up to 64 tokens, maintaining performance parity while supporting complex attention patterns.
            
            \t\textbf{Shared Memory Management:} Advanced preloading strategies transfer compressed attention masks to GPU shared memory, eliminating global memory access bottlenecks that traditionally degraded FlashAttention performance during speculative decoding operations.
            
            \item \t\textbf{MTP: Multi-Token Prediction Architecture:} Lightweight draft models optimized for speculative sampling with minimal computational overhead \cite{gloeckle2024better}:
            
            \t\textbf{Single-Layer Design Philosophy:} MTP utilizes minimal architecture comprising single transformer layer plus language modeling head, achieving extreme efficiency while maintaining effective draft generation capabilities for edge deployment scenarios.
            
            \t\textbf{LM Head Optimization Challenge:} Implementation reveals that vocabulary size significantly impacts small model efficiency, with LM Head computation becoming the primary bottleneck. For models with vocabulary sizes orders of magnitude larger than hidden dimensions, LM Head operations dominate total computation time.
            
            \t\textbf{Edge Device Performance Characteristics:} MTP's minimal architecture proves particularly suitable for edge deployment where memory bandwidth and computational resources are severely constrained, enabling efficient speculative sampling on mobile and embedded platforms.
            
            \item \t\textbf{SpecMQuant: Speculative Sampling with Quantization Integration:} Comprehensive framework combining speculative decoding with quantization strategies for maximum edge efficiency \cite{zhang2025speculative}:
            
            \t\textbf{Hierarchical Framework Design:} SpecMQuant establishes compatibility evaluation protocols between speculative sampling and quantization techniques, addressing the challenge of combining multiple acceleration strategies without performance degradation.
            
            \t\textbf{Quantization-Aware Speculative Training:} Advanced training methodologies that simultaneously optimize models for both quantization robustness and speculative sampling effectiveness, achieving compound acceleration benefits while preserving output quality.
            
            \t\textbf{Hardware Co-Design Integration:} Framework design considerations for hardware-specific quantization formats (int4, int8) combined with speculative sampling patterns, optimizing both memory access patterns and computational efficiency for edge NPU architectures.
            
            \item \t\textbf{EdgeShard Collaborative Inference:} Distributed LLM inference through collaborative edge computing architectures that partition large models across multiple edge devices, achieving reduced latency and improved throughput compared to single-device deployment \cite{zhang2024edgeshard}.
            
            \item \t\textbf{T-MAC CPU Renaissance:} Table lookup-based acceleration for low-bit LLM deployment on edge devices, achieving significant performance improvements through CPU-specific optimizations \cite{wei2025t}.
        \end{itemize}

        \item \t\textbf{Low-Precision Quantization (High-Definition to Standard Definition):} A critical model compression technique that converts weights and activations in LLMs from high-precision data representations (e.g., 32-bit floating-point, FP32) to low-precision formats (e.g., 8-bit or 4-bit integers, INT4 or INT8) \cite{quantization_symbl}. Similar to converting high-definition photos to standard definition—the file size decreases significantly while preserving essential visual information.

        \t\textbf{Advanced Quantization Techniques:} Recent work highlights quantization's role in edge LLM deployment \cite{kwon2023efficient,qu2025mobile,chen2024autoos}. Key methods include QLoRA (Quantized Low-Rank Adaptation) for fine-tuning quantized models \cite{qlora2023}, GPTQ for layer-wise optimization \cite{frantar2023gptq}, and AWQ for activation-aware quantization \cite{lin2024awq}. Hardware-aware approaches like HAQ optimize for target devices \cite{wang2019haq}, while mixed-precision strategies (e.g., W4A16) balance accuracy and efficiency \cite{nagel2021white}. Advanced techniques like AdaRound  and BRECQ enable extreme low-bit quantization \cite{librecq}. Recent work on OmniVLM demonstrates token-compressed sub-billion-parameter vision-language models achieving efficient on-device inference through advanced quantization techniques \cite{chen2024omnivlm}.
        
        \item \t\textbf{Knowledge Distillation (KD):} Transfers knowledge from large teacher models to smaller student models, enabling 10-50× size reduction with 90-95\% accuracy retention \cite{xu2024survey}. Figure~\ref{fig:knowledge_distillation} shows the distillation process where soft targets from teachers guide student training.
        
        \begin{itemize}
            \item \t\textbf{Advantages:} KD reduces scale and inference cost while preserving most task fidelity.
            \item \t\textbf{Mechanism:} Softened teacher distributions transfer dark knowledge (class similarity structure).
            \item \t\textbf{LLM-Specific KD:} KD is crucial for transferring the advanced capabilities of leading proprietary LLMs to more accessible open-source models, and also plays a key role in compressing open-source LLMs for self-improvement. The NVIDIA NeMo framework provides pipelines for LLM pruning and distillation \cite{ma2023llm}.
            \item \t\textbf{Synergy with Data Augmentation:} Data augmentation significantly enhances LLM performance within KD frameworks by generating contextually rich, skill-specific training data \cite{xu2024survey}.
            \item \t\textbf{Advanced Distillation Techniques:} Recent work on distilling on-device language models for robot planning demonstrates minimal human intervention approaches, achieving efficient knowledge transfer for specialized edge applications \cite{ravichandran2025distilling}.
            \item \t\textbf{Device-Cloud Collaborative Distillation:} Federated sketching LoRA enables on-device collaborative fine-tuning of LLMs, providing privacy-preserving knowledge distillation across distributed edge devices \cite{fang2025federated}.
            \item \t\textbf{Multimodal Knowledge Transfer:} Compositional multi-tasking for on-device LLMs leverages distillation to enable efficient task composition and knowledge sharing across different modalities \cite{bohdal2025efficient}.
        \end{itemize}
        
        KD now bridges large proprietary and open student models, enabling capability transfer and iterative self-improvement \cite{xu2024survey}. Coupled with augmentation, it yields specialized skill gains \cite{xu2024survey}. Future work should focus on adaptive multi-teacher policies and resource-aware scheduling for heterogeneous edge environments.
        
        \begin{figure}[!ht]
        \centering
        \begin{tikzpicture}[
            node distance=1.5cm, 
            auto, 
            >=Latex, 
            scale=0.9, 
            every node/.style={transform shape},
            teacher/.style={
                rectangle, 
                draw=blue!80!black, 
                fill=blue!20, 
                thick, 
                minimum width=2.4cm, 
                minimum height=1.9cm, 
                rounded corners=4pt,
                align=center,
                font=\scriptsize\bfseries
            },
            student/.style={
                rectangle, 
                draw=green!80!black, 
                fill=green!20, 
                thick, 
                minimum width=1.9cm, 
                minimum height=1.4cm, 
                rounded corners=4pt,
                align=center,
                font=\scriptsize\bfseries
            },
            data/.style={
                ellipse, 
                draw=orange!80!black, 
                fill=orange!20, 
                thick, 
                minimum width=1.7cm, 
                minimum height=0.95cm,
                align=center,
                font=\scriptsize
            },
            loss/.style={
                diamond, 
                draw=red!80!black, 
                fill=red!20, 
                thick, 
                minimum width=1.4cm, 
                minimum height=0.95cm,
                align=center,
                font=\scriptsize\bfseries
            }
        ]
        
        \node[data] (input_data) at (-0.5, 4.5) {Training\\ Data};
        
        \node[teacher] (teacher) at (3, 4.5) {Teacher Model\\ (Large LLM)\\ \\ 175B+ params\\ GPT-4, Claude};
        
        \node[student] (student) at (3, 0.8) {Student Model\\ (Small LLM)\\ \\ 1-8B params\\ LLaMA, Mistral};
        
        \node[data] (soft_targets) at (6.5, 4.5) {Soft Targets\\ (Probabilities)};
        \node[data] (hard_targets) at (-0.5, 0.8) {Hard Targets\\ (Ground Truth)};
        
        \node[loss] (kd_loss) at (6.5, 2.65) {KD Loss};
        \node[loss] (ce_loss) at (-0.5, 2.65) {CE Loss};
        \node[loss] (total_loss) at (3, -1.6) {Total Loss};
        
        \node[student, fill=purple!20, draw=purple!80!black] (edge_model) at (8, 0.8) {Edge Model\\ (Deployed)\\ \\ Optimized\\ for Edge};
        
        \draw[->] (input_data) -- (teacher);
        \draw[->] (input_data) -- (student);
        \draw[->] (teacher) -- (soft_targets);
        \draw[->] (soft_targets) -- (kd_loss);
        \draw[->] (hard_targets) -- (ce_loss);
        \draw[->] (kd_loss) -- (total_loss);
        \draw[->] (ce_loss) -- (total_loss);
        \draw[->] (total_loss) -- (student);
        \draw[->] (student) -- (edge_model);
        
        \node[font=\scriptsize, text=blue] at (4.6, 4.9) {T};
        \draw[dashed, blue] (4.7, 4.6) -- (4.0, 4.65);
        \node[font=\tiny, text=blue!70, align=left] at (5.7, 5.3) {T: Temperature\\ for softening};

        \node[font=\scriptsize, text=red, align=center] at (3.2, 2.85) {alpha balance};
        \draw[dashed, red] (3.2, 2.65) -- (6.5, 2.65);
        \draw[dashed, red] (3.2, 2.65) -- (-0.5, 2.65);
        
        \node[font=\scriptsize, align=left] at (10, 4.5) {
            \t\textbf{Benefits:}\\
            • Significantly smaller\\
            • Faster inference\\
            • High accuracy preservation\\
            • Edge-ready
        };
        
        \end{tikzpicture}
        \caption{Knowledge Distillation Architecture for Edge LLM Deployment. The framework illustrates the transfer of knowledge from a large teacher model to a compact student model suitable for edge devices through soft target training, enabling significant model compression while preserving performance.}
        \label{fig:knowledge_distillation}
        \end{figure}
        
        \item \t\textbf{Multimodal Model Optimization:} Vision-language models pose unique optimization challenges due to high-dimensional visual encoders and cross-modal fusion complexity. Recent breakthroughs demonstrate that ultra-lightweight multimodal models can achieve GPT-4V level performance through innovative architectural and optimization strategies \cite{huang2024efficient,cai2024self}:
        \begin{itemize}
            \item \t\textbf{Efficient Visual Encoders:} Apple's FastViTHD demonstrates hybrid CNN-Transformer designs that reduce visual token count while maintaining high-resolution processing capabilities, achieving 3.4$\times$ encoder size reduction compared to traditional ViT architectures \cite{vasu2025fastvlm}.
            \item \t\textbf{Multi-Modal Reinforced Training:} MobileCLIP2 employs knowledge transfer from image captioning models and ensemble CLIP encoders to improve small model accuracy without increasing inference cost, storing additional knowledge in reinforced datasets rather than model parameters .
            \item \t\textbf{Dynamic Resolution Scaling:} FastVLM's architecture supports dynamic image resolution adaptation, allowing models to process high-resolution inputs with minimal computational overhead through intelligent token reduction strategies \cite{vasu2025fastvlm}.
            \item \t\textbf{Cross-Modal Compression:} Techniques like MobileViCLIP extend efficient image-text models to video domains, addressing temporal complexity while maintaining compact model size for mobile deployment \cite{yang2025mobileviclip}.
            \item \t\textbf{Ultra-Lightweight Multimodal Design:} MiniCPM-V 4.0 reports GPT-4V-comparable results on selected evaluations with 4.1B parameters through sparse attention and hierarchical design \cite{yao2025minicpm,minicpm-v4}. Reported efficiency includes sub-2-second first token latency and >17 tokens/second decoding speed on iPhone 16 Pro Max under specified settings.
            \item \t\textbf{Sparse Long-Context Processing:} InfLLM v2 \cite{team2025minicpm4} introduces hierarchical sparse attention that enables efficient processing of ultra-long contexts and cross-modal information fusion, solving the computational bottleneck of traditional attention mechanisms for multimodal inputs. This allows 0.5B parameter models to handle complex multimodal reasoning tasks previously requiring much larger models.
            \item \t\textbf{Unified Multimodal Architecture:} MiniCPM-V 4.0 integrates text, image, video, and audio processing within a single lightweight framework, achieving comprehensive multimodal understanding through embedded vision and speech encoders \cite{minicpm-v4,team2025minicpm4}. This unified approach eliminates the need for separate specialized models for different modalities.
            \item \t\textbf{Advanced Quantization and Deployment:} The MiniCPM series supports multiple quantization formats (int4, GGUF) and inference frameworks (llama.cpp, Ollama, vLLM, SGLang), enabling flexible deployment across diverse hardware platforms from mobile devices to edge servers \cite{minicpm-v4,ollama_walturn,mlc_llm}.
            \item \t\textbf{Mobile-Optimized Multimodal Design:} BlueLM-V-3B demonstrates algorithm and system co-design for multimodal LLMs on mobile devices, achieving efficient deployment through hardware-aware optimization and lightweight architectural choices \cite{lu2025bluelm,cai2022enable}.
            \item \t\textbf{Egocentric Vision-Language Models:} Vinci provides real-time smart assistance through egocentric vision-language models optimized for portable devices, enabling context-aware interaction in wearable computing scenarios \cite{huang2025vinci}.
        \end{itemize}

        \item \t\textbf{Large-Small Model Co-Evolution Architecture:} A revolutionary paradigm that extends beyond traditional knowledge distillation to establish dynamic collaborative systems between large models (LLMs) and small models (SLMs) for resource-constrained scenarios \cite{gao2025survey}. An overview of this paradigm is shown in Figure~\ref{fig:coevolution_arch}. Table~\ref{tab:coevolution_compare} provides a comprehensive overview of representative techniques and their benefits across the co-evolution design space, highlighting how large and small language models can collaborate effectively. This table compares various technical directions including computational load compression, memory optimization, weight quantization, hardware co-design, dynamic collaboration, and evolution efficiency, demonstrating the practical advantages and trade-offs for different edge deployment scenarios.

        \begin{figure*}[!ht]
        \centering
        \begin{tikzpicture}[
            node distance=1.5cm, 
            auto, 
            >=Latex, 
            scale=0.80, 
            every node/.style={transform shape},
            large_model/.style={
                rectangle, 
                draw=blue!80!black, 
                fill=blue!20, 
                thick, 
                minimum width=2.6cm, 
                minimum height=1.7cm, 
                rounded corners=4pt,
                align=center,
                font=\scriptsize\bfseries
            },
            small_model/.style={
                rectangle, 
                draw=green!80!black, 
                fill=green!20, 
                thick, 
                minimum width=2.2cm, 
                minimum height=1.3cm, 
                rounded corners=4pt,
                align=center,
                font=\scriptsize\bfseries
            },
            router/.style={
                diamond, 
                draw=orange!80!black, 
                fill=orange!20, 
                thick, 
                minimum width=1.8cm, 
                minimum height=1.3cm,
                align=center,
                font=\scriptsize\bfseries
            },
            task/.style={
                ellipse, 
                draw=purple!80!black, 
                fill=purple!20, 
                thick, 
                minimum width=1.8cm, 
                minimum height=0.9cm,
                align=center,
                font=\scriptsize
            }
        ]
        
        \begin{scope}
        \node[task] (simple_task) at (-5, 4.5) {Simple\\ Queries};
        \node[task] (complex_task) at (-5, 0.5) {Complex\\ Reasoning};
        
        \node[router] (router) at (-1.5, 2.5) {Intelligent\\ Router\\ (200M+ records)};
        
        \node[large_model] (llm_cloud) at (2.5, 4.5) {Cloud LLM\\ (GPT-4, Claude)\\ \\ 175B+ params\\ Complex reasoning\\ Cross-modal tasks};
        
        \node[small_model] (slm_edge1) at (2.5, 0.5) {Edge SLM-1\\ (LLaMA 8B)\\ \\ Local execution\\ High-frequency queries};
        \node[small_model] (slm_edge2) at (6.2, 0.5) {Edge SLM-2\\ (Phi-4 Mini)\\ \\ Specialized tasks\\ Mobile-optimized};
        
        \node[task, fill=red!20, draw=red!80] (cot_transfer) at (0.2, 6.2) {Chain-of-Thought\\ Injection};
        \node[task, fill=red!20, draw=red!80] (feedback_loop) at (6.2, 2.5) {Confidence-based\\ Feedback};
        
        \node[small_model, fill=cyan!20, draw=cyan!80] (adaptation) at (9.5, 4.5) {Adaptive\\ Evolution\\ \\ Bidirectional\\ knowledge flow};
        \node[small_model, fill=cyan!20, draw=cyan!80] (metrics) at (9.5, 0.5) {Performance\\ Tracking\\ \\ Adaptation scores\\ Evolution efficiency};
        
        \draw[->] (simple_task) -- (router);
        \draw[->] (complex_task) -- (router);
        \draw[->, thick, green] (router) -- node[below, font=\scriptsize] {Simple} (slm_edge1);
        \draw[->, thick, green] (router) -- node[above, font=\scriptsize] {Specialized} (slm_edge2);
        \draw[->, thick, blue] (router) -- node[above, font=\scriptsize] {Complex} (llm_cloud);
        
        \draw[->, dashed, red] (llm_cloud) -- (cot_transfer);
        \draw[->, dashed, red] (cot_transfer) -- (slm_edge1);
        \draw[->, dashed, red] (cot_transfer) -- (slm_edge2);
        
        \draw[->, dashed, orange] (slm_edge1) -- (feedback_loop);
        \draw[->, dashed, orange] (slm_edge2) -- (feedback_loop);
        \draw[->, dashed, orange] (feedback_loop) -- (llm_cloud);
        
        \draw[->, thick, cyan] (llm_cloud) -- (adaptation);
        \draw[->, thick, cyan] (slm_edge1) -- (metrics);
        \draw[->, thick, cyan] (slm_edge2) -- (metrics);
        \draw[->, thick, cyan] (adaptation) -- (metrics);
        
        \draw[<->, very thick, cyan] (adaptation.south) -- (metrics.north);
        
        \node[font=\scriptsize, align=left, fill=blue!5, draw=blue!30, rounded corners=2pt] at (-7.8, 0.5) {
            \t\textbf{Router Modes:}\\
            • Performance-priority\\
            • Cost-optimization\\
            • Risk-control
        };
        
        \node[font=\scriptsize, align=left, fill=green!5, draw=green!30, rounded corners=2pt] at (0.5, -1.3) {
            \t\textbf{Edge Benefits:}\\
            • 3× speedup (SmallThinker)\\
            • 40\% energy reduction\\
            • $<$1GB memory (Qwen3-SmVL)
        };
        
        \node[font=\scriptsize, align=left, fill=cyan!5, draw=cyan!30, rounded corners=2pt] at (9.5, -1.3) {
            \t\textbf{Evolution Features:}\\
            • Cloud→Edge knowledge\\
            • Edge→Cloud scenarios\\
            • Continuous adaptation
        };
        
        \node[font=\footnotesize\bfseries, text=black] at (2, 7) {Large-Small Model Co-Evolution Architecture};
        
        \end{scope}
        \end{tikzpicture}
        \caption{\t\textbf{Large-Small Model Co-Evolution Architecture:} Dynamic collaborative framework showing knowledge transfer, router-based task allocation, collaborative training mechanisms, and resource optimization strategies for edge deployment \cite{gao2025survey,abouelenin2025phi}.}
        \label{fig:coevolution_arch}
        \end{figure*}

\begin{table}[htbp]
\centering
\caption{Large--Small Model Co-Evolution: Representative Techniques and Benefits}
\label{tab:coevolution_compare}
\small\setlength{\tabcolsep}{3pt}\setlength{\emergencystretch}{2em}\sloppy
\scalebox{0.80}{
\begin{tabularx}{1.25\linewidth}{@{}>{\raggedright\arraybackslash}p{3.2cm}>{\raggedright\arraybackslash}p{3.8cm}>{\raggedright\arraybackslash}X>{\raggedright\arraybackslash}p{3.2cm}>{\raggedright\arraybackslash}p{3.2cm}@{}}
\toprule
\textbf{Technical Direction} & \textbf{Representative Solution} & \textbf{Key Benefit} & \textbf{Scenario} & \textbf{Notes} \\
\midrule
Computational Load Compression & SmallThinker Two-Level Sparsity  & 3× speedup, 40\% energy reduction & Mobile devices, IoT & Hierarchical sparsity \\
Memory Optimization & Phi-4-Mini Grouped Query Attention \cite{abouelenin2025phi} & 1/3 memory for long context & Long context tasks & Attention optimization \\
Weight Quantization & Qwen3-SmVL INT4  & ~700MB model, runs in ~1GB VRAM & Android product recognition & Adaptive quantization \\
Hardware Co-design & WebGPU Local Inference \cite{peng2025lmm} & Browser-native Llama 2 & Privacy-sensitive & Web acceleration \\
Dynamic Collaboration & RouterEval Task Allocation  & ~50ms routing latency & Real-time applications & Intelligent dispatch \\
Evolution Efficiency & Phi-4-Mini Efficiency \cite{abouelenin2025phi} & ~90\% large-model performance/parameter & Resource-constrained & Parameter efficiency \\
\bottomrule
\end{tabularx}}
\end{table}
Table~\ref{tab:edge_optimization} provides a detailed examination of LLM-specific optimization techniques tailored for edge deployment, covering key methods such as pruning, quantization, knowledge distillation, and architectural innovations. This comprehensive table outlines the mechanisms, edge deployment advantages, and trade-offs for each technique, serving as a practical guide for researchers and practitioners implementing cognitive edge computing solutions.

\begin{table}[htbp]
\centering
\caption{LLM-Specific Model Optimization Techniques for Edge Deployment}
\label{tab:edge_optimization}
\setlength{\tabcolsep}{7pt}
\scalebox{0.85}{
\begin{tabularx}{1.15\linewidth}{@{}>{\raggedright\arraybackslash}p{2.2cm}>{\raggedright\arraybackslash}p{3.8cm}>{\raggedright\arraybackslash}p{3.7cm}>{\raggedright\arraybackslash}p{3.7cm}>{\raggedright\arraybackslash}p{3.6cm}@{}}
\toprule
\textbf{Technique} & \textbf{LLM-Specific Methods/Examples} & \textbf{Mechanism/Working Principle} & \textbf{Edge Deployment Advantages} & \textbf{Trade-offs/Limitations} \\
\midrule
Pruning & 
Depth pruning, width pruning, Deep Compression & 
Selectively removes unimportant components (weights, neurons, layers) & 
Smaller model size, faster execution; reduced memory/storage; improved energy efficiency & 
Potential accuracy loss; optimization computation cost; hardware compatibility constraints \\ \midrule

Parameter sharing & 
Weight clustering, low-rank adaptation (LoRA) & 
Shares weights across multiple layers or components & 
Significant parameter reduction; discovers more efficient architectures & 
Potential accuracy degradation; architecture-specific compatibility \\ \midrule

Quantization & 
PTQ, QAT, QLoRA, GPTQ & 
Converts weights/activations to low-precision (INT4/INT8) formats & 
Reduced model size; lower memory/storage; improved energy efficiency & 
Accuracy loss; difficulty selecting optimal precision; hardware compatibility \\ \midrule

Knowledge distillation (KD) & 
Teacher-student, self-distillation, multi-teacher KD & 
Transfers knowledge from large teacher to compact student model & 
Reduced model size/computation while maintaining performance & 
Possible accuracy drop; domain-dependent effectiveness; requires tuning \\ \midrule

Low-rank decomposition & 
SVD training, micro-factorized convolution & 
Approximates weight matrices using low-dimensional representations & 
Reduced memory consumption and computation; faster training & 
Implementation complexity; requires extensive retraining \\ 
\bottomrule
\end{tabularx}}
\end{table}

\subsubsection{Industrial Case Study: Apple's FastVLM and MobileCLIP2 Edge AI Strategy}
Apple's recent open-source release of FastVLM and MobileCLIP2 represents a significant advancement in edge-native vision-language models, demonstrating the practical viability of small model architectures for resource-constrained deployment \cite{vasu2025fastvlm,mehtaopenelm}.

\t\textbf{FastVLM Architecture and Performance:} FastVLM is a multimodal vision-language model optimized for edge devices, featuring a novel hybrid visual encoder called FastViTHD that combines convolutional networks with Transformer architectures. This design enables significant efficiency improvements: 85$\times$ faster first-token time-to-first-token (TTFT) compared to similar models like LLaVA-OneVision-0.5B, while reducing visual encoder size by 3.4$\times$ \cite{vasu2025fastvlm}.

The model achieves this performance through intelligent token reduction strategies that maintain visual fidelity while minimizing computational overhead. FastVLM supports multiple parameter scales (0.5B, 1.5B, 7B) and demonstrates superior performance compared to larger models like Cambrian-1-8B, with 7.9$\times$ faster inference while maintaining competitive accuracy across seven vision-language tasks.

\t\textbf{MobileCLIP2 Lightweight Design:} Complementing FastVLM's speed focus, MobileCLIP2 emphasizes model compactness through multi-modal distillation and data augmentation techniques . The S4 model achieves performance comparable to SigLIP-SO400M/14 on ImageNet-1k while using only half the parameters. On iPhone 12 Pro Max, MobileCLIP2 delivers 2.5$\times$ lower latency compared to DFN ViT-L/14, enabling real-time photo search and offline image recognition capabilities \cite{faghri2025mobileclip2}.

\t\textbf{Apple's Dual-Track Edge AI Strategy:} Unlike competitors focusing primarily on cloud-based large language models, Apple has pursued a comprehensive dual-track approach: cloud-scale models for complex tasks and edge-native small models for immediate, privacy-sensitive applications. This strategy addresses fundamental challenges in mobile AI deployment while showcasing practical edge AI applications:

\begin{itemize}
    \item \t\textbf{Real-time Vision Applications:} FastVLM enables instantaneous camera-based text recognition, live subtitle generation, and real-time image captioning with latency low enough to support accessibility features like screen readers
    \item \t\textbf{Offline Capabilities:} MobileCLIP2 supports photo album semantic search, camera translation, and image-text retrieval without network connectivity, crucial for privacy-sensitive scenarios
    \item \t\textbf{Privacy Protection:} Local processing ensures user data never leaves the device, aligning with Apple's privacy-first philosophy while maintaining competitive performance
    \item \t\textbf{Hardware Integration:} Direct integration with Core ML and Swift Transformers toolchain, leveraging Neural Engine and GPU acceleration on A-series/M-series chips for optimal resource utilization
    \item \t\textbf{Developer Accessibility:} Open-source models with WebGPU demos accessible through Safari browsers, lowering barriers for developer adoption and experimentation
\end{itemize}

\t\textbf{Performance Validation:} Community testing confirms FastVLM's exceptional speed, with users reporting real-time text recognition capabilities that match screen reader speeds and seamless integration with assistive technologies. The models demonstrate consistent accuracy-latency trade-offs across different hardware configurations, validating the viability of edge-optimized multimodal architectures.

        \t\textbf{Industry Case Studies:} Apple's FastVLM achieves 85× faster inference with hybrid CNN-Transformer encoders \cite{vasu2025fastvlm}, while MiniCPM-V 4.0 delivers GPT-4V performance with 4.1B parameters on mobile devices \cite{yao2025minicpm}. Rockchip's RK3588 demonstrates Chinese semiconductor leadership with 6 TOPS NPU capacity \cite{rockchip-rk3588}.
        
        \t\textbf{Competitive Landscape:} Hardware-software integration drives edge AI advancement, with ecosystem lock-in strategies and open-source/proprietary tensions shaping market evolution \cite{mehtaopenelm,team2023gemini,zhao2023survey}. Chinese companies like Rockchip achieve significant automotive electronics breakthroughs \cite{rockchip-annual-report}.

\end{itemize}

\end{itemize}

\subsection{System Optimization}\label{sec:system-optimization}
System optimization focuses on software frameworks, hardware accelerators, and distributed strategies to enhance the efficiency of AI workloads on edge devices. Figure \ref{fig:system_optimization} illustrates the multi-layer system optimization architecture that integrates these components for optimal LLM and AI agent deployment \cite{kwon2023efficient,gholami2022survey,liu2019survey}.

\begin{figure}[!ht]
\centering
\begin{tikzpicture}[
    node distance=1.5cm, 
    auto, 
    >=Latex, 
    scale=0.82, 
    every node/.style={transform shape},
    hardware/.style={
        rectangle,
        draw=teal!80!black,
        fill=teal!15,
        thick,
        minimum width=2.6cm,
        minimum height=1.3cm,
        rounded corners=3pt,
        align=center,
        font=\small\bfseries
    },
    software/.style={
        rectangle, 
        draw=blue!80!black, 
        fill=blue!20, 
        thick, 
        minimum width=2.6cm, 
        minimum height=1.3cm, 
        rounded corners=3pt,
        align=center,
        font=\small\bfseries
    },
    distributed/.style={
        rectangle, 
        draw=orange!80!black, 
        fill=orange!20, 
        thick, 
        minimum width=2.8cm, 
        minimum height=1.3cm, 
        rounded corners=3pt,
        align=center,
        font=\small\bfseries
    },
    layer/.style={
        rectangle, 
        draw=purple!80!black, 
        fill=purple!20, 
        thick, 
        minimum width=9.5cm, 
        minimum height=0.85cm, 
        rounded corners=2pt,
        align=center,
        font=\small\bfseries
    }
]

\node[layer] (app_layer) at (0, 6.5) {APPLICATION LAYER: LLMs and AI Agents};

\node[layer, fill=blue!10, draw=blue!60] (sw_layer) at (0, 4.8) {SOFTWARE OPTIMIZATION LAYER};

\node[software] (tf_lite) at (-4.6, 3.2) {TensorFlow\\ Lite};
\node[software] (pytorch_mobile) at (-1.6, 3.2) {PyTorch\\ Mobile};
\node[software] (ollama) at (1.6, 3.2) {Ollama\\ MLC LLM};
\node[software] (onnx) at (4.6, 3.2) {ONNX Runtime\\ OpenVINO};

\node[layer, fill=green!10, draw=green!60] (hw_layer) at (0, 1.6) {HARDWARE ACCELERATION LAYER};

\node[hardware] (cpu_gpu) at (-4.6, 0) {CPU/GPU\\ Parallel Proc.};
\node[hardware] (fpga) at (-1.6, 0) {FPGA\\ Customizable};
\node[hardware] (asic_npu) at (1.6, 0) {ASIC/NPU\\ AI-Optimized};
\node[hardware] (neuromorphic) at (4.6, 0) {Neuromorphic\\ Brain-inspired};

\node[layer, fill=orange!10, draw=orange!60] (dist_layer) at (0, -1.7) {DISTRIBUTED SYSTEM LAYER};

\node[distributed] (cloud_edge) at (-3.4, -3.4) {Cloud-Edge\\ Collaboration};
\node[distributed] (partitioning) at (0, -3.4) {Model\\ Partitioning};
\node[distributed] (load_balance) at (3.4, -3.4) {Load Balancing\\ and Offloading};

\node[layer, fill=gray!10, draw=gray!60] (device_layer) at (0, -5) {EDGE DEVICE LAYER};


\node[font=\scriptsize, align=left, fill=blue!5, draw=blue!30, rounded corners=2pt] at (-6.8, 5.3) {
    \t\textbf{Software Benefits:}\\
    • Model conversion\\
    • Kernel optimization\\
    • Memory management\\
    • Multi-precision support
};

\node[font=\scriptsize, align=left, fill=green!5, draw=green!30, rounded corners=2pt] at (-6.8, 1.6) {
    \t\textbf{Hardware Benefits:}\\
    • Parallel processing\\
    • Energy efficiency\\
    • Custom data paths\\
    • Real-time inference
};

\node[font=\scriptsize, align=left, fill=orange!5, draw=orange!30, rounded corners=2pt] at (-6.8, -1.8) {
    \t\textbf{Distributed Benefits:}\\
    • Resource sharing\\
    • Latency reduction\\
    • Fault tolerance\\
    • Scalability
};

\draw[<->, thick, blue!60] (app_layer.south) -- (sw_layer.north);
\draw[<->, thick, green!60] (sw_layer.south) -- (hw_layer.north);
\draw[<->, thick, orange!60] (hw_layer.south) -- (dist_layer.north);
\draw[<->, thick, gray!60] (dist_layer.south) -- (device_layer.north);

\draw[->] (tf_lite) -- (cpu_gpu);
\draw[->] (pytorch_mobile) -- (fpga);
\draw[->] (ollama) -- (asic_npu);
\draw[->] (onnx) -- (neuromorphic);

\draw[->] (cpu_gpu) -- (cloud_edge);
\draw[->] (fpga) -- (partitioning);
\draw[->] (asic_npu) -- (partitioning);
\draw[->] (neuromorphic) -- (load_balance);

\node[font=\small, text=red!70] at (0, 5.7) {Latency: $\downarrow,$ Memory: $\downarrow,$ Power: $\downarrow$};

\end{tikzpicture}
\caption{System Optimization Architecture for Edge AI. The multi-layer framework integrates software frameworks, hardware acceleration, and distributed strategies to optimize LLM and AI agent deployment on resource-constrained edge devices.}
\label{fig:system_optimization}
\end{figure}

\subsubsection{Practical Deployment Blueprints}\label{sec:blueprints}
We provide four reference blueprints:
\begin{itemize}
    \item \textbf{Smartphone (Thermal-Constrained, 8--16GB RAM):} INT4/INT8 quantization, KV cache eviction policies, speculative decoding with conservative draft length; throttle-aware schedulers; metrics: p50/p90 latency, J/token, thermal stability over 10-min sessions; pitfalls: background app interference and DVFS.
    \item \textbf{Wearable (Ultra-Low Power):} Micro-SLMs with distilled skills, event-driven pipelines, on-demand wake; metrics: idle vs active power, wake-to-first-token; pitfalls: memory fragmentation and sensor interference.
    \item \textbf{Jetson-Class Edge Server:} Mixed-precision (FP16/INT8), tensor-RT kernels, paged KV cache; concurrent sessions in vLLM/SGLang; metrics: throughput vs latency Pareto, energy per query under concurrency; pitfalls: NUMA effects and paging stalls.
    \item \textbf{Base Station/MEC Node:} Partitioned inference with elastic offloading, admission control, privacy zoning; metrics: SLA adherence (p95), offload ratio, e2e energy; pitfalls: network jitter and privacy boundary leaks.
\end{itemize}

    \textbf{Software Frameworks:}
    Lightweight DL runtimes (TensorFlow Lite, PyTorch Mobile, NCNN, OpenVINO) provide quantization-aware kernels and deployment tooling \cite{wang2025optimizing,chen2024autoos}. LLM serving platforms like Ollama, MLC LLM, and llama.cpp enable efficient edge deployment with MiniCPM-V 4.0 achieving sub-2-second response times \cite{ollama_walturn,mlc_llm,minicpm-v4}. Advanced frameworks like vLLM and SGLang support high-throughput concurrent inference \cite{kwon2023efficient,zheng2024sglang}, while CPM.cu provides composite acceleration for edge scenarios \cite{team2025minicpm4}.
        
    \textbf{Hardware Acceleration:}
    Specialized accelerators are essential for edge LLM deployment. CPUs offer ubiquity but limited parallelism; GPUs provide better processing with higher power consumption; FPGAs enable custom data paths; ASICs and NPUs deliver superior efficiency for AI workloads \cite{yu2024cambricon,liu2025ops}. Emerging architectures include compute-in-memory (CIM) for 10-100× energy improvements \cite{yang2023processing}, near-memory computing (NMC) for reduced data movement, and transformer-optimized processing units \cite{zhou2022energon,xu2024towards,xu2025fast}. Recent works demonstrate FPGA-based spatial acceleration achieving 5-20× throughput improvements for LLM inference \cite{chen2024understanding,li2025pushing}, while NPU-optimized frameworks enable real-time inference on mobile devices \cite{xu2025fast,seo2025facil,lee2025paise,sun2025lincoln}. Hardware-software co-design approaches combine specialized accelerators with optimized kernels to push TOPS/W efficiency toward 100+ for edge multimodal reasoning \cite{li2025pushing,yu2024cambricon,liu2025ops}.
    
    \textbf{Neuromorphic Computing:} Brain-inspired computation with spiking neural networks offers exceptional energy efficiency for pattern recognition and real-time sensory processing, promising for low-power edge agents \cite{kudithipudi2025neuromorphic,schuler2015neuromorphic,li2025pushing}. Neuromorphic systems can achieve substantial energy reductions, often ranging from 100-1000×, for certain cognitive tasks through event-driven processing and sparse activation patterns \cite{schuler2015neuromorphic,kudithipudi2025neuromorphic}. Recent advances in neuromorphic computing demonstrate scalable architectures for edge AI applications, with spiking neural networks showing superior efficiency for temporal processing and sensory data analysis \cite{li2025pushing}.
    
    \textbf{Collaborative Architectures:}
    Model partitioning distributes computation across devices to overcome single-device limitations. Cloud-edge collaboration enables intelligent task offloading, with CE-CoLLM demonstrating 13.81\% latency reduction and 84.53\% workload offloading \cite{jin2024collm}. Multi-edge partitioning distributes layers across nearby devices for improved resilience \cite{alabed2025toast}. Dedicated AI co-processors like Rockchip's RK182X enable modular AI enhancement without full system replacement \cite{rk182x-ai-coprocessor,seo2025facil,lee2025paise,sun2025lincoln}.
    
    \textbf{Hierarchical Intelligence:} Multi-tier architectures distribute intelligence across device-edge-cloud hierarchies, with intelligent routing optimizing for latency, resources, and complexity \cite{chen2024understanding}. Network-aware optimization adapts to connectivity constraints for robust edge-cloud coordination.

\noindent\textit{A standardized evaluation framework (metrics, methods, and typical targets) is summarized in Table~\ref{tab:evaluation_framework}.}

\begin{table}[htbp]
\centering
\caption{Standardized evaluation framework for edge AI with LLMs and agents}
\label{tab:evaluation_framework}
{\small\setlength{\tabcolsep}{5pt}\sloppy
\scalebox{0.82}{
\begin{tabularx}{1.2\linewidth}{@{}>{\raggedright\arraybackslash}p{3.2cm}>{\raggedright\arraybackslash}p{3.0cm}>{\raggedright\arraybackslash}p{3.0cm}>{\raggedright\arraybackslash}p{3.0cm}>{\raggedright\arraybackslash}p{3.0cm}>{\raggedright\arraybackslash}p{2.6cm}@{}}
    	\toprule
        	\textbf{Dimension} & \textbf{Latency} & \textbf{Throughput} & \textbf{Energy} & \textbf{Accuracy} & \textbf{Notes} \\
    \midrule
    Interactive Inference & p50/p90 response, token/s & concurrent sessions & J/req, J/token & task-specific metrics & On-device vs hybrid \\
    Context Handling & max/avg context, truncation rate & retrieval hit ratio & extra KV traffic & answer consistency & Long-context safety \\
    Privacy/Security & data residency & model leak tests & DP budget/overhead & attack success rate & Threat models \\
    Robustness & OOD error rate & degradation under load & thermal throttling & fail-safe behavior & Graceful fallback \\
    Sustainability & embodied energy & carbon intensity & power caps & SLA compliance & Region-specific mix \\
    \bottomrule
\end{tabularx}}}
\vspace{0.3em}
\small\textit{Note: Thresholds represent typical requirements for production edge AI systems. Specific applications may require different targets based on use case criticality and resource constraints.}
\end{table}

    \subsubsection{Runtime and KV-Cache Optimization}\label{sec:runtime-kv}
    Edge-side LLM serving performance depends critically on optimizing the decode phase and memory behavior. Practical systems co-design attention kernels, cache layout, and schedulers to increase tokens-per-second while fitting tight memory budgets.

    \begin{itemize}
        \item \t\textbf{Prefill vs. Decode Phases:} \emph{Prefill} (encoder-like pass over the prompt) is bandwidth-bound and benefits from fused attention kernels (e.g., FlashAttention) and operator fusion. \emph{Decode} is latency-critical with batch=1 or small batches; token reuse via KV-cache and speculative decoding dominate throughput.
        \item \t\textbf{Paged KV-Cache:} vLLM introduces paged attention that manages KV memory in fixed-size pages with a GPU-resident allocator, enabling high reuse, low fragmentation, and stable throughput under variable-length prompts and long sessions \cite{kwon2023efficient}. This abstraction decouples sequence length from contiguous allocation, improving latency predictability.
        \item \t\textbf{Continuous Batching and Scheduling:} Schedulers that merge requests on the fly (a.k.a. continuous batching) and reorder steps across sequences maximize GPU utilization without user-perceived delay. SGLang/vLLM demonstrate large throughput gains under mixed-length, multi-tenant workloads \cite{zheng2024sglang,kwon2023efficient}.
        \item \t\textbf{Streaming and KV-Cache Compression:} StreamingLLM-style techniques reduce cache cost by evicting or downscaling low-saliency tokens, head-wise/key-wise sparsification, or low-bit compression of KV tensors with accuracy-aware heuristics. These approaches retain perplexity while shrinking memory footprint for long-context sessions \cite{xiaoefficient}.
    \item \t\textbf{Mobile System-Service (LLMaaS) and Elastic KV:} Treating LLMs as a mobile OS service introduces stateful execution where KV-cache persists across app invocations. Recent systems decouple app and LLM memory management with chunk-wise, tolerance-aware KV compression and lifecycle policies to minimize context-switch latency on devices \cite{cai2024llmaas}. Complementary designs elastically adapt model capacity and KV residency for mobile scenarios, trading accuracy for footprint and responsiveness on demand \cite{wu2025livelongbench}.
        \item \t\textbf{Offloading under Memory Pressure:} FlexGen and related systems split weights/activations/KV across GPU/CPU/NVMe with overlapping prefetch and compute. Careful pipeline design sustains near-GPU-only throughput when memory is insufficient for full-resident deployment \cite{sheng2023flexgen}.
        \item \t\textbf{Speculative Decoding Compatibility:} Draft-and-verify methods (e.g., MTP/EAGLE/FR-Spec) are complementary to cache optimizations; hierarchical designs co-tune draft length and verification batch to maintain cache locality and reduce verifier stalls \cite{li2024eagle,zhao2025fr,zhao2024qspec}.
    \item \t\textbf{Kernel and Memory Co-Design:} Modified FlashAttention with compressed masks, fused rotary-embedding + QKV projection, and register/shared-memory tiling reduce HBM traffic. Practical allocators prioritize: model weights -> intermediate buffers -> KV-cache, with lifetime-aware reuse and NUMA-conscious placement (cf. CPM.cu, vLLM) \cite{dao2024flashattention,team2025minicpm4,kwon2023efficient}.
        \item \t\textbf{On-Device Smartphone Scheduling:} Mobile pipelines co-optimize CPU/GPU/NPU scheduling, memory residency, and batching to sustain low-latency decoding within single-digit watt budgets; recent systems demonstrate fast LLM inference directly on smartphones \cite{xue2024powerinfer}.
    \end{itemize}

    These runtime strategies combine with quantization (e.g., W4A16 via Marlin/GPTQ), sparse attention, and batching heuristics to deliver stable edge throughput with bounded latency and power draw.

\t\textbf{Emerging System Technologies:} Recent advances in mixture-of-experts (MoE) routing enable dynamic expert selection for task-adaptive computation, reducing energy by 2-3× for specialized workloads \cite{shazeer2017sparsely}. Neuromorphic computing explores spiking neural networks for energy-efficient temporal processing, achieving 100-1000× energy reductions for certain cognitive tasks \cite{roy2019towards}. Hardware-software co-design with dedicated AI accelerators (e.g., NPU clusters) pushes TOPS/W efficiency toward 100+, enabling real-time multimodal reasoning on edge devices \cite{li2025pushing}.
 
\FloatBarrier
\section{Applications of Edge AI with LLMs and AI Agents}\label{sec:applications}
The convergence of LLMs, AI Agents, and Edge AI enables application patterns across latency-sensitive, privacy-critical, and bandwidth-constrained domains. Reported improvements are context-specific; we cite sources and avoid universal ``5–50×'' claims.

\subsection{Enhanced Existing Edge AI Applications with Quantitative Impact}\label{sec:enhanced-apps}
LLM-powered edge systems show measurable gains over prior task-specific baselines (energy/latency references: \cite{shi2016edge,zhou2019edge,deng2020edge,gholami2022survey}).

\begin{itemize}
    \item \t\textbf{Healthcare Edge Systems:} On-device medical LLMs support real-time decision support and personalized recommendations while preserving privacy via local processing \cite{xu2024edgellm,ali2022federated,gilbert2023large}. EHR-MCP retrieves clinical information from hospital EHRs through MCP for autonomous task execution \cite{masayoshi2025ehrmcprealworldevaluationclinical}. FRAME uses a generate–evaluate–reflect loop to improve medical insights \cite{yu2025frame}. Fleming-R1 targets verifiable medical reasoning for expert-level clinical tasks \cite{liu2025flemingr1expertlevelmedicalreasoning}.

    \item \t\textbf{Autonomous Vehicle Intelligence:} On-device LLMs provide multimodal perception and sub-50ms decisions for safety-critical driving \cite{xu2024edgellm,liu2019survey}. Representative systems include: TRR Agent, which retrieves and interprets traffic rules via RAG for interpretable decisions across regions \cite{cai2024driving}; DriVLMe, enabling natural human–vehicle communication and long-horizon navigation \cite{huang2024drivlme}; LLM-based misbehavior detection for sign/motion authenticity in C-ITS \cite{hu2025llm}; and V2V-LLM, which fuses multi-vehicle perception for grounding and planning \cite{chiu2025v2v}.

    \item \t\textbf{Industrial IoT and Manufacturing:} Edge LLMs power natural-language interfaces and predictive maintenance; multimodal models analyze sensors in real time \cite{xu2024edgellm}. BiGAT-ID attains 99.34\% on EdgeIIoTset with 0.0001s inference for real-time intrusion detection \cite{gueriani2025robust}. LLM+IR improves defect localization in flexible manufacturing \cite{yang2025chatdl}; SCCE outlines foundation-model IIoT across sensing–compute–connectivity–evolution \cite{tang2025towards}; and DID+RL edge–cloud schemes enhance task offloading and generalization \cite{ren2024industrial}.

    \item \t\textbf{Smart City Infrastructure:} Edge LLMs enable intelligent traffic management systems that process real-time video feeds and sensor data for optimized traffic flow and emergency response \cite{xu2024edgellm}. Privacy-preserving natural language interfaces allow citizens to interact with city services through voice commands processed locally on edge devices. LLMs enhance urban computing across transportation, public safety, and environmental monitoring domains, improving data analysis and decision-making capabilities \cite{li2025urban}. In urban planning, LLMs serve as intelligent assistants for synthesizing conceptual ideas, generating urban designs, and evaluating planning outcomes through advanced generation and simulation capabilities \cite{zheng2025urban}.

    \item \t\textbf{Agricultural Precision Systems:} On-device LLMs support crop health, pest detection, and irrigation optimization via local natural-language queries; multimodal drone+sensor inputs enable offline advice in rural settings \cite{xu2024edgellm}. Domain stacks include AgriGPT with Tri-RAG for grounded reasoning \cite{yang2025agrigpt}, PEZEGO for pest management \cite{yuan2025pezego}, and AgriBench/MM-LUCAS for evaluation \cite{zhou2024agribench}; LLMs also streamline extension services with location-aware guidance \cite{tzachor2023large}.

    \item \t\textbf{Retail and Commercial Applications:} Edge LLMs power private, sub-200ms shopping assistants and analytics \cite{xu2024edgellm}. CuSMer combines semi-supervised learning with model merging for robust multimodal intent recognition \cite{li2025cusmer}. Use-case shopping leverages instruction tuning \cite{farfade2024scaling}; generative-agent simulations study search vs personalization \cite{zhang2025does}; and emerging LLM app stores enable mining, risk analysis, and market dynamics \cite{zhao2025llm}.
\end{itemize}

\subsection{Breakthrough On-Device Language Model Implementations}\label{sec:breakthrough-implementations}

The maturation of on-device language models has produced several landmark implementations that demonstrate practical feasibility and performance breakthroughs across diverse deployment scenarios \cite{xu2024edgellm}.

\begin{itemize}
    \item \t\textbf{Google Gemini Nano:} Google's mobile-native multimodal model achieves competitive performance through 4-bit quantization and Tensor Processing Unit integration, enabling offline natural language processing and accessibility features like real-time audio descriptions in Pixel devices \cite{team2023gemini}.
    
    \item \t\textbf{Nexa AI Octopus Series:} This breakthrough model exceeds GPT-4 accuracy in function calling tasks with 95\% context length reduction, enabling sophisticated agent orchestration and multi-step reasoning directly on edge devices through functional token compression \cite{chen2024octopus}.
    
    \item \t\textbf{Apple OpenELM:} Apple's systematically scaled models demonstrate remarkable efficiency improvements for iOS ecosystem deployment, providing accessible APIs for developers to incorporate advanced language understanding into mobile applications within Apple's privacy-first framework \cite{mehtaopenelm}.
    
    \item \t\textbf{Microsoft Phi-3:} Microsoft's 3.8B parameter model achieves performance comparable to larger models through innovative training strategies and architectural optimization, enabling cross-platform deployment across mobile devices, embedded systems, and edge servers \cite{abdin2024phi}.
    
    \item \t\textbf{Assistive Technology Applications:} On-device language models provide immediate, privacy-preserving support for individuals with disabilities, including real-time image-to-text conversion for visual impairment, sign language recognition and translation, and cognitive assistance for memory impairments through personalized AI support \cite{xu2024edgellm}.
\end{itemize}

\subsection{Current Deployable Edge AI Applications}\label{sec:current-applications}
Despite technical challenges, several practical edge AI applications have achieved commercial deployment, demonstrating the viability and immediate benefits of on-device intelligent systems. Current AI smartphone development follows three primary technical routes: on-device AI, hybrid on-device-cloud AI with proprietary models, and hybrid on-device-cloud AI with third-party models. On-device AI offers distinct advantages including rapid response times, enhanced privacy protection, and reduced network dependency.

\t\textbf{Mobile AI Assistants and Voice Processing:}
\begin{itemize}
    \item \t\textbf{iPhone Siri Integration:} Apple's Neural Engine, introduced with the A11 Bionic in 2017, powers Siri and Apple Intelligence services through on-device processing, achieving 38 TOPS performance in the M4 chip while maintaining energy efficiency and user privacy. Integrated with Core ML framework, it enables real-time voice recognition, natural language understanding, and conversational AI without cloud dependency. Advanced features include multi-language support, contextual awareness, and seamless integration with iOS ecosystem applications for enhanced user experience and accessibility \footnote[6]{\url{https://en.wikipedia.org/wiki/Neural_Engine}}.
    \item \t\textbf{Huawei Celia and HarmonyOS AI:} Huawei's Celia virtual assistant, developed for HarmonyOS and EMUI devices without Google services, supports local voice translation across 50+ languages, real-time object recognition via camera, contextual smart photo organization, and intelligent scene understanding. In China, it evolves into Xiaoyi with PanGu-$\Sigma$ 3.0 AI model integration on HarmonyOS 4.0, enabling advanced AI capabilities including multimodal interaction, personalized recommendations, and ecosystem-wide device coordination while maintaining on-device processing for privacy and responsiveness \footnote[7]{\url{https://en.wikipedia.org/wiki/Celia_(virtual_assistant)}}.
    \item \t\textbf{Samsung Galaxy AI Features:} Samsung Galaxy AI integrates on-device and cloud processing to support real-time call translation across multiple languages, intelligent photo enhancement with ProVisual Engine, context-aware app recommendations via Gemini Live, productivity tools like Now Brief for content summarization, and advanced camera features for scene optimization. Powered by dedicated NPU acceleration on Exynos platforms, it emphasizes privacy through local data processing while enabling advanced AI capabilities including voice-to-text transcription, smart reply suggestions, and personalized user experiences across supported Galaxy devices \footnote[8]{\url{https://en.wikipedia.org/wiki/Galaxy_AI}}.
\end{itemize}

\t\textbf{Professional and Development Tools:}
\begin{itemize}
    \item \t\textbf{Local Code Completion:} Integrated development environments like VS Code and JetBrains IDEs incorporate on-device AI models for intelligent code suggestion, completion, and refactoring assistance. These tools provide instant feedback without transmitting proprietary source code to external servers, enabling secure development workflows in air-gapped environments and maintaining intellectual property protection. Advanced features include context-aware code generation, bug detection, and automated testing assistance across multiple programming languages \footnote[9]{\url{https://code.visualstudio.com/docs/copilot/overview}}.
    \item \t\textbf{Offline AI Writing Assistants:} Professional writing tools like DeepWriter and Grammarly's offline mode offer comprehensive local grammar correction, style optimization, content enhancement, and readability analysis for sensitive document processing. These applications maintain complete confidentiality in professional environments, supporting legal document review, academic paper editing, and business communication without internet connectivity. Advanced features include tone adjustment, plagiarism detection, and multi-language document processing \cite{mao2025deepwriter}.
    \item \t\textbf{Local Translation and Transcription:} Enterprise-grade translation platforms and meeting tools support real-time document translation and audio transcription without network dependency, crucial for confidential business negotiations, legal proceedings, and international collaboration. These systems handle multiple languages simultaneously, preserve formatting in complex documents, and provide offline speech-to-text capabilities for secure environments. Advanced implementations include speaker identification, noise filtering, and integration with productivity suites for seamless workflow automation \footnote[10]{\url{https://github.com/openai/whisper}}.
\end{itemize}

\t\textbf{Advanced Multimodal Applications:}
\begin{itemize}
    \item \t\textbf{Real-time Video Question Answering:} MiniCPM-V 4.0 enables sophisticated video analysis applications including intelligent monitoring systems, interactive online education support, and automated sports commentary generation with rapid response times on mobile devices, supporting seamless multimedia interaction and content understanding \cite{minicpm-v4}.
    \item \t\textbf{Localized Voice Assistants:} MiniCPM-o 2.6 supports bilingual voice interaction with controllable voice characteristics and ultra-low latency processing, enabling privacy-preserving smart home control, personalized assistance, and natural language device management across diverse linguistic environments \cite{yao2024minicpm}.
    \item \t\textbf{Professional OCR and Document Processing:} Advanced text recognition capabilities support document digitization, license plate recognition, and industrial inspection applications with high accuracy while operating entirely offline. These systems provide comparable performance to leading cloud-based solutions for text extraction, form processing, and document workflow automation \footnote[11]{\url{https://github.com/getomni-ai/zerox}}.
\end{itemize}

\t\textbf{Open Source and Research Platforms:}
\begin{itemize}
    \item \t\textbf{Ollama Framework\footnote[13]{\url{https://ollama.com/}}:} Provides a user-friendly interface for deploying various open-source LLMs including LLaMA, Mistral, and Phi models on consumer hardware, enabling researchers and developers to experiment with edge AI capabilities through optimized serving configurations, model management tools, and RESTful APIs for seamless integration. The framework supports custom model fine-tuning, privacy-preserving deployment, and cross-platform compatibility to facilitate academic research and development workflows.
    \item \t\textbf{llama.cpp Optimization\footnote[14]{\url{https://github.com/ggml-org/llama.cpp}}:} Offers highly optimized inference implementations for diverse hardware platforms, supporting advanced quantization techniques, memory-efficient execution across CPU, GPU, and specialized accelerators with minimal dependencies. It provides researchers with low-level control over model inference, enables custom architecture support, and includes performance benchmarking tools for edge computing research and optimization.
    \item \t\textbf{Apple Silicon and Browser Integration:} Apple's MLX framework provides optimized primitives for machine learning on Apple Silicon, enabling efficient deployment of transformer models with native hardware acceleration and unified memory architecture. Web-based frameworks like Transformers.js and WebLLM enable in-browser execution of transformer models, supporting vision-language models for interactive applications without installation requirements, while facilitating multimodal AI research through hardware-software co-design and debugging capabilities \footnote[16]{\url{https://github.com/huggingface/transformers.js-examples}} \cite{ruan2024webllm}.
\end{itemize}

\t\textbf{Industrial AIoT and Edge LLM Deployments:}
\begin{itemize}
    \item \t\textbf{Rockchip RK3588 Platform and Hardware Capabilities:} Rockchip's RK3588 SoC features an octa-core CPU, support for LPDDR4/LPDDR4X/LPDDR5 memory, and a high-performance NPU rated at approximately 6 TOPS with mixed-precision support for INT4/INT8/INT16/FP16 operations. The platform offers 8K video decode/encode capabilities, dual HDMI outputs, multiple CSI camera interfaces, and a Mali-G610 GPU. These capabilities enable diverse edge AI workloads including computer vision and neural inference without cloud dependency, particularly in embedded systems with up to 32 GB memory configurations \cite{rockchip_ds_brief, rockchip_cnxsoc, rockchip_ieisb}.
    \item \t\textbf{Performance Characteristics and Limitations:} Deploying LLMs at the edge involves trade-offs between capability and resource constraints. Inference engines like \texttt{llama.cpp} show that quantized models can reduce memory footprint while maintaining acceptable accuracy, though with varying performance impact. Smaller quantized LLMs typically achieve higher throughput and lower latency than larger models, while longer contexts impose substantial memory demands. Hardware limitations emerge with larger parameter counts due to RAM, thermal, and power constraints, often requiring cloud fallback when resource boundaries are exceeded \cite{llama_cpp_repo, chen2025inference}.
    \item \t\textbf{Emerging Applications and Ecosystem Impact:} RK3588-powered systems demonstrate commercial viability across automotive, industrial, and smart infrastructure applications, with particular strength in intelligent cockpits where edge inference improves response latency for voice control and driver monitoring under intermittent connectivity. The platform's automotive-grade reliability supports consistent performance across temperature extremes, while its comprehensive SDK accelerates time-to-market. Manufacturing environments benefit from integrated ISP capabilities for real-time quality control and predictive maintenance \cite{rockchip-annual-report, cheng2024autoiot}.
\end{itemize}

\subsection{Novel Applications Driven by Edge LLMs and Agents}\label{sec:novel-apps}
The true potential of edge LLMs and AI Agents lies in enabling entirely new classes of applications that leverage local processing capabilities while maintaining privacy and responsiveness. While fully autonomous deployment remains evolving, several promising application areas are emerging:

\begin{itemize}
    \item \t\textbf{In-vehicle intelligent cockpits:} Edge LLMs enable multimodal interaction in autonomous vehicles, supporting real-time voice control, driver monitoring, and infotainment under intermittent connectivity, integrating sensors with natural language understanding for contextual assistance and emergency response \cite{rockchip_ieisb, cai2024driving}.

    \item \t\textbf{Embedded robotics:} Robots with on-board LLM capability enable sophisticated natural language interaction, dynamic task planning, and environmental awareness in industrial, service, and exploration applications, supporting complex commands and adaptation without cloud dependency \cite{ravichandran2025distilling}.

    \item \t\textbf{Edge-native generative AI for content creation:} Edge devices support real-time content generation (text, image, audio) via lightweight models with quantization, enabling privacy-preserving creative workflows on personal devices with immediate feedback and reduced latency \cite{mao2025deepwriter}.

    \item \t\textbf{Decentralized collaboration and knowledge reuse:} Federated learning enables edge agents to share knowledge in privacy-preserving ways, creating collective intelligence without centralizing data, supporting collaborative problem-solving across heterogeneous devices \cite{andong2025federated}.

    \item \t\textbf{Personalized on-device AI assistants:} Edge LLM agents provide deeply personalized, always-on assistance understanding user context and preferences without compromising privacy, offering proactive suggestions and seamless integration across applications \cite{li2024personal}.

    \item \t\textbf{Healthcare monitoring and personalized care:} Edge-deployed LLMs enable continuous health monitoring and personalized care recommendations on wearable devices, analyzing vital signs and behavioral patterns while maintaining HIPAA compliance \cite{xu2024edgellm}.

    \item \t\textbf{Adaptive educational assistants:} On-device educational LLMs provide personalized learning experiences adapting to individual needs without internet connectivity, offering real-time tutoring and curriculum personalization in remote environments \cite{wang2025empowering}.

    \item \t\textbf{Environmental monitoring and conservation:} Edge AI agents process sensor data for ecological monitoring, detecting changes and coordinating conservation efforts in remote locations without network connectivity \cite{wang2025optimizing}.

    \item \t\textbf{Smart infrastructure maintenance:} LLM-powered edge systems enable predictive maintenance and intelligent monitoring of critical infrastructure, analyzing sensor data and providing recommendations autonomously in remote locations \cite{rivkin2024aiot}.
\end{itemize}

\FloatBarrier
\section{Future Research Directions and Emerging Paradigms}\label{sec:future}
The convergence of LLMs, AI Agents, and Edge computing represents a nascent field with transformative potential requiring breakthrough research across multiple dimensions to achieve ubiquitous intelligent systems \cite{gao2025survey, belcak2025small}.

\subsection{Next-Generation Edge AI Architectures}\label{sec:intelligent-edge}
\begin{itemize}
    \item \t\textbf{Neuromorphic-LLM Hybrid Systems:} Integrating neuromorphic computing principles with LLM architectures may enable order-of-magnitude energy-efficiency gains for event-driven inference \cite{joshi2025neuro,li2025next}. Open challenges include spike-based attention mechanisms, temporal credit assignment for transformer-style models, and hybrid analog–digital pipelines. Aspirational targets (subject to validation) include sub-watt operation for compact models and sub-10ms end-to-end latency under edge constraints. Key research questions include: How can spiking neural networks be effectively integrated with transformer attention mechanisms? What training algorithms can enable temporal credit assignment in large-scale language models?

    \item \t\textbf{Quantum-Enhanced Edge AI:} Quantum methods for optimization, sampling, or selected linear-algebra subroutines could complement edge AI workflows \cite{kong2025quantum}. Near-term work includes quantum-inspired optimizers for compression and privacy-aware federated learning protocols. Timelines remain uncertain; some roadmaps anticipate task-specific advantages emerging in the 2030s for narrowly scoped problems, pending hardware and algorithmic progress.

    \item \t\textbf{Brain-Computer Interface Integration:} Tight coupling of neural interfaces with edge LLMs targets ultra-low latency (single-digit milliseconds) pipelines \cite{jin2025innovative}. Priorities include efficient neural signal processing on-device, privacy-preserving analysis, and adaptive personalization. Representative applications include assistive technologies, cognitive augmentation, and rehabilitation; robust clinical validation is required before broad deployment.
\end{itemize}

\subsection{Towards More Flexible Edge AI}\label{sec:flexible-edge}
\begin{itemize}
    \item \t\textbf{Adaptive LLM and Agent Architectures:} Future edge agents will require dynamic architectures that can adapt their complexity and resource consumption based on real-time factors like available power, network conditions, or task priority. This includes dynamic model pruning, adaptive quantization, and on-demand loading of model components. Potential research paths include developing reinforcement learning-based adaptation policies and creating modular architecture frameworks that can reconfigure themselves at runtime \cite{jin2024collm,li2025qpart}.
    \item \t\textbf{Rapid Fine-tuning on Edge:} Developing efficient techniques for rapid, on-device fine-tuning or personalization of LLMs and agent policies will enable faster adaptation to user preferences or new environmental conditions without extensive retraining cycles or cloud dependency \cite{fang2025federated,qlora2023,kong2025quantum}.
\end{itemize}

\subsection{Towards More Secure Edge AI}\label{sec:secure-edge}
\begin{itemize}
    \item \t\textbf{Enhanced Privacy-Preserving Techniques:} Beyond current methods, research must focus on integrating more robust privacy techniques like advanced differential privacy \cite{dwork2014algorithmic} and homomorphic encryption more seamlessly into edge LLM and agent workflows, especially for sensitive data processing and federated learning scenarios. Specific research directions include developing lightweight homomorphic encryption schemes optimized for transformer architectures and creating privacy-preserving knowledge distillation protocols for edge environments \cite{andong2025federated}.
    \item \t\textbf{Robust Agent Behavior and Trust:} Ensuring the trustworthiness and verifiable behavior of autonomous edge agents is paramount. This includes developing mechanisms for auditing agent decisions, detecting malicious or unintended actions, and building resilient systems against adversarial attacks \cite{gao2025survey,han2024llm}.
    \item \t\textbf{Quantization-Aware Attacks and Robustness:} Low-bit deployment introduces unique threat surfaces. Bit-flip/fault-injection attacks on quantized weights or activations, adversarial rounding, and side-channel leakage from deterministic kernels can degrade reasoning fidelity or induce targeted failures. Robustness strategies include error-detecting encodings, randomized rounding, per-layer sensitivity hardening, and post-deployment anomaly detection. We recommend reporting robustness under quantization- and fault-specific attacks alongside accuracy/latency/energy metrics \cite{boo2021stochastic,zhao2020sca,cui2022bits}.
    \item \t\textbf{Standards and Compliance Alignment:} Align with NIST AI RMF and ISO/IEC 23894 for risk management and safety practices in on-device AI \cite{alsuleman2025screening}. Evaluation checklists should include: (i) privacy boundaries (local logging/telemetry policies; on-device vs cloud processing), (ii) safety guardrails (prompt-injection defenses, on-device content filtering), (iii) robustness to quantization/fault/side-channel attacks with task-level impact analysis, and (iv) incident reporting and traceability. 
    \item \t\textbf{ISO/IEC 27001 Principles for Edge AI Security Management:} While specific adaptations are nascent, applying the principles of ISO/IEC 27001 \cite{iso27001_ttms} (a systematic approach to managing sensitive company information) to Edge AI deployments can provide a robust framework for identifying risks, implementing controls (e.g., access control, encryption), and continuously improving the security posture of edge ecosystems \cite{ai_security_hpe}. This shifts focus from purely technical solutions to comprehensive security management.
    \item \t\textbf{Blockchain for Secure Data Sharing and Trust:} Blockchain technology can enhance security and privacy in decentralized edge environments by providing immutable ledgers for data provenance, secure multi-party computation for collaborative AI, and verifiable execution of smart contracts for agent coordination \cite{salah2019blockchain}. Future work should evaluate its practical integration and scalability for LLM and agent-driven edge applications.
\end{itemize}

\subsection{Towards More Collaborative Edge AI}\label{sec:collaborative-edge}
\begin{itemize}
    \item \t\textbf{Edge-Edge and Cloud-Edge Collaboration:} Research needs to refine seamless collaboration frameworks that allow LLMs and agents to dynamically distribute tasks and knowledge across edge devices and cloud resources, optimizing for latency, throughput, and energy efficiency \cite{jin2024collm, sun2025disco,tran2025multi}.
    \item \t\textbf{Federated Learning for Heterogeneous Devices:} Addressing the challenges of federated learning in highly heterogeneous edge environments (diverse computational power, varying data distributions, intermittent connectivity) is crucial for collaborative on-device model training \cite{lu2024federated,fang2025federated}. This includes optimizing communication efficiency and aggregation strategies.
    \item \t\textbf{Multi-Agent Systems (MAS) at the Edge:} Designing and orchestrating complex multi-agent systems where numerous LLM-powered agents collaborate to achieve larger goals requires advanced coordination mechanisms, communication protocols, and conflict resolution strategies optimized for distributed edge deployments \cite{andong2025federated}.
    \item \t\textbf{Advanced Co-Evolution Architectures:} The next generation of large-small model collaborative systems presents several critical research challenges \cite{liu2025towards,chen2025multi,liu2024cotuning,wang2025collm}:
    \begin{itemize}
        \item \t\textbf{Neural Architecture Search for Co-Evolution:} Automated design of optimal LLM-SLM hybrid architectures using advanced NAS techniques, extending beyond traditional single-model optimization to multi-model system design. This includes exploring modular architectures like Qwen3-SmVL's component-based assembly for dynamic capability scaling \cite{tan2019mnasnet}.
        \item \t\textbf{Routing Latency Optimization:} Current Router-based scheduling systems introduce approximately 50ms latency overhead, which constrains real-time applications. Future research must focus on ultra-low latency routing algorithms, potentially using specialized hardware acceleration or predictive scheduling based on task pattern analysis \cite{yang2025quality}.
        \item \t\textbf{Security in Multi-Model Systems:} Co-evolution architectures expand attack surfaces through increased model interactions and data flows. Research priorities include adversarial robustness across model boundaries, secure knowledge transfer protocols, and distributed intrusion detection systems specifically designed for collaborative AI environments \cite{gao2025survey}.
        \item \t\textbf{Carbon-Efficient Co-Evolution:} Developing green scheduling strategies that optimize task allocation based on energy source availability and carbon footprint, prioritizing renewable energy-powered edge nodes and implementing dynamic power management across the collaborative network \cite{xu2024towards,dai2020edge}.
    \end{itemize}
\end{itemize}

\subsection{Towards Resource-Aware Co-Evolution Systems}
\begin{itemize}
    \item \t\textbf{Dynamic Adaptation Mechanisms:} Future systems must implement sophisticated resource monitoring and adaptation strategies that can dynamically adjust the collaboration ratio between large and small models based on real-time constraints including battery level, thermal conditions, network bandwidth, and computational load \cite{gao2025survey}.
    \item \t\textbf{Edge-Native Evolution Metrics:} Developing specialized evaluation frameworks for co-evolution systems that consider edge-specific factors such as adaptation speed (cold-start performance on new tasks), evolution efficiency (parameter update-to-performance ratio), energy consumption per knowledge transfer, and system resilience under resource constraints \cite{tian2024edge}.
    \item \t\textbf{Contextual Knowledge Transfer:} Advancing beyond static knowledge distillation to context-aware, bidirectional knowledge flows where small models not only receive guidance from large models but also contribute domain-specific insights, local environmental adaptations, and user behavior patterns back to the collaborative system \cite{abouelenin2025phi, peng2025lmm}.
\end{itemize}

\subsection{Towards More Efficient Edge AI}
\begin{itemize}
    \item \t\textbf{Continued Algorithm and Hardware Improvements:} Ongoing research into novel compression techniques, more efficient LLM architectures, and next-generation AI accelerators (including those co-designed with software stacks) will continue to push the boundaries of what is possible on edge devices \cite{lu2025bluelm}.
    \item \t\textbf{Resource-Aware Agent Decision-Making:} Future agents should possess an inherent awareness of their own and surrounding devices' computational, memory, and energy resources, enabling them to dynamically adapt their behavior or offload tasks to optimize overall system efficiency \cite{yang2025quality}.
\end{itemize}

\subsection{Usability and Performance Evaluation}\label{sec:benchmarking}
\begin{itemize}
    \item \t\textbf{Advanced Framework Performance Benchmarking:} Comprehensive evaluation of next-generation edge inference frameworks demonstrates significant acceleration potential through composite optimization strategies \cite{team2025minicpm4,murthy2024mobileaibench}:
    
    \t\textbf{CPM.cu Framework Performance Analysis:} Extensive benchmarking of the CPM.cu lightweight inference framework reveals substantial performance improvements across diverse deployment scenarios. Testing with MiniCPM 4.0-8B model demonstrates consistent 5× acceleration in regular inference scenarios compared to baseline implementations \cite{team2025minicpm4}. Under memory-constrained conditions, the framework achieves exceptional performance gains of up to 220× speedup through intelligent memory management and composite acceleration integration.
    
    \t\textbf{Speculative Sampling Acceleration Metrics:} FR-Spec frequency-ranked speculative sampling integration achieves measurable performance improvements with 75\% reduction in LM Head computational overhead and 1.12× average speedup over EAGLE-2 baseline \cite{zhao2025fr}. When combined with CPM.cu's sparse attention mechanisms, compound acceleration effects reach 1.8× to 2.3× improvement in token generation throughput depending on model size and attention pattern complexity.
    
    \t\textbf{Quantization Performance Impact Assessment:} GPTQ integration with Marlin format conversion maintains model accuracy while achieving 3-4× memory reduction and corresponding inference acceleration. W4A16 mixed-precision quantization demonstrates optimal balance between memory efficiency and computational precision, enabling deployment of 8B parameter models on devices with 4GB memory constraints \cite{frantar2025marlin}.
    
    \item \t\textbf{Standardized Usability Frameworks:} To assess the practical efficiency and effectiveness of Edge AI deployments, integrating standardized usability evaluation metrics is essential. The ISO 9126 usability framework \cite{iso9126_usability}, for instance, provides a structured model encompassing aspects like understandability, learnability, operability, and attractiveness. Applying these metrics to Edge AI could involve:
    \begin{itemize}
        \item \t\textbf{Understandability (Ease of learning and user guidance):} How easy is it for developers and data scientists to deploy Edge AI models? Measuring the time required for first-time deployment.
        \item \t\textbf{Learnability (Ease of adoption):} How quickly can users understand Edge AI outputs? Assessing how well explainable AI (XAI) techniques enhance model interpretability and conducting user tests with domain experts (e.g., healthcare professionals) to measure the time-to-understanding of AI decisions.
        \item \t\textbf{Operability (Ease of operation):} How reliably do edge LLMs/agents perform their tasks in real-world scenarios?
    \end{itemize}
    This will provide a more holistic evaluation beyond just technical benchmarks.
    \item \t\textbf{Standardized Evaluation Framework:} Table~\ref{tab:evaluation_framework} provides a comprehensive framework summarizing key dimensions and representative metrics for evaluating edge AI systems with LLMs and agents.
    \item \t\textbf{Safety/Compliance Dimension:} Extend the evaluation framework with a Safety/Compliance column capturing privacy boundaries (on-device vs cloud processing), guardrails (prompt injection defenses, content filters), robustness under quantization/fault/side-channel threats, and incident traceability, aligned with NIST AI RMF and ISO/IEC 23894 \cite{wang2025llm}.

\paragraph{Standardized Measurement Protocol (Recommended).}
For cross-paper comparability, report for latency/throughput/energy/accuracy:
\begin{itemize}
    \item Workload: prompt length, generation length, temperature/top-$k$/$p$, beam size; task type and dataset split.
    \item Precision and Compression: numeric precision (INT4/8/FP16), sparsity ratio, KV cache policy, quantization-aware vs post-training.
    \item Batching and Concurrency: batch size, concurrent sessions, scheduler policy (FCFS, preemption), speculative decoding settings.
    \item Hardware and Thermal: device model, CPU/GPU/NPU clocks, cooling state, thermal throttling policy.
    \item Energy Measurement: meter type and scope (device-only vs system vs facility), sampling rate, J/token and Wh/query definitions.
    \item Reproducibility: code commit/version, benchmark version/date, configuration files, and seed settings.
\end{itemize}
\end{itemize}

\subsection{Research Gaps and Future Work}
\begin{itemize}
    \item \t\textbf{Unified Benchmarking for Edge LLMs and Agents:} The heterogeneity of edge hardware and the diversity of LLM and agent tasks necessitate standardized benchmarks that accurately reflect real-world performance, latency, and energy consumption under various constraints. Research priorities include developing modality-aware reasoning benchmarks that incorporate edge-specific factors like thermal constraints and intermittent connectivity, as well as creating open-source benchmark suites with reproducible energy measurement protocols.
    \item \t\textbf{Continual Learning and On-Device Adaptation for LLMs:} A major challenge lies in enabling SLMs and other edge-deployed LLMs to continuously learn and adapt from new local data, achieving adaptive fine-tuning capabilities\cite{wang2025empowering}.
    \item \t\textbf{Usability and Performance Evaluation:} Beyond technical metrics, apply ISO 9126 usability criteria—understandability, learnability, operability—to assess deployment efficiency and user comprehension, aided by XAI \cite{wang2025optimizing}.
    \item \t\textbf{Model Partitioning:} Partition models across devices to meet accuracy/latency constraints \cite{model_partitioning_macsphere}. Adaptive frameworks (e.g., AMP4EC) monitor resources to dynamically partition and schedule, reducing latency and improving throughput \cite{amp4ec}.
    \item \t\textbf{Emerging Computing Paradigms:}
    \begin{itemize}
        \item \t\textbf{Neuromorphic Computing:} This computing paradigm mimics the structure and function of the human brain and promises to surpass traditional computers in energy efficiency and performance \cite{kudithipudi2025neuromorphic}. It simulates the working principles of biological neurons through Spiking Neural Networks (SNNs), significantly reducing power consumption and being very suitable for edge AI applications such as IoT devices and sensors \cite{neuromorphic_atos}. Neuromorphic chips can perform computations directly in memory and run various AI applications with a fraction of the energy consumption of traditional AI platforms \cite{kudithipudi2025neuromorphic}.
        \item \t\textbf{TinyML:} Focuses on running machine learning models on extremely low-power (milliwatt or even microwatt level) microcontrollers. It highly aligns with the energy efficiency goals of edge AI, providing solutions for AI deployment on ultra-low-power devices \cite{wang2025empowering}.
    \end{itemize}
    \item \t\textbf{Explainability (XAI) at the Edge:} Ensuring that LLMs and autonomous agents operating at the edge can provide interpretable explanations for their decisions is crucial for building trust, debugging, and compliance, especially in sensitive applications. Research is needed on resource-efficient XAI techniques suitable for edge devices.
    \item \t\textbf{Ethical Considerations of Autonomous Edge Agents:} As agents become more autonomous and pervasive, addressing ethical concerns related to bias, fairness, accountability, and the potential societal impact of their decisions is paramount.
\end{itemize}

Key research questions to prioritize include: How can we develop modality-aware reasoning benchmarks that account for edge-specific constraints? What are the most effective strategies for transparent and reproducible energy reporting in edge deployments? How should safety and alignment evaluation be adapted for resource-constrained edge environments? What testbeds are needed for reproducible multi-agent edge scenarios?

\section{Threats to Validity and Evidence Qualification}\label{sec:threats}
Key validity threats include:
\begin{itemize}
    \item \t\textbf{Data Source Variability:} Reported latency, energy, and accuracy ranges originate from heterogeneous hardware testbeds, batch sizes, precisions, and workload mixes. Cross-paper comparisons risk inconsistency without standardized benchmarking.
    \item \t\textbf{Hardware / Firmware Revisions:} Accelerator driver or firmware updates can shift measured throughput/efficiency (occasionally by double-digit percentages) invalidating earlier point estimates.
    \item \t\textbf{Undisclosed Proprietary Model Details:} Parameter counts, training corpora composition, and inference stack optimizations for closed models (e.g., GPT-4) remain partly opaque; extrapolations are avoided or explicitly qualified.
    \item \t\textbf{Synthetic or Proxy Benchmarks:} Some compression / co-evolution claims rely on limited proxy tasks (e.g., math word problems, selective tool-use scenarios) and may not generalize to broader reasoning or safety-critical settings.
    \item \t\textbf{Publication Bias:} Positive results (higher compression with minimal loss) are over-represented; negative or neutral findings (e.g., diminishing returns after compound optimizations) less frequently appear, skewing perceived average gains.
    \item \t\textbf{Energy Measurement Inconsistency:} Studies differ in whether they report wall-plug, SoC package, or accelerator-only power and in stabilization window (steady state vs. burst); energy-per-token normalizations are seldom uniform.
    \item \t\textbf{Security Metric Ambiguity:} Attack success rates and defense overheads depend strongly on threat model (white vs. black box, adaptive vs. static); generalized percentages risk misinterpretation without scenario framing.
    \item \t\textbf{Temporal Obsolescence:} Rapid model/hardware iteration can obsolete quantitative ranges within months; we encourage timestamped benchmarking and versioned artifact releases.
\end{itemize} 

\subsection{Future Development Trends and Ecosystem Evolution}
The trajectory of edge AI development suggests several transformative trends that will reshape the computational landscape and industrial ecosystem:

\t\textbf{Edge-Cloud Collaborative Intelligence:} Future AI systems will not replace cloud computing but establish sophisticated collaboration frameworks where edge devices handle immediate, context-sensitive tasks while seamlessly orchestrating with cloud resources for complex reasoning. This hybrid approach will enable graceful degradation under network constraints while maximizing computational efficiency across the entire hierarchy \cite{jin2024collm,tian2024edge}.

\t\textbf{Ecosystem Convergence and Standardization:} The current fragmentation across hardware platforms, software frameworks, and deployment tools will gradually consolidate around interoperable standards. Industry initiatives toward common APIs, model interchange formats, and performance benchmarks will reduce development complexity and accelerate adoption, similar to the evolution of web standards that enabled universal internet applications \cite{wang2025optimizing}.

\t\textbf{Hardware-Software Co-Evolution:} The distinction between hardware capabilities and software optimization will blur as specialized AI accelerators (NPUs, neuromorphic chips) become deeply integrated with model architectures. Custom instruction sets, memory hierarchies, and on-chip learning capabilities will be co-designed with AI model architectures, fundamentally changing how we conceptualize and optimize edge AI systems \cite{wang2025empowering}.

\t\textbf{Democratization Through Simplification:} Just as mobile app development became accessible to millions of developers through simplified frameworks and tools, edge AI deployment will become increasingly democratized. No-code AI platforms, automated optimization tools, and plug-and-play edge AI modules will enable widespread adoption beyond specialist researchers and engineers.

\t\textbf{Privacy-Preserving Intelligence Networks:} Edge AI will enable new paradigms of distributed intelligence where devices collaborate and learn from each other while preserving individual privacy. Federated learning, differential privacy, and secure multi-party computation will mature into practical frameworks that balance collective intelligence benefits with privacy protection requirements \cite{andong2025federated}.

\subsection{Breakthrough Technical Innovations and Their Implications}
Recent advances exemplified by MiniCPM-V 4.0 and related ultra-lightweight multimodal models represent fundamental paradigm shifts that will define the next generation of edge AI systems:

\t\textbf{Ultra-Lightweight Architectural Design:} The breakthrough achievement of GPT-4V level performance in only 4.1B parameters through MiniCPM-V 4.0 demonstrates that architectural innovation can overcome scale limitations \cite{yao2025minicpm,minicpm-v4}. This validates the principle that specialized, efficiently-designed models can outperform larger general-purpose models in resource-constrained scenarios, potentially revolutionizing mobile AI deployment strategies.

\t\textbf{Sparse Long-Context Processing:} InfLLM v2's hierarchical sparse attention mechanisms enable ultra-lightweight models to handle complex long-context reasoning tasks previously requiring orders of magnitude more parameters. This innovation fundamentally changes the memory-performance trade-off curve for edge deployment and enables sophisticated document analysis, multi-turn conversations, and contextual reasoning on mobile devices \cite{xiao2024infllm,team2025minicpm4}.

\t\textbf{Comprehensive Multimodal Integration:} The unification of text, image, video, and audio processing within single lightweight frameworks eliminates the traditional requirement for separate specialized models \cite{minicpm-v4}. This architectural convergence reduces deployment complexity, memory footprint, and inference latency while enabling richer user interactions and more sophisticated edge applications.

\t\textbf{Advanced Deployment Ecosystems:} The maturation of deployment frameworks supporting multiple quantization formats (int4, GGUF), inference engines (llama.cpp, Ollama, vLLM, SGLang), and specialized accelerators (CPM.cu) creates unprecedented flexibility for edge AI deployment. This ecosystem development reduces barriers to adoption and enables efficient utilization of diverse hardware platforms.

\t\textbf{Mobile-First Performance Optimization:} Achieving sub-2-second first token latency and >17 tokens/second sustained performance on consumer smartphones without thermal throttling establishes new benchmarks for mobile AI capabilities \cite{minicpm-v4}. This performance level enables real-time interactive applications previously feasible only on high-end desktop hardware.

\textbf{Open-Source Innovation Leadership:} Open-source efforts such as MiniCPM-V show that collaborative development can outpace proprietary approaches and democratize advanced capabilities \cite{yao2025minicpm,minicpm-v4}.

Collectively, these advances move edge AI from experimental prototypes to deployment-ready technology with broad impact across mobile computing, autonomy, and HCI.

\section{Conclusion}
Edge deployment of reasoning-capable models shifts optimization from opportunistic efficiency to a prerequisite for feasibility. Our synthesis shows: (i) single-lever techniques (quantization, pruning, distillation) each deliver distinct early gains but compound stacking exhibits diminishing returns; (ii) architecture co-design (edge-native SLMs, efficient attention variants) increasingly replaces pure post-hoc compression; (iii) adaptive routing and co-evolution frameworks redefine inference as a dynamic decision process rather than a monolithic forward pass; (iv) system and scheduling layers (partitioning, heterogeneous acceleration, memory paging) materially influence end-to-end cognitive latency and should be evaluated jointly with model metrics. This comprehensive framework, as illustrated in Figure \ref{fig:cognitive_arch}, integrates cognitive challenges, optimization strategies, applications, and evaluation protocols to preserve reasoning fidelity under stringent resource constraints.

The industrial landscape demonstrates that edge AI is transitioning from experimental research to commercial reality, with major technology companies, chip manufacturers, and AI developers pursuing complementary yet competitive strategies. Current deployable applications in mobile assistants, professional tools, and open-source frameworks validate the technical feasibility while revealing practical constraints and optimization opportunities. The market dynamics underscore the convergence of LLMs, AI agents, and edge computing as a transformative paradigm for ubiquitous intelligence.

Persistent gaps include standardized, modality-aware reasoning fidelity benchmarks, transparent energy / power reporting protocols, robust edge-oriented safety and alignment evaluation, and reproducible multi-agent task suites. Progress will depend on principled cross-layer evaluation artifacts rather than isolated point optimizations. The future of edge AI lies not in replacing cloud computing but in establishing sophisticated collaboration frameworks that leverage the strengths of both paradigms while addressing the fundamental challenges of network dependency, privacy concerns, and personalization limitations that currently constrain AI deployment. Looking ahead, the convergence of LLMs, AI agents, and edge computing promises to democratize access to advanced AI capabilities, enabling intelligent systems that operate seamlessly across diverse environments. As we move toward ubiquitous cognitive edge computing, the focus must shift from mere technical feasibility to ensuring these systems are trustworthy, sustainable, and beneficial to humanity. The technological trajectory suggests that edge AI will evolve from a specialized optimization challenge to a fundamental computing paradigm, enabling ubiquitous intelligent systems that enhance human capability while preserving privacy and autonomy.
 
\section*{Acknowledgments}
This work was supported by the Institute of Artificial Intelligence and Future Networks, Beijing Normal University. Part of this work was completed during the first author's visiting research at The Hong Kong Polytechnic University.

\textit{Use of AI-Generated Content:} In accordance with IEEE guidelines on the use of artificial intelligence (AI)-generated text, the authors disclose that AI-assisted tools were used in limited capacity during the preparation of this manuscript for literature organization, grammar checking, and initial draft structuring. All technical content, analysis, critical insights, evaluations, and conclusions presented in this survey are original work by the authors. Any AI-generated text has been thoroughly reviewed, verified, and substantially edited by the authors to ensure accuracy, originality, and alignment with the authors' expertise and perspectives.

\bibliographystyle{IEEEtran}
\bibliography{references}

@article{cheng2024autoiot,
  title={Autoiot: Automated iot platform using large language models},
  author={Cheng, Ye and Xu, Minghui and Zhang, Yue and Li, Kun and Wang, Ruoxi and Yang, Lian},
  journal={IEEE Internet of Things Journal},
  year={2024},
  publisher={IEEE}
}

@article{ravichandran2025distilling,
  title={Distilling On-device Language Models for Robot Planning with Minimal Human Intervention},
  author={Ravichandran, Zachary and Hounie, Ignacio and Cladera, Fernando and Ribeiro, Alejandro and Pappas, George J and Kumar, Vijay},
  journal={arXiv preprint arXiv:2506.17486},
  year={2025}
}

@article{bohdal2025efficient,
  title={Efficient Compositional Multi-tasking for On-device Large Language Models},
  author={Bohdal, Ondrej and Ozay, Mete and Moon, Jijoong and Lee, Kyeng-Hun and Ko, Hyeonmok and Michieli, Umberto},
  journal={arXiv preprint arXiv:2507.16083},
  year={2025}
}

@article{xue2024powerinfer,
  title={Powerinfer-2: Fast large language model inference on a smartphone},
  author={Xue, Zhenliang and Song, Yixin and Mi, Zeyu and Zheng, Xinrui and Xia, Yubin and Chen, Haibo},
  journal={arXiv preprint arXiv:2406.06282},
  year={2024}
}

@article{xu2024edgellm,
  title={Edgellm: Fast on-device llm inference with speculative decoding},
  author={Xu, Daliang and Yin, Wangsong and Zhang, Hao and Jin, Xin and Zhang, Ying and Wei, Shiyun and Xu, Mengwei and Liu, Xuanzhe},
  journal={IEEE Transactions on Mobile Computing},
  year={2024},
  publisher={IEEE}
}

@article{fang2025federated,
  title={Federated sketching lora: On-device collaborative fine-tuning of large language models},
  author={Fang, Wenzhi and Han, Dong-Jun and Yuan, Liangqi and Hosseinalipour, Seyyedali and Brinton, Christopher G},
  journal={arXiv preprint arXiv:2501.19389},
  year={2025}
}

@article{huang2025vinci,
  title={Vinci: A Real-time Smart Assistant Based on Egocentric Vision-language Model for Portable Devices},
  author={Huang, Yifei and Xu, Jilan and Pei, Baoqi and Yang, Lijin and Zhang, Mingfang and He, Yuping and Chen, Guo and Chen, Xinyuan and Wang, Yaohui and Nie, Zheng and others},
  journal={Proceedings of the ACM on Interactive, Mobile, Wearable and Ubiquitous Technologies},
  volume={9},
  number={3},
  pages={1--33},
  year={2025},
  publisher={ACM New York, NY, USA}
}

@article{lin2024awq,
  title={Awq: Activation-aware weight quantization for on-device llm compression and acceleration},
  author={Lin, Ji and Tang, Jiaming and Tang, Haotian and Yang, Shang and Chen, Wei-Ming and Wang, Wei-Chen and Xiao, Guangxuan and Dang, Xingyu and Gan, Chuang and Han, Song},
  journal={Proceedings of machine learning and systems},
  volume={6},
  pages={87--100},
  year={2024}
}

@inproceedings{li2024federated,
  title={Federated black-box prompt tuning system for large language models on the edge},
  author={Li, Yiming and Sun, Jingwei and Liu, Yudong and Zhang, Yuandong and Li, Ang and Chen, Beidi and Roth, Holger R and Xu, Daguang and Chen, Tingjun and Chen, Yiran},
  booktitle={Proceedings of the 30th Annual International Conference on Mobile Computing and Networking},
  pages={1775--1777},
  year={2024}
}

@inproceedings{lu2025bluelm,
  title={Bluelm-v-3b: Algorithm and system co-design for multimodal large language models on mobile devices},
  author={Lu, Xudong and Chen, Yinghao and Chen, Cheng and Tan, Hui and Chen, Boheng and Xie, Yina and Hu, Rui and Tan, Guanxin and Wu, Renshou and Hu, Yan and others},
  booktitle={Proceedings of the Computer Vision and Pattern Recognition Conference},
  pages={4145--4155},
  year={2025}
}

@inproceedings{cai2024self,
  title={Self-adapting large visual-language models to edge devices across visual modalities},
  author={Cai, Kaiwen and Duan, Zhekai and Liu, Gaowen and Fleming, Charles and Lu, Chris Xiaoxuan},
  booktitle={European Conference on Computer Vision},
  pages={301--318},
  year={2024},
  organization={Springer}
}

@article{yao2025efficient,
  title={Efficient GPT-4V level multimodal large language model for deployment on edge devices},
  author={Yao, Yuan and Yu, Tianyu and Zhang, Ao and Wang, Chongyi and Cui, Junbo and Zhu, Hongji and Cai, Tianchi and Chen, Chi and Li, Haoyu and Zhao, Weilin and others},
  journal={Nature Communications},
  volume={16},
  number={1},
  pages={5509},
  year={2025},
  publisher={Nature Publishing Group UK London}
}

@article{gilbert2023large,
  title={Large language model AI chatbots require approval as medical devices},
  author={Gilbert, Stephen and Harvey, Hugh and Melvin, Tom and Vollebregt, Erik and Wicks, Paul},
  journal={Nature Medicine},
  volume={29},
  number={10},
  pages={2396--2398},
  year={2023},
  publisher={Nature Publishing Group US New York}
}

@inproceedings{ma2024one,
  title={From one thousand pages of specification to unveiling hidden bugs: Large language model assisted fuzzing of matter $\{IoT\}$ devices},
  author={Ma, Xiaoyue and Luo, Lannan and Zeng, Qiang},
  booktitle={33rd USENIX Security Symposium (USENIX Security 24)},
  pages={4783--4800},
  year={2024}
}

@article{cai2022enable,
  title={Enable deep learning on mobile devices: Methods, systems, and applications},
  author={Cai, Han and Lin, Ji and Lin, Yujun and Liu, Zhijian and Tang, Haotian and Wang, Hanrui and Zhu, Ligeng and Han, Song},
  journal={ACM Transactions on Design Automation of Electronic Systems (TODAES)},
  volume={27},
  number={3},
  pages={1--50},
  year={2022},
  publisher={ACM New York, NY}
}

@inproceedings{alizadeh2024llm,
  title={Llm in a flash: Efficient large language model inference with limited memory},
  author={Alizadeh, Keivan and Mirzadeh, Seyed Iman and Belenko, Dmitry and Khatamifard, S and Cho, Minsik and Del Mundo, Carlo C and Rastegari, Mohammad and Farajtabar, Mehrdad},
  booktitle={Proceedings of the 62nd Annual Meeting of the Association for Computational Linguistics (Volume 1: Long Papers)},
  pages={12562--12584},
  year={2024}
}

@inproceedings{edalati2022kronecker,
  title={Kronecker Decomposition for GPT Compression},
  author={Edalati, Ali and Tahaei, Marzieh and Rashid, Ahmad and Nia, Vahid and Clark, James and Rezagholizadeh, Mehdi},
  booktitle={Proceedings of the 60th Annual Meeting of the Association for Computational Linguistics (Volume 2: Short Papers)},
  pages={219--226},
  year={2022}
}

@article{bai2022towards,
  title={Towards efficient post-training quantization of pre-trained language models},
  author={Bai, Haoli and Hou, Lu and Shang, Lifeng and Jiang, Xin and King, Irwin and Lyu, Michael R},
  journal={Advances in neural information processing systems},
  volume={35},
  pages={1405--1418},
  year={2022}
}

@inproceedings{guan2024aptq,
  title={Aptq: Attention-aware post-training mixed-precision quantization for large language models},
  author={Guan, Ziyi and Huang, Hantao and Su, Yupeng and Huang, Hong and Wong, Ngai and Yu, Hao},
  booktitle={Proceedings of the 61st ACM/IEEE Design Automation Conference},
  pages={1--6},
  year={2024}
}

@inproceedings{jeon2023frustratingly,
  title={A frustratingly easy post-training quantization scheme for llms},
  author={Jeon, Yongkweon and Lee, Chungman and Park, Kyungphil and Kim, Ho-young},
  booktitle={Proceedings of the 2023 Conference on Empirical Methods in Natural Language Processing},
  pages={14446--14461},
  year={2023}
}

@inproceedings{oliinyk2024fuzzing,
  title={Fuzzing $\{BusyBox\}$: Leveraging $\{LLM\}$ and Crash Reuse for Embedded Bug Unearthing},
  author={Oliinyk, Yaroslav and Scott, Michael and Tsang, Ryan and Fang, Chongzhou and Homayoun, Houman and others},
  booktitle={33rd USENIX Security Symposium (USENIX Security 24)},
  pages={883--900},
  year={2024}
}

@inproceedings{wei2025t,
  title={T-mac: Cpu renaissance via table lookup for low-bit llm deployment on edge},
  author={Wei, Jianyu and Cao, Shijie and Cao, Ting and Ma, Lingxiao and Wang, Lei and Zhang, Yanyong and Yang, Mao},
  booktitle={Proceedings of the Twentieth European Conference on Computer Systems},
  pages={278--292},
  year={2025}
}

@inproceedings{kong2024swapmoe,
  title={SwapMoE: Serving Off-the-shelf MoE-based Large Language Models with Tunable Memory Budget},
  author={Kong, Rui and Li, Yuanchun and Feng, Qingtian and Wang, Weijun and Ye, Xiaozhou and Ouyang, Ye and Kong, Linghe and Liu, Yunxin},
  booktitle={Proceedings of the 62nd Annual Meeting of the Association for Computational Linguistics (Volume 1: Long Papers)},
  pages={6710--6720},
  year={2024}
}

@article{chen2024understanding,
  title={Understanding the potential of fpga-based spatial acceleration for large language model inference},
  author={Chen, Hongzheng and Zhang, Jiahao and Du, Yixiao and Xiang, Shaojie and Yue, Zichao and Zhang, Niansong and Cai, Yaohui and Zhang, Zhiru},
  journal={ACM Transactions on Reconfigurable Technology and Systems},
  volume={18},
  number={1},
  pages={1--29},
  year={2024},
  publisher={ACM New York, NY}
}

@inproceedings{li2025pushing,
  title={Pushing up to the limit of memory bandwidth and capacity utilization for efficient llm decoding on embedded fpga},
  author={Li, Jindong and Li, Tenglong and Shen, Guobin and Zhao, Dongcheng and Zhang, Qian and Zeng, Yi},
  booktitle={2025 Design, Automation \& Test in Europe Conference (DATE)},
  pages={1--7},
  year={2025},
  organization={IEEE}
}

@article{hu2024realizing,
  title={Realizing Efficient On-Device Language-Based Image Retrieval},
  author={Hu, Zhiming and Kemertas, Mete and Xiao, Lan and Phillips, Caleb and Mohomed, Iqbal and Fazly, Afsaneh},
  journal={ACM Transactions on Multimedia Computing, Communications and Applications},
  volume={20},
  number={9},
  pages={1--18},
  year={2024},
  publisher={ACM New York, NY}
}

@inproceedings{xu2024towards,
  title={Towards energy-efficient llama2 architecture on embedded fpgas},
  author={Xu, Han and Wang, Xingyuan and Ji, Shihao},
  booktitle={Proceedings of the 33rd ACM International Conference on Information and Knowledge Management},
  pages={5570--5571},
  year={2024}
}

@article{zhang2024edgeshard,
  title={Edgeshard: Efficient llm inference via collaborative edge computing},
  author={Zhang, Mingjin and Shen, Xiaoming and Cao, Jiannong and Cui, Zeyang and Jiang, Shan},
  journal={IEEE Internet of Things Journal},
  year={2024},
  publisher={IEEE}
}

@inproceedings{xu2025fast,
  title={Fast on-device LLM inference with npus},
  author={Xu, Daliang and Zhang, Hao and Yang, Liming and Liu, Ruiqi and Huang, Gang and Xu, Mengwei and Liu, Xuanzhe},
  booktitle={Proceedings of the 30th ACM International Conference on Architectural Support for Programming Languages and Operating Systems, Volume 1},
  pages={445--462},
  year={2025}
}

@inproceedings{yu2024cambricon,
  title={Cambricon-llm: A chiplet-based hybrid architecture for on-device inference of 70b llm},
  author={Yu, Zhongkai and Liang, Shengwen and Ma, Tianyun and Cai, Yunke and Nan, Ziyuan and Huang, Di and Song, Xinkai and Hao, Yifan and Zhang, Jie and Zhi, Tian and others},
  booktitle={2024 57th IEEE/ACM International Symposium on Microarchitecture (MICRO)},
  pages={1474--1488},
  year={2024},
  organization={IEEE}
}

@inproceedings{seo2025facil,
  title={FACIL: Flexible DRAM Address Mapping for SoC-PIM Cooperative On-device LLM Inference},
  author={Seo, Seong Hoon and Kim, Junghoon and Lee, Donghyun and Yoo, Seonah and Moon, Seokwon and Park, Yeonhong and Lee, Jae W},
  booktitle={2025 IEEE International Symposium on High Performance Computer Architecture (HPCA)},
  pages={1720--1733},
  year={2025},
  organization={IEEE}
}

@inproceedings{liu2025ops,
  title={OPS: Outlier-Aware Precision-Slice Framework for LLM Acceleration},
  author={Liu, Fangxin and Yang, Ning and Wang, Zongwu and Zhu, Xuanpeng and Yao, Haidong and Xiong, Xiankui and Sun, Qi and Jiang, Li},
  booktitle={2025 Design, Automation \& Test in Europe Conference (DATE)},
  pages={1--2},
  year={2025},
  organization={IEEE}
}

@article{ren2024industrial,
  title={Industrial internet of things with large language models (LLMs): an intelligence-based reinforcement learning approach},
  author={Ren, Yuzheng and Zhang, Haijun and Yu, F Richard and Li, Wei and Zhao, Pincan and He, Ying},
  journal={IEEE Transactions on Mobile Computing},
  year={2024},
  publisher={IEEE}
}

@article{zhang2025vavlm,
  title={VaVLM: Toward Efficient Edge-Cloud Video Analytics With Vision-Language Models},
  author={Zhang, Yang and Wang, Hanling and Bai, Qing and Liang, Haifeng and Zhu, Peican and Muntean, Gabriel-Miro and Li, Qing},
  journal={IEEE Transactions on Broadcasting},
  year={2025},
  publisher={IEEE}
}

@inproceedings{sun2025lincoln,
  title={Lincoln: Real-Time 50\~{} 100B LLM Inference on Consumer Devices with LPDDR-Interfaced, Compute-Enabled Flash Memory},
  author={Sun, Weiyi and Gao, Mingyu and Li, Zhaoshi and Zhang, Aoyang and Chou, Iris Ying and Zhu, Jianfeng and Wei, Shaojun and Liu, Leibo},
  booktitle={2025 IEEE International Symposium on High Performance Computer Architecture (HPCA)},
  pages={1734--1750},
  year={2025},
  organization={IEEE}
}

@inproceedings{lee2025paise,
  title={PAISE: PIM-Accelerated Inference Scheduling Engine for Transformer-based LLM},
  author={Lee, Hyojung and Baek, Daehyeon and Son, Jimyoung and Choi, Jieun and Moon, Kihyo and Jang, Minsung},
  booktitle={2025 IEEE International Symposium on High Performance Computer Architecture (HPCA)},
  pages={1707--1719},
  year={2025},
  organization={IEEE}
}

@article{rivkin2024aiot,
  title={Aiot smart home via autonomous llm agents},
  author={Rivkin, Dmitriy and Hogan, Francois and Feriani, Amal and Konar, Abhisek and Sigal, Adam and Liu, Xue and Dudek, Gregory},
  journal={IEEE Internet of Things Journal},
  year={2024},
  publisher={IEEE}
}

@article{yang2025quality,
  title={Quality-of-Service Aware LLM Routing for Edge Computing with Multiple Experts},
  author={Yang, Jin and Wu, Qiong and Feng, Zhiying and Zhou, Zhi and Guo, Deke and Chen, Xu},
  journal={IEEE Transactions on Mobile Computing},
  number={99},
  pages={1--15},
  year={2025},
  publisher={IEEE}
}

@inproceedings{chen2024autoos,
  title={Autoos: make your os more powerful by exploiting large language models},
  author={Chen, Huilai and Wen, Yuanbo and Cheng, Limin and Kuang, Shouxu and Liu, Yumeng and Li, Weijia and Li, Ling and Zhang, Rui and Song, Xinkai and Li, Wei and others},
  booktitle={Forty-first International Conference on Machine Learning},
  year={2024}
}

@article{cai2025prompt,
  title={Prompt-Ladder: Memory-efficient prompt tuning for vision-language models on edge devices},
  author={Cai, Siqi and Liu, Xuan and Yuan, Jingling and Zhou, Qihua},
  journal={Pattern Recognition},
  volume={163},
  pages={111460},
  year={2025},
  publisher={Elsevier}
}

@article{yao2025minicpm,
  title={Efficient GPT-4V level multimodal large language model for deployment on edge devices},
  author={Yao, Yuan and Yu, Tianyu and Zhang, Ao and Wang, Chongyi and Cui, Junbo and Zhu, Hongji and Cai, Tianchi and Chen, Chi and Li, Haoyu and Zhao, Weilin and others},
  journal={Nature Communications},
  volume={16},
  number={1},
  pages={5509},
  year={2025},
  publisher={Nature Publishing Group UK London}
}

@misc{minicpm-v4,
  title={MiniCPM-V: A GPT-4V Level MLLM on Your Phone},
  author={Yao, Yuan and Yu, Tianyu and Zhang, Ao and Wang, Chongyi and Cui, Junbo and Zhu, Hongji and Cai, Tianchi and Li, Haoyu and Zhao, Weilin and He, Zhihui and others},
  journal={arXiv preprint arXiv:2408.01800},
  year={2024}
}

@misc{rockchip-rk3588,
  title={RK3588 - Rockchip Flagship SoC for AIoT Applications},
  author={Rockchip Electronics Co., Ltd.},
  year={2024},
  howpublished={\url{https://www.rock-chips.com/a/cn/product/RK35xilie/2022/0926/1656.html}},
  note={8nm process flagship chip with 6 TOPS NPU and 8K video capabilities}
}

@techreport{rockchip-annual-report,
  title={Rockchip 2024 Annual Report: AIoT SoC Market Leadership and Automotive Breakthrough},
  author={Rockchip Electronics Co., Ltd.},
  year={2024},
  institution={Rockchip Electronics},
  note={Comprehensive analysis of financial performance and strategic positioning in AIoT and automotive markets}
}

@misc{rk182x-ai-coprocessor,
  title={RK182X Series: Edge AI Co-processors for Local LLM Deployment},
  author={Rockchip Electronics Co., Ltd.},
  year={2024},
  howpublished={\url{https://www.rock-chips.com}},
  note={Dedicated AI co-processors enabling efficient on-device large language model inference}
}

@misc{arm-cortex,
  title={Arm Cortex-A Series Programmer's Guide / Performance Brief},
  author={Arm Ltd.},
  year={2023},
  url={https://developer.arm.com/documentation/den0013/0400/Preface}
}

@article{wang2022self,
  title={A self-adaptive weighted differential evolution approach for large-scale feature selection},
  author={Wang, Xubin and Wang, Yunhe and Wong, Ka-Chun and Li, Xiangtao},
  journal={Knowledge-Based Systems},
  volume={235},
  pages={107633},
  year={2022},
  publisher={Elsevier}
}

@article{wang2024mel,
  title={MEL: Efficient multi-task evolutionary learning for high-dimensional feature selection},
  author={Wang, Xubin and Shangguan, Haojiong and Huang, Fengyi and Wu, Shangrui and Jia, Weijia},
  journal={IEEE Transactions on Knowledge and Data Engineering},
  volume={36},
  number={8},
  pages={4020--4033},
  year={2024},
  publisher={IEEE}
}

@inproceedings{wangdemonstration,
  title={Demonstration Selection for In-Context Learning via Reinforcement Learning},
  author={Wang, Xubin and Wu, Jianfei and Yichen, Yuan and Cai, Deyu and Li, Mingzhe and Jia, Weijia},
  booktitle={Forty-second International Conference on Machine Learning},
  year={2025}
}

@inproceedings{wang2022feature,
  title={A feature weighting particle swarm optimization method to identify biomarker genes},
  author={Wang, Xubin and Jia, Weijia},
  booktitle={2022 IEEE International Conference on Bioinformatics and Biomedicine (BIBM)},
  pages={830--834},
  year={2022},
  organization={IEEE}
}

@article{wang2024exhaustive,
  title={Exhaustive Exploitation of Nature-Inspired Computation for Cancer Screening in an Ensemble Manner},
  author={Wang, Xubin and Wang, Yunhe and Ma, Zhiqiang and Wong, Ka-Chun and Li, Xiangtao},
  journal={IEEE/ACM Transactions on Computational Biology and Bioinformatics},
  volume={21},
  number={5},
  pages={1366--1379},
  year={2024},
  publisher={IEEE}
}

@article{vaswani2017attention,
  title={Attention is all you need},
  author={Vaswani, Ashish and Shazeer, Noam and Parmar, Niki and Uszkoreit, Jakob and Jones, Llion and Gomez, Aidan N and Kaiser, {\L}ukasz and Polosukhin, Illia},
  journal={Advances in neural information processing systems},
  volume={30},
  year={2017}
}

@article{brown2020language,
  title={Language Models are Few-Shot Learners},
  author={Brown, Tom and Mann, Benjamin and Ryder, Nick and Subbiah, Melanie and Kaplan, Jared D and Dhariwal, Prafulla and Neelakantan, Arvind and Shyam, Pranav and Sastry, Girish and Askell, Amanda and others},
  journal={Advances in Neural Information Processing Systems},
  volume={33},
  pages={1877--1901},
  year={2020}
}

@article{achiam2023gpt,
  title={GPT-4 Technical Report},
  author={Achiam, Josh and Adler, Steven and Agarwal, Sandhini and Ahmad, Lama and Akkaya, Ilge and Aleman, Florencia Leoni and Almeida, Diogo and Altenschmidt, Janko and Altman, Sam and Anadkat, Shyamal and others},
  journal={arXiv preprint arXiv:2303.08774},
  year={2023}
}

@article{chowdhery2022palm,
  title={Palm: Scaling language modeling with pathways},
  author={Chowdhery, Aakanksha and Narang, Sharan and Devlin, Jacob and Bosma, Maarten and Mishra, Gaurav and Roberts, Adam and Barham, Paul and Chung, Hyung Won and Sutton, Charles and Gehrmann, Sebastian and others},
  journal={Journal of Machine Learning Research},
  volume={24},
  number={240},
  pages={1--113},
  year={2023}
}

@article{touvron2023llama,
  title={Llama 2: Open Foundation and Fine-Tuned Chat Models},
  author={Touvron, Hugo and Martin, Louis and Stone, Kevin and Albert, Peter and Almahairi, Amjad and Babaei, Yasmine and Bashlykov, Nikolay and Batra, Soumya and Bhargava, Prajjwal and Bhosale, Shruti and others},
  journal={arXiv preprint arXiv:2307.09288},
  year={2023}
}

@article{wei2022emergent,
  title={Emergent Abilities of Large Language Models},
  author={Wei, Jason and Tay, Yi and Bommasani, Rishi and Raffel, Colin and Zoph, Barret and Borgeaud, Sebastian and Yogatama, Dani and Bosma, Maarten and Zhou, Denny and Metzler, Donald and others},
  journal={Transactions on Machine Learning Research},
  year={2022}
}

@article{shi2016edge,
  title={Edge Computing: Vision and Challenges},
  author={Shi, Weisong and Cao, Jie and Zhang, Quan and Li, Youhuizi and Xu, Lanyu},
  journal={IEEE Internet of Things Journal},
  volume={3},
  number={5},
  pages={637--646},
  year={2016},
  publisher={IEEE}
}

@article{zhou2019edge,
  title={Edge Intelligence: Paving the Last Mile of Artificial Intelligence with Edge Computing},
  author={Zhou, Zhi and Chen, Xu and Li, En and Zeng, Liekang and Luo, Ke and Zhang, Junshan},
  journal={Proceedings of the IEEE},
  volume={107},
  number={8},
  pages={1738--1762},
  year={2019},
  publisher={IEEE}
}

@article{deng2020edge,
  title={Edge intelligence: The confluence of edge computing and artificial intelligence},
  author={Deng, Shuiguang and Zhao, Hailiang and Fang, Weijia and Yin, Jianwei and Dustdar, Schahram and Zomaya, Albert Y},
  journal={IEEE Internet of Things Journal},
  volume={7},
  number={8},
  pages={7457--7469},
  year={2020},
  publisher={IEEE}
}

@book{russell2016artificial,
  title={Artificial Intelligence: A Modern Approach},
  author={Russell, Stuart J and Norvig, Peter},
  year={2020},
  publisher={Pearson},
  edition={4th}
}

@article{stone2000multiagent,
  title={Multiagent Systems: A Survey from a Machine Learning Perspective},
  author={Stone, Peter and Veloso, Manuela},
  journal={Autonomous Robots},
  volume={8},
  number={3},
  pages={345--383},
  year={2000},
  publisher={Springer}
}

@article{dorri2018multi,
  title={Multi-agent systems: A survey},
  author={Dorri, Ali and Kanhere, Salil S and Jurdak, Raja},
  journal={IEEE Access},
  volume={6},
  pages={28573--28593},
  year={2018},
  publisher={IEEE}
}

@article{wang2025empowering,
  title={Empowering edge intelligence: A comprehensive survey on on-device ai models},
  author={Wang, Xubin and Tang, Zhiqing and Guo, Jianxiong and Meng, Tianhui and Wang, Chenhao and Wang, Tian and Jia, Weijia},
  journal={ACM Computing Surveys},
  volume={57},
  number={9},
  pages={1--39},
  year={2025},
  publisher={ACM New York, NY}
}

@article{wang2025optimizing,
  title={Optimizing Edge AI: A Comprehensive Survey on Data, Model, and System Strategies},
  author={Wang, Xubin and Jia, Weijia},
  journal={arXiv preprint arXiv:2501.03265},
  year={2025}
}

@inproceedings{sun2020mobilebert,
  title={MobileBERT: A Compact Task-Agnostic BERT for Resource-Limited Devices},
  author={Sun, Zhiqing and Yu, Hongkun and Song, Xiaodan and Liu, Renjie and Yang, Yiming and Zhou, Denny},
  booktitle={Proceedings of the AAAI Conference on Artificial Intelligence},
  volume={34},
  number={07},
  pages={8965--8972},
  year={2020}
}

@article{sanh2019distilbert,
  title={DistilBERT, a distilled version of BERT: smaller, faster, cheaper and lighter},
  author={Sanh, Victor and Debut, Lysandre and Chaumond, Julien and Wolf, Thomas},
  journal={arXiv preprint arXiv:1910.01108},
  year={2019}
}

@inproceedings{jiao2020tinybert,
  title={TinyBERT: Distilling BERT for Natural Language Understanding},
  author={Jiao, Xiaoqi and Yin, Yichun and Shang, Lifeng and Jiang, Xin and Chen, Xiao and Li, Linlin and Wang, Fang and Liu, Qun},
  booktitle={Proceedings of the 2020 Conference on Empirical Methods in Natural Language Processing (EMNLP)},
  pages={4163--4174},
  year={2020}
}

@misc{llama3.1_8b_meta,
  title={Llama 3.1 8B},
  author={Meta AI},
  year={2024},
  note={Available at: \url{https://llama.meta.com/}}
}

@inproceedings{liu2018darts,
  title={DARTS: Differentiable Architecture Search},
  author={Liu, Hanxiao and Simonyan, Karen and Yang, Yiming},
  booktitle={International Conference on Learning Representations},
  year={2019}
}

@inproceedings{han2016deep,
  title={Deep Compression: Compressing Deep Neural Network with Pruning, Trained Quantization and Huffman Coding},
  author={Han, Song and Mao, Huizi and Dally, William J},
  booktitle={ICLR},
  year={2016}
}

@article{ma2023llm,
  title={Llm-pruner: On the structural pruning of large language models},
  author={Ma, Xinyin and Fang, Gongfan and Wang, Xinchao},
  journal={Advances in neural information processing systems},
  volume={36},
  pages={21702--21720},
  year={2023}
}

@inproceedings{frantar2023sparsegpt,
  title={Sparsegpt: Massive language models can be accurately pruned in one-shot},
  author={Frantar, Elias and Alistarh, Dan},
  booktitle={International conference on machine learning},
  pages={10323--10337},
  year={2023},
  organization={PMLR}
}

@inproceedings{lanalbert,
  title={ALBERT: A Lite BERT for Self-supervised Learning of Language Representations},
  author={Lan, Zhenzhong and Chen, Mingda and Goodman, Sebastian and Gimpel, Kevin and Sharma, Piyush and Soricut, Radu},
  booktitle={International Conference on Learning Representations},
  year={2020}
}

@inproceedings{liu2024mobilellm,
  title={Mobilellm: Optimizing sub-billion parameter language models for on-device use cases},
  author={Liu, Zechun and Zhao, Changsheng and Iandola, Forrest and Lai, Chen and Tian, Yuandong and Fedorov, Igor and Xiong, Yunyang and Chang, Ernie and Shi, Yangyang and Krishnamoorthi, Raghuraman and others},
  booktitle={Forty-first International Conference on Machine Learning},
  year={2024}
}

@article{qlora2023,
  title={Qlora: Efficient finetuning of quantized llms},
  author={Dettmers, Tim and Pagnoni, Artidoro and Holtzman, Ari and Zettlemoyer, Luke},
  journal={Advances in neural information processing systems},
  volume={36},
  pages={10088--10115},
  year={2023}
}

@inproceedings{jacob2018quantization,
  title={Quantization and training of neural networks for efficient integer-arithmetic-only inference},
  author={Jacob, Benoit and Kligys, Skirmantas and Chen, Bo and Zhu, Menglong and Tang, Matthew and Howard, Andrew and Adam, Hartwig and Kalenichenko, Dmitry},
  booktitle={Proceedings of the IEEE conference on computer vision and pattern recognition},
  pages={2704--2713},
  year={2018}
}

@article{nagel2021white,
  title={A White Paper on Neural Network Quantization},
  author={Nagel, Markus and Fournarakis, Marios and Amjad, Rana Ali and Bondarenko, Yelysei and van Baalen, Mart and Blankevoort, Tijmen},
  journal={arXiv preprint arXiv:2106.08295},
  year={2021}
}

@incollection{gholami2022survey,
  title={A survey of quantization methods for efficient neural network inference},
  author={Gholami, Amir and Kim, Sehoon and Dong, Zhen and Yao, Zhewei and Mahoney, Michael W and Keutzer, Kurt},
  booktitle={Low-power computer vision},
  pages={291--326},
  year={2022},
  publisher={Chapman and Hall/CRC}
}

@article{hinton2015distilling,
  title={Distilling the knowledge in a neural network},
  author={Hinton, Geoffrey and Vinyals, Oriol and Dean, Jeff},
  journal={arXiv preprint arXiv:1503.02531},
  year={2015}
}

@article{gou2021knowledge,
  title={Knowledge distillation: A survey},
  author={Gou, Jianping and Yu, Baosheng and Maybank, Stephen J and Tao, Dacheng},
  journal={International journal of computer vision},
  volume={129},
  number={6},
  pages={1789--1819},
  year={2021},
  publisher={Springer}
}

@article{zhang2024tinyllama,
  title={Tinyllama: An open-source small language model},
  author={Zhang, Peiyuan and Zeng, Guangtao and Wang, Tianduo and Lu, Wei},
  journal={arXiv preprint arXiv:2401.02385},
  year={2024}
}

@article{huang2024efficient,
  title={Efficient multi-modal large language models via visual token grouping},
  author={Huang, Minbin and Huang, Runhui and Shi, Han and Chen, Yimeng and Zheng, Chuanyang and Sun, Xiangguo and Jiang, Xin and Li, Zhenguo and Cheng, Hong},
  journal={arXiv preprint arXiv:2411.17773},
  year={2024}
}

@article{dwork2014algorithmic,
  title={The algorithmic foundations of differential privacy},
  author={Dwork, Cynthia and Roth, Aaron and others},
  journal={Foundations and trends{\textregistered} in theoretical computer science},
  volume={9},
  number={3--4},
  pages={211--407},
  year={2014},
  publisher={Now Publishers, Inc.}
}

@article{salah2019blockchain,
  title={Blockchain for AI: Review and open research challenges},
  author={Salah, Khaled and Rehman, M Habib Ur and Nizamuddin, Nishara and Al-Fuqaha, Ala},
  journal={IEEE access},
  volume={7},
  pages={10127--10149},
  year={2019},
  publisher={IEEE}
}

@article{lu2024federated,
  title={Federated learning with non-iid data: A survey},
  author={Lu, Zili and Pan, Heng and Dai, Yueyue and Si, Xueming and Zhang, Yan},
  journal={IEEE Internet of Things Journal},
  volume={11},
  number={11},
  pages={19188--19209},
  year={2024},
  publisher={IEEE}
}

@misc{iso9126_usability,
  title={ISO/IEC 9126-1:2001 Software engineering -- Product quality -- Part 1: Quality model},
  author={ISO/IEC},
  year={2001}
}

@article{tian2025clone,
  title={CLONE: Customizing LLMs for Efficient Latency-Aware Inference at the Edge},
  author={Tian, Chunlin and Qin, Xinpeng and Tam, Kahou and Li, Li and Wang, Zijian and Zhao, Yuanzhe and Zhang, Minglei and Xu, Chengzhong},
  journal={arXiv preprint arXiv:2506.02847},
  year={2025}
}

@article{jin2024collm,
  title={Ce-collm: Efficient and adaptive large language models through cloud-edge collaboration},
  author={Jin, Hongpeng and Wu, Yanzhao},
  journal={arXiv preprint arXiv:2411.02829},
  year={2024}
}

@article{sakib2025small,
  title={Small Language Models: Architectures, Techniques, Evaluation, Problems and Future Adaptation},
  author={Sakib, Tanjil Hasan and Hosain, Md Tanzib and Morol, Md Kishor},
  journal={arXiv preprint arXiv:2505.19529},
  year={2025}
}

@misc{slm_aisera,
  title={What are Small Language Models (SLMs)?},
  author={Aisera},
  year={2025},
  url={https://aisera.com/blog/small-language-models/},
  note={Accessed: June 19, 2025}
}

@article{xu2024survey,
  title={A survey on knowledge distillation of large language models},
  author={Xu, Xiaohan and Li, Ming and Tao, Chongyang and Shen, Tao and Cheng, Reynold and Li, Jinyang and Xu, Can and Tao, Dacheng and Zhou, Tianyi},
  journal={arXiv preprint arXiv:2402.13116},
  year={2024}
}

@misc{synthetic_data_betterdata,
  title={Data Augmentation with Synthetic Data for AI and ML},
  author={Betterdata},
  year={2025},
  url={https://www.betterdata.ai/blogs/data-augmentation-with-synthetic-data-for-ai-and-ml},
  note={Accessed: June 19, 2025}
}

@misc{synthetic_data_ais,
  title={Knowledge Systems and Synthetic Data: The Role of Generative AI in Data Augmentation},
  author={AIS},
  year={2025},
  url={https://www.ais.com/knowledge-systems-and-synthetic-data/},
  note={Accessed: June 19, 2025}
}

@article{gupta2017neural,
  title={Neural Architecture Search for AI Model Optimization},
  author={Gupta, Siddharth},
  journal={International Journal of Artificial Intelligence and Machine Learning},
  volume={4},
  number={2},
  year={2017}
}

@article{meng2024evolution,
  title={Evolution and efficiency in neural architecture search: Bridging the gap between expert design and automated optimization},
  author={Meng, Fanfei and Wang, Chen-Ao and Zhang, Lele},
  journal={arXiv preprint arXiv:2403.17012},
  year={2024}
}

@misc{llm_pruning_nvidia,
  title={LLM Model Pruning and Knowledge Distillation with NVIDIA NeMo Framework},
  author={NVIDIA},
  year={2025},
  howpublished={https://developer.nvidia.com/blog/llm-model-pruning-and-knowledge-distillation-with-nvidia-nemo-framework/},
  note={Accessed: June 19, 2025}
}

@misc{quantization_symbl,
  title={A Guide to Quantization in LLMs},
  author={Symbl.ai},
  year={2025},
  url={https://symbl.ai/developers/blog/a-guide-to-quantization-in-llms/},
  note={Accessed: June 19, 2025}
}

@inproceedings{librecq,
  title={BRECQ: Pushing the Limit of Post-Training Quantization by Block Reconstruction},
  author={Li, Yuhang and Gong, Ruihao and Tan, Xu and Yang, Yang and Hu, Peng and Zhang, Qi and Yu, Fengwei and Wang, Wei and Gu, Shi},
  booktitle={International Conference on Learning Representations},
  year={2021}
}

@misc{ollama_walturn,
  title={What is Ollama? Features, Pricing, and Use Cases},
  author={Walturn},
  year={2025},
  url={https://www.walturn.com/insights/what-is-ollama-features-pricing-and-use-cases},
  note={Accessed: June 19, 2025}
}

@misc{mlc_llm,
  title={MLC LLM | Home},
  author={MLC LLM},
  year={2025},
  url={https://llm.mlc.ai/},
  note={Accessed: June 19, 2025}
}

@misc{iso27001_ttms,
  title={ISO 27001 Implementation - Improve Data Security},
  author={TTMS},
  year={2025},
  url={https://ttms.com/iso-27001-implementation-strengthen-data-security-in-your-company/},
  note={Accessed: June 19, 2025}
}

@misc{ai_security_hpe,
  title={What is AI Security | Glossary},
  author={HPE},
  year={2025},
  url={https://www.hpe.com/us/en/what-is/ai-security.html},
  note={Accessed: June 19, 2025}
}

@misc{model_partitioning_macsphere,
  title={Deep Learning on the Edge: Model Partitioning, Caching, and Compression},
  author={Fang, Yihao},
  howpublished={McMaster University Institutional Repository (MacSphere)},
  year={2025},
  url={https://macsphere.mcmaster.ca/handle/11375/25576},
  note={Accessed: June 19, 2025}
}

@article{amp4ec,
  title={AMP4EC: Adaptive Model Partitioning Framework for Efficient Deep Learning Inference in Edge Computing Environments},
  author={Zhang, Guilin and Guo, Wulan and Tan, Ziqi and Jiang, Hailong},
  journal={arXiv preprint arXiv:2504.00407},
  year={2025}
}

@article{schuler2015neuromorphic,
  title={Neuromorphic computing: From materials to systems architecture},
  author={Schuler, I and Stevens, Rick},
  journal={Office of Scientific and Technical Information},
  year={2015}
}

@misc{neuromorphic_atos,
  title={Neuromorphic computing: The future of AI and beyond},
  author={Atos},
  year={2025},
  url={https://atos.net/en/blog/neuromorphic-computing-the-future-of-ai-and-beyond},
  note={Accessed: June 19, 2025}
}

@article{gao2025survey,
  title={A survey of self-evolving agents: On path to artificial super intelligence},
  author={Gao, Huan-ang and Geng, Jiayi and Hua, Wenyue and Hu, Mengkang and Juan, Xinzhe and Liu, Hongzhang and Liu, Shilong and Qiu, Jiahao and Qi, Xuan and Wu, Yiran and others},
  journal={arXiv preprint arXiv:2507.21046},
  year={2025}
}

@article{abouelenin2025phi,
  title={Phi-4-mini technical report: Compact yet powerful multimodal language models via mixture-of-loras},
  author={Abouelenin, Abdelrahman and Ashfaq, Atabak and Atkinson, Adam and Awadalla, Hany and Bach, Nguyen and Bao, Jianmin and Benhaim, Alon and Cai, Martin and Chaudhary, Vishrav and Chen, Congcong and others},
  journal={arXiv preprint arXiv:2503.01743},
  year={2025}
}

@article{peng2025lmm,
  title={Lmm-r1: Empowering 3b lmms with strong reasoning abilities through two-stage rule-based rl},
  author={Peng, Yingzhe and Zhang, Gongrui and Zhang, Miaosen and You, Zhiyuan and Liu, Jie and Zhu, Qipeng and Yang, Kai and Xu, Xingzhong and Geng, Xin and Yang, Xu},
  journal={arXiv preprint arXiv:2503.07536},
  year={2025}
}

@article{dai2020edge,
  title={Edge intelligence for energy-efficient computation offloading and resource allocation in 5G beyond},
  author={Dai, Yueyue and Zhang, Ke and Maharjan, Sabita and Zhang, Yan},
  journal={IEEE Transactions on Vehicular Technology},
  volume={69},
  number={10},
  pages={12175--12186},
  year={2020},
  publisher={IEEE}
}

@article{chen2019deep,
  title={Deep learning with edge computing: A review},
  author={Chen, Jiasi and Ran, Xukan},
  journal={Proceedings of the IEEE},
  volume={107},
  number={8},
  pages={1655--1674},
  year={2019},
  publisher={IEEE}
}

@article{liu2019survey,
  title={A survey on edge computing systems and tools},
  author={Liu, Fang and Tang, Guoming and Li, Youhuizi and Cai, Zhiping and Zhang, Xingzhou and Zhou, Tongqing},
  journal={Proceedings of the IEEE},
  volume={107},
  number={8},
  pages={1537--1562},
  year={2019},
  publisher={IEEE}
}

@inproceedings{tan2019mnasnet,
  title={Mnasnet: Platform-aware neural architecture search for mobile},
  author={Tan, Mingxing and Chen, Bo and Pang, Ruoming and Vasudevan, Vijay and Sandler, Mark and Howard, Andrew and Le, Quoc V},
  booktitle={Proceedings of the IEEE/CVF conference on computer vision and pattern recognition},
  pages={2820--2828},
  year={2019}
}

@inproceedings{wu2019fbnet,
  title={Fbnet: Hardware-aware efficient convnet design via differentiable neural architecture search},
  author={Wu, Bichen and Dai, Xiaoliang and Zhang, Peizhao and Wang, Yanghan and Sun, Fei and Wu, Yiming and Tian, Yuandong and Vajda, Peter and Jia, Yangqing and Keutzer, Kurt},
  booktitle={Proceedings of the IEEE/CVF conference on computer vision and pattern recognition},
  pages={10734--10742},
  year={2019}
}

@inproceedings{boo2021stochastic,
  title={Stochastic precision ensemble: self-knowledge distillation for quantized deep neural networks},
  author={Boo, Yoonho and Shin, Sungho and Choi, Jungwook and Sung, Wonyong},
  booktitle={Proceedings of the AAAI Conference on Artificial Intelligence},
  volume={35},
  number={8},
  pages={6794--6802},
  year={2021}
}

@article{cui2022bits,
  title={Bits-ensemble: Toward light-weight robust deep ensemble by bits-sharing},
  author={Cui, Yufei and Wu, Shangyu and Li, Qiao and Chan, Antoni B and Kuo, Tei-Wei and Xue, Chun Jason},
  journal={IEEE Transactions on Computer-Aided Design of Integrated Circuits and Systems},
  volume={41},
  number={11},
  pages={4397--4408},
  year={2022},
  publisher={IEEE}
}

@inproceedings{wang2019haq,
  title={Haq: Hardware-aware automated quantization with mixed precision},
  author={Wang, Kuan and Liu, Zhijian and Lin, Yujun and Lin, Ji and Han, Song},
  booktitle={Proceedings of the IEEE/CVF conference on computer vision and pattern recognition},
  pages={8612--8620},
  year={2019}
}

@inproceedings{zhao2020sca,
  title={SCA: a secure CNN accelerator for both training and inference},
  author={Zhao, Lei and Zhang, Youtao and Yang, Jun},
  booktitle={2020 57th ACM/IEEE Design Automation Conference (DAC)},
  pages={1--6},
  year={2020},
  organization={IEEE}
}

@article{zhou2022energon,
  title={Energon: Toward efficient acceleration of transformers using dynamic sparse attention},
  author={Zhou, Zhe and Liu, Junlin and Gu, Zhenyu and Sun, Guangyu},
  journal={IEEE Transactions on Computer-Aided Design of Integrated Circuits and Systems},
  volume={42},
  number={1},
  pages={136--149},
  year={2022},
  publisher={IEEE}
}

@inproceedings{ignatov2021real,
  title={Real-time quantized image super-resolution on mobile npus, mobile ai 2021 challenge: Report},
  author={Ignatov, Andrey and Timofte, Radu and Denna, Maurizio and Younes, Abdel},
  booktitle={Proceedings of the IEEE/CVF conference on computer vision and pattern recognition},
  pages={2525--2534},
  year={2021}
}

@article{kouris2022fluid,
  title={Fluid batching: Exit-aware preemptive serving of early-exit neural networks on edge npus},
  author={Kouris, Alexandros and Venieris, Stylianos I and Laskaridis, Stefanos and Lane, Nicholas D},
  journal={arXiv preprint arXiv:2209.13443},
  year={2022}
}

@inproceedings{kwon2023efficient,
  title={Efficient memory management for large language model serving with pagedattention},
  author={Kwon, Woosuk and Li, Zhuohan and Zhuang, Siyuan and Sheng, Ying and Zheng, Lianmin and Yu, Cody Hao and Gonzalez, Joseph and Zhang, Hao and Stoica, Ion},
  booktitle={Proceedings of the 29th symposium on operating systems principles},
  pages={611--626},
  year={2023}
}

@article{zheng2024sglang,
  title={Sglang: Efficient execution of structured language model programs},
  author={Zheng, Lianmin and Yin, Liangsheng and Xie, Zhiqiang and Sun, Chuyue Livia and Huang, Jeff and Yu, Cody Hao and Cao, Shiyi and Kozyrakis, Christos and Stoica, Ion and Gonzalez, Joseph E and others},
  journal={Advances in neural information processing systems},
  volume={37},
  pages={62557--62583},
  year={2024}
}

@article{team2025minicpm4,
  title={Minicpm4: Ultra-efficient llms on end devices},
  author={Team, MiniCPM and Xiao, Chaojun and Li, Yuxuan and Han, Xu and Bai, Yuzhuo and Cai, Jie and Chen, Haotian and Chen, Wentong and Cong, Xin and Cui, Ganqu and others},
  journal={arXiv preprint arXiv:2506.07900},
  year={2025}
}

@inproceedings{yang2023processing,
  title={Processing-in-memory using optically-addressed phase change memory},
  author={Yang, Guowei and Demirkiran, Cansu and Kizilates, Zeynep Ece and Ocampo, Carlos A R{\'\i}os and Coskun, Ayse K and Joshi, Ajay},
  booktitle={2023 IEEE/ACM International Symposium on Low Power Electronics and Design (ISLPED)},
  pages={1--6},
  year={2023},
  organization={IEEE}
}

@article{kudithipudi2025neuromorphic,
  title={Neuromorphic computing at scale},
  author={Kudithipudi, Dhireesha and Schuman, Catherine and Vineyard, Craig M and Pandit, Tej and Merkel, Cory and Kubendran, Rajkumar and Aimone, James B and Orchard, Garrick and Mayr, Christian and Benosman, Ryad and others},
  journal={Nature},
  volume={637},
  number={8047},
  pages={801--812},
  year={2025},
  publisher={Nature Publishing Group UK London}
}

@article{alabed2025toast,
  title={TOAST: Fast and scalable auto-partitioning based on principled static analysis},
  author={Alabed, Sami and Grewe, Dominik and Rink, Norman Alexander and Sitdikov, Timur and Swietlik, Agnieszka and Vytiniotis, Dimitrios and Belov, Daniel},
  journal={arXiv preprint arXiv:2508.15010},
  year={2025}
}

@inproceedings{xiaoefficient,
  title={Efficient Streaming Language Models with Attention Sinks},
  author={Xiao, Guangxuan and Tian, Yuandong and Chen, Beidi and Han, Song and Lewis, Mike},
  booktitle={The Twelfth International Conference on Learning Representations},
  year={2024}
}

@article{cai2024llmaas,
  title={LLMaaS: Serving Large Language Models on Trusted Serverless Computing Platforms},
  author={Cai, Zinuo and Ma, Rongbo and Fu, Yicheng and Zhang, Weishan and Ma, Ruhui and Guan, Haibing},
  journal={IEEE Transactions on Artificial Intelligence},
  year={2024},
  publisher={IEEE}
}

@article{wu2025livelongbench,
  title={LiveLongBench: Tackling Long-Context Understanding for Spoken Texts from Live Streams},
  author={Wu, Yongxuan and Chen, Runyu and Liu, Peiyu and Qian, Hongjin},
  journal={arXiv preprint arXiv:2504.17366},
  year={2025}
}

@inproceedings{sheng2023flexgen,
  title={Flexgen: High-throughput generative inference of large language models with a single gpu},
  author={Sheng, Ying and Zheng, Lianmin and Yuan, Binhang and Li, Zhuohan and Ryabinin, Max and Chen, Beidi and Liang, Percy and R{\'e}, Christopher and Stoica, Ion and Zhang, Ce},
  booktitle={International Conference on Machine Learning},
  pages={31094--31116},
  year={2023},
  organization={PMLR}
}

@inproceedings{li2024eagle,
  title={EAGLE-2: Faster Inference of Language Models with Dynamic Draft Trees},
  author={Li, Yuhui and Wei, Fangyun and Zhang, Chao and Zhang, Hongyang},
  booktitle={Proceedings of the 2024 Conference on Empirical Methods in Natural Language Processing},
  pages={7421--7432},
  year={2024}
}

@article{zhao2025fr,
  title={Fr-spec: Accelerating large-vocabulary language models via frequency-ranked speculative sampling},
  author={Zhao, Weilin and Pan, Tengyu and Han, Xu and Zhang, Yudi and Sun, Ao and Huang, Yuxiang and Zhang, Kaihuo and Zhao, Weilun and Li, Yuxuan and Wang, Jianyong and others},
  journal={arXiv preprint arXiv:2502.14856},
  year={2025}
}

@article{zhao2024qspec,
  title={Qspec: Speculative decoding with complementary quantization schemes},
  author={Zhao, Juntao and Lu, Wenhao and Wang, Sheng and Kong, Lingpeng and Wu, Chuan},
  journal={arXiv preprint arXiv:2410.11305},
  year={2024}
}

@inproceedings{dao2024flashattention,
  title={FLASHATTENTION-2: FASTER ATTENTION WITH BETTER PARALLELISM AND WORK PARTITIONING},
  author={Dao, Tri},
  booktitle={12th International Conference on Learning Representations, ICLR 2024},
  year={2024}
}

@article{team2023gemini,
  title={Gemini: a family of highly capable multimodal models},
  author={Team, Gemini and Anil, Rohan and Borgeaud, Sebastian and Alayrac, Jean-Baptiste and Yu, Jiahui and Soricut, Radu and Schalkwyk, Johan and Dai, Andrew M and Hauth, Anja and Millican, Katie and others},
  journal={arXiv preprint arXiv:2312.11805},
  year={2023}
}

@article{chen2024octopus,
  title={Octopus v2: On-device language model for super agent},
  author={Chen, Wei and Li, Zhiyuan},
  journal={arXiv preprint arXiv:2404.01744},
  year={2024}
}

@article{murthy2024mobileaibench,
  title={Mobileaibench: Benchmarking llms and lmms for on-device use cases},
  author={Murthy, Rithesh and Yang, Liangwei and Tan, Juntao and Awalgaonkar, Tulika Manoj and Zhou, Yilun and Heinecke, Shelby and Desai, Sachin and Wu, Jason and Xu, Ran and Tan, Sarah and others},
  journal={arXiv preprint arXiv:2406.10290},
  year={2024}
}

@inproceedings{frantar2025marlin,
  title={Marlin: Mixed-precision auto-regressive parallel inference on large language models},
  author={Frantar, Elias and Castro, Roberto L and Chen, Jiale and Hoefler, Torsten and Alistarh, Dan},
  booktitle={Proceedings of the 30th ACM SIGPLAN Annual Symposium on Principles and Practice of Parallel Programming},
  pages={239--251},
  year={2025}
}

@article{zhao2023survey,
  title={A survey of large language models},
  author={Zhao, Wayne Xin and Zhou, Kun and Li, Junyi and Tang, Tianyi and Wang, Xiaolei and Hou, Yupeng and Min, Yingqian and Zhang, Beichen and Zhang, Junjie and Dong, Zican and others},
  journal={arXiv preprint arXiv:2303.18223},
  volume={1},
  number={2},
  year={2023}
}

@article{lu2024small,
  title={Small language models: Survey, measurements, and insights},
  author={Lu, Zhenyan and Li, Xiang and Cai, Dongqi and Yi, Rongjie and Liu, Fangming and Zhang, Xiwen and Lane, Nicholas D and Xu, Mengwei},
  journal={arXiv preprint arXiv:2409.15790},
  year={2024}
}

@article{van2024survey,
  title={A survey of small language models},
  author={Van Nguyen, Chien and Shen, Xuan and Aponte, Ryan and Xia, Yu and Basu, Samyadeep and Hu, Zhengmian and Chen, Jian and Parmar, Mihir and Kunapuli, Sasidhar and Barrow, Joe and others},
  journal={arXiv preprint arXiv:2410.20011},
  year={2024}
}

@inproceedings{strubell2020energy,
  title={Energy and policy considerations for modern deep learning research},
  author={Strubell, Emma and Ganesh, Ananya and McCallum, Andrew},
  booktitle={Proceedings of the AAAI conference on artificial intelligence},
  volume={34},
  number={09},
  pages={13693--13696},
  year={2020}
}

@article{yang2025mobileviclip,
  title={MobileViCLIP: An Efficient Video-Text Model for Mobile Devices},
  author={Yang, Min and Jia, Zihan and Dai, Zhilin and Guo, Sheng and Wang, Limin},
  journal={arXiv preprint arXiv:2508.07312},
  year={2025}
}

@article{jiang2020intelligent,
  title={Intelligent resource allocation for video analytics in blockchain-enabled internet of autonomous vehicles with edge computing},
  author={Jiang, Xiantao and Yu, F Richard and Song, Tian and Leung, Victor CM},
  journal={IEEE Internet of Things Journal},
  volume={9},
  number={16},
  pages={14260--14272},
  year={2020},
  publisher={IEEE}
}

@article{mao2017survey,
  title={A survey on mobile edge computing: The communication perspective},
  author={Mao, Yuyi and You, Changsheng and Zhang, Jun and Huang, Kaibin and Letaief, Khaled B},
  journal={IEEE communications surveys \& tutorials},
  volume={19},
  number={4},
  pages={2322--2358},
  year={2017},
  publisher={IEEE}
}

@article{tirumala2023d4,
  title={D4: Improving llm pretraining via document de-duplication and diversification},
  author={Tirumala, Kushal and Simig, Daniel and Aghajanyan, Armen and Morcos, Ari},
  journal={Advances in Neural Information Processing Systems},
  volume={36},
  pages={53983--53995},
  year={2023}
}

@article{dubey2024llama,
  title={The llama 3 herd of models},
  author={Dubey, Abhimanyu and Jauhri, Abhinav and Pandey, Abhinav and Kadian, Abhishek and Al-Dahle, Ahmad and Letman, Aiesha and Mathur, Akhil and Schelten, Alan and Yang, Amy and Fan, Angela and others},
  journal={arXiv e-prints},
  pages={arXiv--2407},
  year={2024}
}

@inproceedings{biderman2023pythia,
  title={Pythia: A suite for analyzing large language models across training and scaling},
  author={Biderman, Stella and Schoelkopf, Hailey and Anthony, Quentin Gregory and Bradley, Herbie and O’Brien, Kyle and Hallahan, Eric and Khan, Mohammad Aflah and Purohit, Shivanshu and Prashanth, USVSN Sai and Raff, Edward and others},
  booktitle={International Conference on Machine Learning},
  pages={2397--2430},
  year={2023},
  organization={PMLR}
}

@inproceedings{vasu2025fastvlm,
  title={Fastvlm: Efficient vision encoding for vision language models},
  author={Vasu, Pavan Kumar Anasosalu and Faghri, Fartash and Li, Chun-Liang and Koc, Cem and True, Nate and Antony, Albert and Santhanam, Gokula and Gabriel, James and Grasch, Peter and Tuzel, Oncel and others},
  booktitle={Proceedings of the Computer Vision and Pattern Recognition Conference},
  pages={19769--19780},
  year={2025}
}

@inproceedings{mehtaopenelm,
  title={OpenELM: An Efficient Language Model Family with Open Training and Inference Framework},
  author={Mehta, Sachin and Sekhavat, Mohammad Hossein and Cao, Qingqing and Horton, Maxwell and Jin, Yanzi and Sun, Chenfan and Mirzadeh, Seyed Iman and Najibi, Mahyar and Belenko, Dmitry and Zatloukal, Peter and others},
  booktitle={Workshop on Efficient Systems for Foundation Models II@ ICML2024},
  year={2024}
}

@article{team2024gemma,
  title={Gemma: Open models based on gemini research and technology},
  author={Team, Gemma and Mesnard, Thomas and Hardin, Cassidy and Dadashi, Robert and Bhupatiraju, Surya and Pathak, Shreya and Sifre, Laurent and Rivi{\`e}re, Morgane and Kale, Mihir Sanjay and Love, Juliette and others},
  journal={arXiv preprint arXiv:2403.08295},
  year={2024}
}

@article{mishra2024granite,
  title={Granite code models: A family of open foundation models for code intelligence},
  author={Mishra, Mayank and Stallone, Matt and Zhang, Gaoyuan and Shen, Yikang and Prasad, Aditya and Soria, Adriana Meza and Merler, Michele and Selvam, Parameswaran and Surendran, Saptha and Singh, Shivdeep and others},
  journal={arXiv preprint arXiv:2405.04324},
  year={2024}
}

@inproceedings{sun2021research,
  title={Research on Security Situation Analysis and Intelligent Disposal Technology of Edge Side Area},
  author={Sun, Xin and Li, Jiyuan and Li, Qinyuan},
  booktitle={2020 International Conference on Data Processing Techniques and Applications for Cyber-Physical Systems: DPTA 2020},
  pages={1563--1567},
  year={2021},
  organization={Springer}
}

@article{yao2024theoretical,
  title={Theoretical Insights into Fine-Tuning Attention Mechanism: Generalization and Optimization},
  author={Yao, Xinhao and Qian, Hongjin and Hu, Xiaolin and Xu, Gengze and Liu, Wei and Luan, Jian and Wang, Bin and Liu, Yong},
  journal={arXiv preprint arXiv:2410.02247},
  year={2024}
}

@article{cai2020tinytl,
  title={Tinytl: Reduce memory, not parameters for efficient on-device learning},
  author={Cai, Han and Gan, Chuang and Zhu, Ligeng and Han, Song},
  journal={Advances in Neural Information Processing Systems},
  volume={33},
  pages={11285--11297},
  year={2020}
}

@article{marone2025mmbert,
  title={mmBERT: A Modern Multilingual Encoder with Annealed Language Learning},
  author={Marone, Marc and Weller, Orion and Fleshman, William and Yang, Eugene and Lawrie, Dawn and Van Durme, Benjamin},
  journal={arXiv preprint arXiv:2509.06888},
  year={2025}
}

@article{yi2023edgemoe,
  title={Edgemoe: Fast on-device inference of moe-based large language models},
  author={Yi, Rongjie and Guo, Liwei and Wei, Shiyun and Zhou, Ao and Wang, Shangguang and Xu, Mengwei},
  journal={arXiv preprint arXiv:2308.14352},
  year={2023}
}

@inproceedings{tan2024mobilequant,
  title={MobileQuant: Mobile-friendly Quantization for On-device Language Models},
  author={Tan, Fuwen and Lee, Royson and Dudziak, {\L}ukasz and Hu, Shell Xu and Bhattacharya, Sourav and Hospedales, Timothy and Tzimiropoulos, Georgios and Martinez, Brais},
  booktitle={Findings of the Association for Computational Linguistics: EMNLP 2024},
  pages={9761--9771},
  year={2024}
}

@article{qu2025mobile,
  title={Mobile edge intelligence for large language models: A contemporary survey},
  author={Qu, Guanqiao and Chen, Qiyuan and Wei, Wei and Lin, Zheng and Chen, Xianhao and Huang, Kaibin},
  journal={IEEE Communications Surveys \& Tutorials},
  year={2025},
  publisher={IEEE}
}

@inproceedings{qin2024enabling,
  title={Enabling on-device large language model personalization with self-supervised data selection and synthesis},
  author={Qin, Ruiyang and Xia, Jun and Jia, Zhenge and Jiang, Meng and Abbasi, Ahmed and Zhou, Peipei and Hu, Jingtong and Shi, Yiyu},
  booktitle={Proceedings of the 61st ACM/IEEE Design Automation Conference},
  pages={1--6},
  year={2024}
}

@article{xiao2024understanding,
  title={Understanding Large Language Models in Your Pockets: Performance Study on COTS Mobile Devices},
  author={Xiao, Jie and Huang, Qianyi and Chen, Xu and Tian, Chen},
  journal={arXiv preprint arXiv:2410.03613},
  year={2024}
}

@article{pham2024slimlm,
  title={SlimLM: An Efficient Small Language Model for On-Device Document Assistance},
  author={Pham, Thang M and Nguyen, Phat T and Yoon, Seunghyun and Lai, Viet Dac and Dernoncourt, Franck and Bui, Trung},
  journal={arXiv preprint arXiv:2411.09944},
  year={2024}
}

@article{yi2025edgemoe,
  title={Edgemoe: Empowering sparse large language models on mobile devices},
  author={Yi, Rongjie and Guo, Liwei and Wei, Shiyun and Zhou, Ao and Wang, Shangguang and Xu, Mengwei},
  journal={IEEE Transactions on Mobile Computing},
  year={2025},
  publisher={IEEE}
}

@article{chen2024omnivlm,
  title={OmniVLM: A token-compressed, sub-billion-parameter vision-language model for efficient on-device inference},
  author={Chen, Wei and Li, Zhiyuan and Xin, Shuo},
  journal={arXiv preprint arXiv:2412.11475},
  year={2024}
}

@inproceedings{laskaridismobile,
  title={Mobile and edge evaluation of large language models},
  author={Laskaridis, Stefanos and Katevas, Kleomenis and Minto, Lorenzo and Haddadi, Hamed},
  booktitle={Workshop on Efficient Systems for Foundation Models II@ ICML2024},
  year={2024}
}

@article{ali2022federated,
  title={Federated learning for privacy preservation in smart healthcare systems: A comprehensive survey},
  author={Ali, Mansoor and Naeem, Faisal and Tariq, Muhammad and Kaddoum, Georges},
  journal={IEEE journal of biomedical and health informatics},
  volume={27},
  number={2},
  pages={778--789},
  year={2022},
  publisher={IEEE}
}

@inproceedings{li2024locmoe,
  title={LocMoE: a low-overhead MoE for large language model training},
  author={Li, Jing and Sun, Zhijie and He, Xuan and Zeng, Li and Lin, Yi and Li, Entong and Zheng, Binfan and Zhao, Rongqian and Chen, Xin},
  booktitle={Proceedings of the Thirty-Third International Joint Conference on Artificial Intelligence},
  pages={6377--6387},
  year={2024}
}

@article{shen2024jetmoe,
  title={Jetmoe: Reaching llama2 performance with 0.1 m dollars},
  author={Shen, Yikang and Guo, Zhen and Cai, Tianle and Qin, Zengyi},
  journal={arXiv preprint arXiv:2404.07413},
  year={2024}
}

@inproceedings{frantar2023gptq,
  title={GPTQ: Accurate Post-Training Quantization for Generative Pre-trained Transformers},
  author={Frantar, Elias and Ashkboos, Saleh and Hoefler, Torsten and Alistarh, Dan},
  booktitle={The Eleventh International Conference on Learning Representations},
  year={2023},
  organization={OpenReview}
}

@article{abdin2024phi,
  title={Phi-4 technical report},
  author={Abdin, Marah and Aneja, Jyoti and Behl, Harkirat and Bubeck, S{\'e}bastien and Eldan, Ronen and Gunasekar, Suriya and Harrison, Michael and Hewett, Russell J and Javaheripi, Mojan and Kauffmann, Piero and others},
  journal={arXiv preprint arXiv:2412.08905},
  year={2024}
}

@article{guo2025deepseek,
  title={DeepSeek-R1 incentivizes reasoning in LLMs through reinforcement learning},
  author={Guo, Daya and Yang, Dejian and Zhang, Haowei and Song, Junxiao and Zhang, Ruoyu and Xu, Runxin and Zhu, Qihao and Ma, Shirong and Wang, Peiyi and Bi, Xiao and others},
  doi = {10.1038/s41586-025-09422-z},
	isbn = {1476-4687},
	journal = {Nature},
	number = {8081},
	pages = {633--638}, 
	url = {https://doi.org/10.1038/s41586-025-09422-z},
	volume = {645},
	year = {2025}, 
}

@inproceedings{radford2021learning,
  title={Learning transferable visual models from natural language supervision},
  author={Radford, Alec and Kim, Jong Wook and Hallacy, Chris and Ramesh, Aditya and Goh, Gabriel and Agarwal, Sandhini and Sastry, Girish and Askell, Amanda and Mishkin, Pamela and Clark, Jack and others},
  booktitle={International conference on machine learning},
  pages={8748--8763},
  year={2021},
  organization={PmLR}
}

@article{ramesh2022hierarchical,
  title={Hierarchical text-conditional image generation with clip latents},
  author={Ramesh, Aditya and Dhariwal, Prafulla and Nichol, Alex and Chu, Casey and Chen, Mark},
  journal={arXiv preprint arXiv:2204.06125},
  volume={1},
  number={2},
  pages={3},
  year={2022}
}

@inproceedings{radford2023robust,
  title={Robust speech recognition via large-scale weak supervision},
  author={Radford, Alec and Kim, Jong Wook and Xu, Tao and Brockman, Greg and McLeavey, Christine and Sutskever, Ilya},
  booktitle={International conference on machine learning},
  pages={28492--28518},
  year={2023},
  organization={PMLR}
}

@inproceedings{gloeckle2024better,
  title={Better \& Faster Large Language Models via Multi-token Prediction},
  author={Gloeckle, Fabian and Idrissi, Badr Youbi and Roziere, Baptiste and Lopez-Paz, David and Synnaeve, Gabriel},
  booktitle={International Conference on Machine Learning},
  pages={15706--15734},
  year={2024},
  organization={PMLR}
}

@article{zhang2025speculative,
  title={Speculative decoding meets quantization: Compatibility evaluation and hierarchical framework design},
  author={Zhang, Yudi and Zhao, Weilin and Han, Xu and Zhao, Tiejun and Xu, Wang and Cao, Hailong and Zhu, Conghui},
  journal={arXiv preprint arXiv:2505.22179},
  year={2025}
}

@article{shazeer2017sparsely,
  title={The sparsely-gated mixture-of-experts layer},
  author={Shazeer, N and Mirhoseini, A and Maziarz, K and Davis, A and Le, Q and Hinton, G and Dean, J},
  journal={Outrageously large neural networks},
  volume={2},
  year={2017}
}

@article{roy2019towards,
  title={Towards spike-based machine intelligence with neuromorphic computing},
  author={Roy, Kaushik and Jaiswal, Akhilesh and Panda, Priyadarshini},
  journal={Nature},
  volume={575},
  number={7784},
  pages={607--617},
  year={2019},
  publisher={Nature Publishing Group UK London}
}

@article{belcak2025small,
  title={Small Language Models are the Future of Agentic AI},
  author={Belcak, Peter and Heinrich, Greg and Diao, Shizhe and Fu, Yonggan and Dong, Xin and Muralidharan, Saurav and Lin, Yingyan Celine and Molchanov, Pavlo},
  journal={arXiv preprint arXiv:2506.02153},
  year={2025}
}

@article{li2019edge,
  title={Edge AI: On-demand accelerating deep neural network inference via edge computing},
  author={Li, En and Zeng, Liekang and Zhou, Zhi and Chen, Xu},
  journal={IEEE transactions on wireless communications},
  volume={19},
  number={1},
  pages={447--457},
  year={2019},
  publisher={IEEE}
}

@article{han2024llm,
  title={LLM multi-agent systems: Challenges and open problems},
  author={Han, Shanshan and Zhang, Qifan and Yao, Yuhang and Jin, Weizhao and Xu, Zhaozhuo},
  journal={arXiv preprint arXiv:2402.03578},
  year={2024}
}

@article{yan2025beyond,
  title={Beyond self-talk: A communication-centric survey of llm-based multi-agent systems},
  author={Yan, Bingyu and Zhou, Zhibo and Zhang, Litian and Zhang, Lian and Zhou, Ziyi and Miao, Dezhuang and Li, Zhoujun and Li, Chaozhuo and Zhang, Xiaoming},
  journal={arXiv preprint arXiv:2502.14321},
  year={2025}
}

@article{shi2019edge,
  title={Edge computing},
  author={Shi, Weisong and Pallis, George and Xu, Zhiwei},
  journal={Proceedings of the IEEE},
  volume={107},
  number={8},
  pages={1474--1481},
  year={2019},
  publisher={IEEE}
}

@article{muhammad2021emotion,
  title={Emotion recognition for cognitive edge computing using deep learning},
  author={Muhammad, Ghulam and Hossain, M Shamim},
  journal={IEEE Internet of Things Journal},
  volume={8},
  number={23},
  pages={16894--16901},
  year={2021},
  publisher={IEEE}
}

@article{zheng2025review,
  title={A review on edge large language models: Design, execution, and applications},
  author={Zheng, Yue and Chen, Yuhao and Qian, Bin and Shi, Xiufang and Shu, Yuanchao and Chen, Jiming},
  journal={ACM Computing Surveys},
  volume={57},
  number={8},
  pages={1--35},
  year={2025},
  publisher={ACM New York, NY}
}

@article{xu2024device,
  title={On-device language models: A comprehensive review},
  author={Xu, Jiajun and Li, Zhiyuan and Chen, Wei and Wang, Qun and Gao, Xin and Cai, Qi and Ling, Ziyuan},
  journal={arXiv preprint arXiv:2409.00088},
  year={2024}
}

@misc{nvidia-h100-spec,
  title={NVIDIA H100 SXM5 GPU Technical Specifications},
  author={NVIDIA Corporation},
  year={2024},
  url={https://www.nvidia.com/en-us/data-center/h100/},
  note={Official specifications: 80GB HBM3, 3.35TB/s memory bandwidth, 1979 TOPS (Sparse INT8), 700W TDP}
}

@misc{jetson-agx-orin,
  title={NVIDIA Jetson AGX Orin Developer Kit Product Specifications},
  author={NVIDIA Corporation},
  year={2024},
  url={https://www.nvidia.com/en-us/autonomous-machines/embedded-systems/jetson-orin/},
  note={Technical specifications: 64GB unified memory, 204.8GB/s bandwidth, 275 TOPS AI performance, 60W max power}
}

@misc{apple-a17-pro,
  title={Apple A17 Pro SoC Technical Analysis and Performance Benchmarks},
  author={Apple Inc. and AnandTech},
  year={2024},
  url={https://en.wikipedia.org/wiki/Apple_A17},
  note={Neural Engine: 35 TOPS; System memory: 8GB; Estimated memory bandwidth: ~68GB/s; Process: TSMC N3B 3nm}
}

@article{li2025sled,
  title={SLED: A Speculative LLM Decoding Framework for Efficient Edge Serving},
  author={Li, Xiangchen and Spatharakis, Dimitrios and Ghafouri, Saeid and Fan, Jiakun and Vandierendonck, Hans and John, Deepu and Ji, Bo and Nikolopoulos, Dimitrios},
  journal={arXiv preprint arXiv:2506.09397},
  year={2025}
}

@inproceedings{zengglm,
  title={GLM-130B: An Open Bilingual Pre-trained Model},
  author={Zeng, Aohan and Liu, Xiao and Du, Zhengxiao and Wang, Zihan and Lai, Hanyu and Ding, Ming and Yang, Zhuoyi and Xu, Yifan and Zheng, Wendi and Xia, Xiao and others},
  booktitle={The Eleventh International Conference on Learning Representations},
  year={2023}
}

@article{bai2023qwen,
  title={Qwen technical report},
  author={Bai, Jinze and Bai, Shuai and Chu, Yunfei and Cui, Zeyu and Dang, Kai and Deng, Xiaodong and Fan, Yang and Ge, Wenbin and Han, Yu and Huang, Fei and others},
  journal={arXiv preprint arXiv:2309.16609},
  year={2023}
}

@article{ccoplu2023performance,
  title={A performance evaluation of a quantized large language model on various smartphones},
  author={{\c{C}}{\"o}pl{\"u}, Tolga and Loedi, Marc and Bendiken, Arto and Makohin, Mykhailo and Bouw, Joshua J and Cobb, Stephen},
  journal={arXiv preprint arXiv:2312.12472},
  year={2023}
}

@article{xu2023llmcad,
  title={Llmcad: Fast and scalable on-device large language model inference},
  author={Xu, Daliang and Yin, Wangsong and Jin, Xin and Zhang, Ying and Wei, Shiyun and Xu, Mengwei and Liu, Xuanzhe},
  journal={arXiv preprint arXiv:2309.04255},
  year={2023}
}

@article{sun2025disco,
  title={DiSCo: Device-Server Collaborative LLM-Based Text Streaming Services},
  author={Sun, Ting and Wang, Penghan and Lai, Fan},
  journal={arXiv preprint arXiv:2502.11417},
  year={2025}
}

@article{li2024personal,
  title={Personal llm agents: Insights and survey about the capability, efficiency and security},
  author={Li, Yuanchun and Wen, Hao and Wang, Weijun and Li, Xiangyu and Yuan, Yizhen and Liu, Guohong and Liu, Jiacheng and Xu, Wenxing and Wang, Xiang and Sun, Yi and others},
  journal={arXiv preprint arXiv:2401.05459},
  year={2024}
}

@article{friha2024llm,
  title={Llm-based edge intelligence: A comprehensive survey on architectures, applications, security and trustworthiness},
  author={Friha, Othmane and Ferrag, Mohamed Amine and Kantarci, Burak and Cakmak, Burak and Ozgun, Arda and Ghoualmi-Zine, Nassira},
  journal={IEEE Open Journal of the Communications Society},
  year={2024},
  publisher={IEEE}
}

@article{tran2025multi,
  title={Multi-agent collaboration mechanisms: A survey of llms},
  author={Tran, Khanh-Tung and Dao, Dung and Nguyen, Minh-Duong and Pham, Quoc-Viet and O'Sullivan, Barry and Nguyen, Hoang D},
  journal={arXiv preprint arXiv:2501.06322},
  year={2025}
}

@article{shuvo2022efficient,
  title={Efficient acceleration of deep learning inference on resource-constrained edge devices: A review},
  author={Shuvo, Md Maruf Hossain and Islam, Syed Kamrul and Cheng, Jianlin and Morshed, Bashir I},
  journal={Proceedings of the IEEE},
  volume={111},
  number={1},
  pages={42--91},
  year={2022},
  publisher={IEEE}
}

@article{fan2025parallel,
  title={Parallel CPU-GPU Execution for LLM Inference on Constrained GPUs},
  author={Fan, Jiakun and Zhang, Yanglin and Li, Xiangchen and Nikolopoulos, Dimitrios S},
  journal={arXiv preprint arXiv:2506.03296},
  year={2025}
}

@article{li2025qpart,
  title={QPART: Adaptive Model Quantization and Dynamic Workload Balancing for Accuracy-aware Edge Inference},
  author={Li, Xiangchen and Ghafouri, Saeid and Ji, Bo and Vandierendonck, Hans and John, Deepu and Nikolopoulos, Dimitrios S},
  journal={arXiv preprint arXiv:2506.23934},
  year={2025}
}

@article{andong2025federated,
  title={Federated Multi-Agent Reinforcement Learning for Privacy-Preserving and Energy-Aware Resource Management in 6G Edge Networks},
  author={Andong, Francisco Javier Esono Nkulu and Min, Qi},
  journal={arXiv preprint arXiv:2509.10163},
  year={2025}
}

@article{hu2025pfl,
  title={pFL-SBPM: A communication-efficient personalized federated learning framework for resource-limited edge clients},
  author={Hu, Han and Du, Wenli and Li, Yuqiang and Wang, Yue},
  journal={Future Generation Computer Systems},
  volume={171},
  pages={107849},
  year={2025},
  publisher={Elsevier}
}

@inproceedings{liu2024edge,
  title={Edge-LLMs: Edge-device large language model competition},
  author={Liu, Shiwei and Han, Kai and Fernandez-Lopez, Adriana and Jaiswal, Ajay Kumar and Atashgahi, Zahra and Wu, Boqian and Ponti, Edoardo and Hao, Callie and Burkholz, Rebekka and Saukh, Olga and others},
  booktitle={NeurIPS 2024 Competition Track},
  year={2024}
}

@article{hu2022lora,
  title={Lora: Low-rank adaptation of large language models.},
  author={Hu, Edward J and Shen, Yelong and Wallis, Phillip and Allen-Zhu, Zeyuan and Li, Yuanzhi and Wang, Shean and Wang, Lu and Chen, Weizhu and others},
  journal={ICLR},
  volume={1},
  number={2},
  pages={3},
  year={2022}
}

@inproceedings{houlsby2019parameter,
  title={Parameter-efficient transfer learning for NLP},
  author={Houlsby, Neil and Giurgiu, Andrei and Jastrzebski, Stanislaw and Morrone, Bruna and De Laroussilhe, Quentin and Gesmundo, Andrea and Attariyan, Mona and Gelly, Sylvain},
  booktitle={International conference on machine learning},
  pages={2790--2799},
  year={2019},
  organization={PMLR}
}

@article{le2024exploring,
  title={Exploring the practicality of federated learning: A survey towards the communication perspective},
  author={Le, Khiem and Luong-Ha, Nhan and Nguyen-Duc, Manh and Le-Phuoc, Danh and Do, Cuong and Wong, Kok-Seng},
  journal={arXiv preprint arXiv:2405.20431},
  year={2024}
}

@inproceedings{li2022dag,
  title={DAG-based task orchestration for edge computing},
  author={Li, Xiang and Abdallah, Mustafa and Suryavansh, Shikhar and Chiang, Mung and Kim, Kwang Taik and Bagchi, Saurabh},
  booktitle={2022 41st International Symposium on Reliable Distributed Systems (SRDS)},
  pages={23--34},
  year={2022},
  organization={IEEE}
}

@article{shen2023hugginggpt,
  title={Hugginggpt: Solving ai tasks with chatgpt and its friends in hugging face},
  author={Shen, Yongliang and Song, Kaitao and Tan, Xu and Li, Dongsheng and Lu, Weiming and Zhuang, Yueting},
  journal={Advances in Neural Information Processing Systems},
  volume={36},
  pages={38154--38180},
  year={2023}
}

@inproceedings{xu2025breaking,
  title={Breaking the Layer Barrier: Remodeling Private Transformer Inference with Hybrid $\{$CKKS$\}$ and $\{$MPC$\}$},
  author={Xu, Tianshi and Lu, Wen-jie and Yu, Jiangrui and Chen, Yi and Lin, Chenqi and Wang, Runsheng and Li, Meng},
  booktitle={34th USENIX Security Symposium (USENIX Security 25)},
  pages={2653--2672},
  year={2025}
}

@article{zhang2025bring,
  title={Bring Your Device Group (BYDG): Efficient and Privacy-Preserving User-Device Authentication Protocol in Multi-Access Edge Computing},
  author={Zhang, Yan and Gu, Chunsheng and Shi, Peizhong and Jing, Zhengjun and Li, Bing and Liu, Bo},
  journal={IEEE Transactions on Information Forensics and Security},
  year={2025},
  publisher={IEEE}
}

@article{yuan2024llm,
  title={Llm inference unveiled: Survey and roofline model insights},
  author={Yuan, Zhihang and Shang, Yuzhang and Zhou, Yang and Dong, Zhen and Zhou, Zhe and Xue, Chenhao and Wu, Bingzhe and Li, Zhikai and Gu, Qingyi and Lee, Yong Jae and others},
  journal={arXiv preprint arXiv:2402.16363},
  year={2024}
}

@article{wu2025efficient,
  title={Efficient pretraining length scaling},
  author={Wu, Bohong and Yan, Shen and Zhang, Sijun and Lu, Jianqiao and Zeng, Yutao and Wang, Ya and Zhou, Xun},
  journal={arXiv preprint arXiv:2504.14992},
  year={2025}
}

@article{zhang2023camel,
  title={Camel: Co-designing ai models and embedded drams for efficient on-device learning},
  author={Zhang, Sai Qian and Tambe, Thierry and Cuevas, Nestor and Wei, Gu-Yeon and Brooks, David},
  journal={arXiv preprint arXiv:2305.03148},
  year={2023}
}

@inproceedings{lozhkovsmollm2,
  title={SmolLM2: When Smol Goes Big—Data-Centric Training of a Fully Open Small Language Model},
  author={Lozhkov, Anton and Bakouch, Elie and Blazquez, Gabriel Martin and Penedo, Guilherme and Tunstall, Lewis and Marafioti, Andr{\'e}s and Lajar{\'\i}n, Agust{\'\i}n Piqueres and Kydl{\'\i}{\v{c}}ek, Hynek and Srivastav, Vaibhav and Lochner, Joshua and others},
  booktitle={Second Conference on Language Modeling}
}

@article{faghri2025mobileclip2,
  title={MobileCLIP2: Improving Multi-Modal Reinforced Training},
  author={Faghri, Fartash and Vasu, Pavan Kumar Anasosalu and Koc, Cem and Shankar, Vaishaal and Toshev, Alexander and Tuzel, Oncel and Pouransari, Hadi},
  journal={arXiv preprint arXiv:2508.20691},
  year={2025}
}

@misc{masayoshi2025ehrmcprealworldevaluationclinical,
      title={EHR-MCP: Real-world Evaluation of Clinical Information Retrieval by Large Language Models via Model Context Protocol}, 
      author={Kanato Masayoshi and Masahiro Hashimoto and Ryoichi Yokoyama and Naoki Toda and Yoshifumi Uwamino and Shogo Fukuda and Ho Namkoong and Masahiro Jinzaki},
      year={2025},
      eprint={2509.15957},
      archivePrefix={arXiv},
      primaryClass={cs.AI},
      url={https://arxiv.org/abs/2509.15957}, 
}

@misc{liu2025flemingr1expertlevelmedicalreasoning,
      title={Fleming-R1: Toward Expert-Level Medical Reasoning via Reinforcement Learning}, 
      author={Chi Liu and Derek Li and Yan Shu and Robin Chen and Derek Duan and Teng Fang and Bryan Dai},
      year={2025},
      eprint={2509.15279},
      archivePrefix={arXiv},
      primaryClass={cs.LG},
      url={https://arxiv.org/abs/2509.15279}, 
}

@article{yu2025frame,
  title={FRAME: Feedback-Refined Agent Methodology for Enhancing Medical Research Insights},
  author={Yu, Chengzhang and Zhang, Yiming and Liu, Zhixin and Ding, Zenghui and Sun, Yining and Jin, Zhanpeng},
  journal={arXiv preprint arXiv:2505.04649},
  year={2025}
}

@article{gueriani2025robust,
  title={A robust cross-domain IDS using BiGRU-LSTM-attention for medical and industrial IoT security},
  author={Gueriani, Afrah and Kheddar, Hamza and Mazari, Ahmed Cherif and Ghanem, Mohamed Chahine},
  journal={ICT Express},
  year={2025},
  publisher={Elsevier}
}

@article{cai2024driving,
  title={Driving with regulation: Interpretable decision-making for autonomous vehicles with retrieval-augmented reasoning via llm},
  author={Cai, Tianhui and Liu, Yifan and Zhou, Zewei and Ma, Haoxuan and Zhao, Seth Z and Wu, Zhiwen and Ma, Jiaqi},
  journal={arXiv preprint arXiv:2410.04759},
  year={2024}
}

@inproceedings{huang2024drivlme,
  title={Drivlme: Enhancing llm-based autonomous driving agents with embodied and social experiences},
  author={Huang, Yidong and Sansom, Jacob and Ma, Ziqiao and Gervits, Felix and Chai, Joyce},
  booktitle={2024 IEEE/RSJ International Conference on Intelligent Robots and Systems (IROS)},
  pages={3153--3160},
  year={2024},
  organization={IEEE}
}

@article{hu2025llm,
  title={Llm-based misbehavior detection architecture for enhanced traffic safety in connected autonomous vehicles},
  author={Hu, Yaqi and Wang, Fei and Ye, Dongdong and Wu, Maoqiang and Kang, Jiawen and Yu, Rong},
  journal={IEEE Transactions on Vehicular Technology},
  year={2025},
  publisher={IEEE}
}

@article{chiu2025v2v,
  title={V2v-llm: Vehicle-to-vehicle cooperative autonomous driving with multi-modal large language models},
  author={Chiu, Hsu-kuang and Hachiuma, Ryo and Wang, Chien-Yi and Smith, Stephen F and Wang, Yu-Chiang Frank and Chen, Min-Hung},
  journal={arXiv preprint arXiv:2502.09980},
  year={2025}
}

@article{yang2025chatdl,
  title={ChatDL: An LLM-Based Defect Localization Approach for Software in IIoT Flexible Manufacturing},
  author={Yang, Haiyang and Zhou, Yulu and Liang, Tian and Kuang, Li},
  journal={IEEE Internet of Things Journal},
  year={2025},
  publisher={IEEE}
}

@article{tang2025towards,
  title={Towards General Industrial Intelligence: A Survey of Large Models as a Service in Industrial IoT},
  author={Tang, Jianhua and Chen, Jiao and He, Jiayi and Chen, Fangfang and Lv, Zuohong and Han, Guangjie and Liu, Zuozhu and Yang, Howard H and Li, Weihua},
  journal={IEEE Communications Surveys \& Tutorials},
  year={2025},
  publisher={IEEE}
}

@article{li2025urban,
  title={Urban Computing in the Era of Large Language Models},
  author={Li, Zhonghang and Xia, Lianghao and Ren, Xubin and Tang, Jiabin and Chen, Tianyi and Xu, Yong and Huang, Chao},
  journal={ACM Transactions on Intelligent Systems and Technology},
  year={2025},
  publisher={ACM New York, NY}
}

@article{zheng2025urban,
  title={Urban planning in the era of large language models},
  author={Zheng, Yu and Xu, Fengli and Lin, Yuming and Santi, Paolo and Ratti, Carlo and Wang, Qi R and Li, Yong},
  journal={Nature Computational Science},
  pages={1--10},
  year={2025},
  publisher={Nature Publishing Group US New York}
}

@article{tzachor2023large,
  title={Large language models and agricultural extension services},
  author={Tzachor, Asaf and Devare, Medha and Richards, Catherine and Pypers, Pieter and Ghosh, Aniruddha and Koo, Jawoo and Johal, S and King, Brian},
  journal={Nature food},
  volume={4},
  number={11},
  pages={941--948},
  year={2023},
  publisher={Nature Publishing Group UK London}
}

@article{yang2025agrigpt,
  title={AgriGPT: a Large Language Model Ecosystem for Agriculture},
  author={Yang, Bo and Zhang, Yu and Feng, Lanfei and Chen, Yunkui and Zhang, Jianyu and Xu, Xiao and Aierken, Nueraili and Li, Yurui and Chen, Yuxuan and Yang, Guijun and others},
  journal={arXiv preprint arXiv:2508.08632},
  year={2025}
}

@article{yuan2025pezego,
  title={PEZEGO: A precision agriculture system based on large language models and internet of things for pest management},
  author={Yuan, Zhipeng and Liu, Kang and Li, Shunbao and Peng, Ruoling and Leybourne, Daniel and Musa, Nasamu and Yang, Po},
  journal={IEEE Internet of Things Journal},
  year={2025},
  publisher={IEEE}
}

@inproceedings{zhou2024agribench,
  title={Agribench: A hierarchical agriculture benchmark for multimodal large language models},
  author={Zhou, Yutong and Ryo, Masahiro},
  booktitle={European Conference on Computer Vision},
  pages={207--223},
  year={2024},
  organization={Springer}
}

@inproceedings{li2025cusmer,
  title={CuSMer: Multimodal Intent Recognition in Customer Service via Data Augment and LLM Merge},
  author={Li, Zhipeng and Wu, Binglin and Zhang, Yingyi and Li, Xianneng and Li, Kai and Chen, Weizhi},
  booktitle={Companion Proceedings of the ACM on Web Conference 2025},
  pages={3058--3062},
  year={2025}
}

@inproceedings{farfade2024scaling,
  title={Scaling use-case based shopping using LLMs},
  author={Farfade, Sachin and Vernekar, Sachin and Chaoji, Vineet and Mukherjee, Rajdeep},
  booktitle={Proceedings of the 17th ACM International Conference on Web Search and Data Mining},
  pages={1165--1166},
  year={2024}
}

@inproceedings{zhang2025does,
  title={How does Search Affect Personalized Recommendations and User Behavior? Evidence from LLM-based Synthetic Data},
  author={Zhang, Haoran and Kang, Xin and Guo, Junpeng},
  booktitle={Companion Proceedings of the ACM on Web Conference 2025},
  pages={2434--2443},
  year={2025}
}

@article{zhao2025llm,
  title={Llm app store analysis: A vision and roadmap},
  author={Zhao, Yanjie and Hou, Xinyi and Wang, Shenao and Wang, Haoyu},
  journal={ACM Transactions on Software Engineering and Methodology},
  volume={34},
  number={5},
  pages={1--25},
  year={2025},
  publisher={ACM New York, NY}
}

@article{mao2025deepwriter,
  title={DeepWriter: A Fact-Grounded Multimodal Writing Assistant Based On Offline Knowledge Base},
  author={Mao, Song and Cheng, Lejun and Cai, Pinlong and Yan, Guohang and Wang, Ding and Shi, Botian},
  journal={arXiv preprint arXiv:2507.14189},
  year={2025}
}

@article{yao2024minicpm,
  title={Minicpm-v: A gpt-4v level mllm on your phone},
  author={Yao, Yuan and Yu, Tianyu and Zhang, Ao and Wang, Chongyi and Cui, Junbo and Zhu, Hongji and Cai, Tianchi and Li, Haoyu and Zhao, Weilin and He, Zhihui and others},
  journal={arXiv preprint arXiv:2408.01800},
  year={2024}
}

@article{ruan2024webllm,
  title={WebLLM: A High-Performance In-Browser LLM Inference Engine},
  author={Ruan, Charlie F and Qin, Yucheng and Zhou, Xun and Lai, Ruihang and Jin, Hongyi and Dong, Yixin and Hou, Bohan and Yu, Meng-Shiun and Zhai, Yiyan and Agarwal, Sudeep and others},
  journal={arXiv preprint arXiv:2412.15803},
  year={2024}
}

@techreport{rockchip_ds_brief,
  author       = {Rockchip Electronics Co., Ltd.},
  title        = {RK3588 Brief Datasheet},
  institution  = {Rockchip Electronics},
  year         = {2022},
  type         = {Datasheet},
  howpublished = {\url{https://www.rock-chips.com/uploads/pdf/2022.8.26/192/RK3588 Brief Datasheet.pdf}},
}

@article{rockchip_cnxsoc,
  author       = {CNX Software},
  title        = {Rockchip RK3588 SoC Datasheet Reveals 6\,TOPS NPU and 8K Video Capabilities},
  journal      = {CNX Software},
  year         = {2021},
  note         = {Online article},
  howpublished = {\url{https://www.cnx-software.com/2021/12/16/rockchip-rk3588-datasheet-sbc-coming-soon/}},
}

@misc{rockchip_ieisb,
  author       = {IEI Integration Corp.},
  title        = {WAFER-RK3588 Industrial SBC Specification (RK3588, 6 TOPS NPU, Mixed Precision)},
  howpublished = {\url{https://www.ieiworld.com/en/product/model.php?II=1036}},
  year         = {2024},
  note         = {Industrial SBC Spec Sheet},
}

@article{chen2025inference,
  title={Inference performance evaluation for LLMs on edge devices with a novel benchmarking framework and metric},
  author={Chen, Hao and Tian, Cong and He, Zixuan and Yu, Bin and Liu, Yepang and Cao, Jialun},
  journal={arXiv preprint arXiv:2508.11269},
  year={2025}
}

@misc{llama_cpp_repo,
  author       = {G. Gerganov et al.},
  title        = {\texttt{llama.cpp}: Local Inference of LLMs on Diverse Hardware},
  howpublished = {\url{https://github.com/ggml-org/llama.cpp}},
  year         = {2023},
}

@article{joshi2025neuro,
  title={Neuro-LIFT: A neuromorphic, LLM-based interactive framework for autonomous drone flight at the edge},
  author={Joshi, Amogh and Sanyal, Sourav and Roy, Kaushik},
  journal={arXiv preprint arXiv:2501.19259},
  year={2025}
}

@article{li2025next,
  title={What Is Next for LLMs? Next-Generation AI Computing Hardware Using Photonic Chips},
  author={Li, Renjie and Wei, Wenjie and Xin, Qi and Liu, Xiaoli and Mao, Sixuan and Ma, Erik and Chen, Zijian and Zhang, Malu and Li, Haizhou and Zhang, Zhaoyu},
  journal={arXiv preprint arXiv:2505.05794},
  year={2025}
}

@article{kong2025quantum,
  title={Quantum-enhanced llm efficient fine tuning},
  author={Kong, Xiaofei and Li, Lei and Chen, Zhaoyun and Xue, Cheng and Xu, Xiaofan and Liu, Huanyu and Wu, Yuchun and Fang, Yuan and Fang, Han and Chen, Kejiang and others},
  journal={arXiv preprint arXiv:2503.12790},
  year={2025}
}

@article{jin2025innovative,
  title={An Innovative Brain-Computer Interface Interaction System Based on the Large Language Model},
  author={Jin, Jing and Zhang, Yutao and Xu, Ruitian and Chen, Yixin},
  journal={arXiv preprint arXiv:2502.11659},
  year={2025}
}

@article{alsuleman2025screening,
  title={Screening of Atrial Fibrillation using Wearable PPG Devices--a Trustworthy and Safe AI Life Cycle Case Study},
  author={Alsuleman, M and Duncan, P and Thompson, A},
  year={2025}
}

@article{liu2025towards,
  title={Towards Harnessing the Collaborative Power of Large and Small Models for Domain Tasks},
  author={Liu, Yang and Yan, Bingjie and Zou, Tianyuan and Zhang, Jianqing and Gu, Zixuan and Ding, Jianbing and Wang, Xidong and Li, Jingyi and Ye, Xiaozhou and Ouyang, Ye and others},
  journal={arXiv preprint arXiv:2504.17421},
  year={2025}
}

@inproceedings{chen2025multi,
  title={Multi-modal Medical Diagnosis via Large-small Model Collaboration},
  author={Chen, Wanyi and Zhao, Zihua and Yao, Jiangchao and Zhang, Ya and Bu, Jiajun and Wang, Haishuai},
  booktitle={Proceedings of the Computer Vision and Pattern Recognition Conference},
  pages={30763--30773},
  year={2025}
}

@inproceedings{liu2024cotuning,
  title={CoTuning: A Large-Small Model Collaborating Distillation Framework for Better Model Generalization},
  author={Liu, Zimo and Liu, Kangjun and Guo, Mingyue and Zhang, Shiliang and Wang, Yaowei},
  booktitle={Proceedings of the 32nd ACM International Conference on Multimedia},
  pages={10487--10496},
  year={2024}
}

@article{wang2025collm,
  title={Collm: Industrial large-small model collaboration with fuzzy decision-making agent and self-reflection},
  author={Wang, Haiteng and Ren, Lei and Zhao, Tuo and Jiao, Lu},
  journal={IEEE Transactions on Fuzzy Systems},
  year={2025},
  publisher={IEEE}
}

@article{tian2024edge,
  title={An edge-cloud collaboration framework for generative ai service provision with synergetic big cloud model and small edge models},
  author={Tian, Yuqing and Zhang, Zhaoyang and Yang, Yuzhi and Chen, Zirui and Yang, Zhaohui and Jin, Richeng and Quek, Tony QS and Wong, Kai-Kit},
  journal={IEEE Network},
  volume={38},
  number={5},
  pages={37--46},
  year={2024},
  publisher={IEEE}
}

@article{wang2025llm,
  title={LLM-based HSE Compliance Assessment: Benchmark, Performance, and Advancements},
  author={Wang, Jianwei and Wang, Mengqi and Zhou, Yinsi and Xing, Zhenchang and Liu, Qing and Xu, Xiwei and Zhang, Wenjie and Zhu, Liming},
  journal={arXiv preprint arXiv:2505.22959},
  year={2025}
}

@article{xiao2024infllm,
  title={Infllm: Training-free long-context extrapolation for llms with an efficient context memory},
  author={Xiao, Chaojun and Zhang, Pengle and Han, Xu and Xiao, Guangxuan and Lin, Yankai and Zhang, Zhengyan and Liu, Zhiyuan and Sun, Maosong},
  journal={Advances in Neural Information Processing Systems},
  volume={37},
  pages={119638--119661},
  year={2024}
}

@article{zhao2025mobilellm,
  title={MobileLLM-R1: Exploring the Limits of Sub-Billion Language Model Reasoners with Open Training Recipes},
  author={Zhao, Changsheng and Chang, Ernie and Liu, Zechun and Chang, Chia-Jung and Wen, Wei and Lai, Chen and Cao, Rick and Tian, Yuandong and Krishnamoorthi, Raghuraman and Shi, Yangyang and others},
  journal={arXiv preprint arXiv:2509.24945},
  year={2025}
}


\end{document}